\documentclass[a4paper,12pt,twoside]{book}
\input{Preambule}
\input{PreambuleGarde}

\usepackage[acronym, automake]{glossaries-extra}
\makeglossaries
\newacronym{CNN}{CNN}{Convolutional Neuronal Network}

\newacronym{C3D}{C3D}{Convolutional 3 Dimensions}

\newacronym{CNIL}{CNIL}{Commission Nationale de l'Informatique et des Libertés}

\newacronym{CRNN}{CRNN}{Convolutional Recurrent Neuronal Network}

\newacronym{CGRU}{CGRU}{Convolutional Gated Recurrent Unit}

\newacronym{FPS}{FPS}{Frame Per Second}

\newacronym{GAN}{GAN}{Generative Adversarial Networks}

\newacronym{GB}{GB}{Giga Byte}

\newacronym{GHz}{GHz}{Giga Hertz}

\newacronym{GRU}{GRU}{Gated Recurrent Unit}

\newacronym{KNN}{KNN}{K-Nearest Neighbours}

\newacronym{LIASD}{LIASD}{Laboratoire d'Intelligence Artificielle et Science des Données}

\newacronym{LSTM}{LSTM}{Long Short Term Memory}

\newacronym{MLP}{MLP}{Multi Layer Perceptron}

\newacronym{RAM}{RAM}{Random Acess Memory}

\newacronym{RCNN}{RCNN}{Regional Convolutional Neuronal Network}

\newacronym{RGPD}{RGPD}{Règlement Général sur la Protection des Données}

\newacronym{RNN}{RNN}{Recurrent Neronal Network}

\newacronym{RoI}{RoI}{Region of Interest}

\newacronym{SAM}{SAM}{Segment Anything Mode}

\newacronym{TCN}{TCN}{Temporal Convolutional Network}

\newacronym{SVM}{SVM}{Support Vector Machine}

\newacronym{VGG}{VGG}{Visual Geometry Group}

\newacronym{ViT}{ViT}{Vision Transformer}

\newacronym{YOLO}{YOLO}{You Only Looks Once}

\newglossaryentry{DPM}{
name = {Deformable Parts Model (DPM)},
description = { : is an object detection algorithm that uses deformable parts models to represent objects in an image, allowing it to handle variations in shape and size in the training data.},
first =  {Deformable Parts Model (DPM)},
text = {DPM}
}

\newglossaryentry{DIoU}{
name = {Distance Intersection over Union (DIoU)},
description = { : This is a variant of IoU that measures the similarity between two sets by considering both the distance between the centers of the regions and their intersection.},
first = {Distance Intersection over Union (DIoU)},
text = {DIoU}
}

\newglossaryentry{DINO}{
name = {DIstillation with NO label (DINO)},
description = { : A learning technique where a model called the teacher model is trained to transfer its knowledge to a smaller model called the student model. The student model learns to mimic the teacher model's predictions using the teacher's predictions as supervision, rather than using labeled data.},
first = {DIstillation with NO label (DINO)},
text = {DINO}
}

\newglossaryentry{HOG}{
name = {Histogram of Oriented Gradients (HOG)},
description = { : is an image descriptor technique used in computer vision to detect object contours in an image. It uses a histogram of oriented gradients to represent the local features of the image.},
first = {Histogram of Oriented Gradients (HOG)},
text = {HOG}
}

\newglossaryentry{IoU}{
name = {Intersection over Union (IoU)},
description = { : A measure of similarity between two sets used to evaluate the quality of object detections; IoU measures the proportion of overlap between the region predicted by the model and the actual region of the object.},
first = {Intersection over Union (IoU)},
text = {IoU}
}

\newglossaryentry{MAP}{
name = {Mean Average Precision (MAP)},
description = { : A global measure of object detection quality that accounts for both precision (the number of correct detections) and recall (the number of detections over the total number of objects). The average precision for each class is used to evaluate the object detection model.},
first = {Mean Average Precision (MAP)},
text = {MAP}
}

\newglossaryentry{NMS}{
name = {Non-Max Suppression (NMS)},
description = { : A technique used in object detection to eliminate overlapping regions (or duplicated regions) by keeping only the region with the highest probability of being the target object.},
first = {Non-Max Suppression (NMS)},
text = {NMS}
}

\newglossaryentry{DIoU-NMS}{
name = {Distance Intersection over Union Non-Max Suppression (DIoU NMS)},
description = { : This is a combination of DIoU and NMS used to eliminate overlapping regions while using a more precise similarity measure.},
first = {Distance Intersection over Union Non-Max Suppression (DIoU NMS)},
text = {DIoU-NMS}
}

\newglossaryentry{CIoU}{
name = {Complete Intersection over Union (CIoU)},
description = { : This is an improvement of IoU that takes into account the distance between the centers of the predicted and actual regions as well as their orientation. CIoU provides a more accurate measure of the quality of object detections.},
first = {Complete Intersection over Union (CIoU)},
text = {CIoU}
}

\graphicspath{{images/}}

\author{Fabien \textsc{Poirier}}
\fulltitle{Real-Time Anomaly Detection in Video Streams}
\title{Anomaly Detection in Videos}
\specialite{Computer Science}
\directeur{Professeur Gilles \textsc{Bernard}, U. Paris 8, Paragraphe}
\encadrant{Assistant professor Rakia \textsc{Jaziri}, U. Paris 8, Paragraphe}
\date{14/09/2023}

\jurya{M. Pierre \textsc{Gançarski}}{Professor}{U. Strasbourg}{ICube}{President}
\juryb{M. Kurosh \textsc{Madani}}{Professor}{U. Paris Est Créteil}{LISSI}{Reviewer}
\juryc{M. Mustapha \textsc{Lebbah}}{Professor}{U. Paris Saclay}{David}{Reviewer}
\juryd{M. Haythem \textsc{Elghazel}}{Assistant professor}{U. Lyon 1}{LIRIS}{Examiner}
\jurye{Mme Rakia \textsc{Jaziri}}{Assistant professor}{U. Paris 8}{Paragraphe}{Co-director}
\juryf{M. Gilles \textsc{Bernard}}{Professor}{U. Paris 8}{Paragraphe}{Director}

\ecole{Paris 8 University, Saint-Denis}

\begin{document}

\fancyhead{}
\fancyhead[RO,LE]{\thepage}
\fancyhead[RE]{\textit{\leftmark}}
\fancyhead[LO]{\textit{\rightmark}}
\renewcommand{\headrulewidth}{0.4pt}

\fancyfoot{}
\fancyfoot[LE]{\theauthor}
\fancyfoot[RO]{2023}
\fancyfoot[LO]{\scshape Anomaly Detection in Videos}
\fancyfoot[RE]{Paris 8 University}

\fancyhfoffset[LO]{0in}

\dominitoc
\doparttoc

\pagedegarde

\csname @openrightfalse\endcsname
\section*{Acknowledgements} 
\markboth{Acknowledgements}{} 
\addcontentsline{toc}{section}{\protect{Acknowledgements}}

\noindent I would like to thank all the individuals who contributed to the success of this thesis, starting with my supervisors, Ms. Rakia Jaziri and Mr. Gilles Bernard. \ \vspace{1\baselineskip}

\noindent My company Othello, and especially its leader Camille Srour, for his trust and for ensuring that this thesis was conducted under the best possible conditions by providing all the necessary tools. \ \vspace{1\baselineskip}

\noindent Furthermore, I would like to express my sincere gratitude to the reviewers and members of the jury for their valuable advice and contributions, which greatly enhanced this work. \ \vspace{1\baselineskip}

\noindent Lastly, I would also like to thank my family and friends who supported me during these 4 years.
\newpage\section*{Abstract}
\markboth{Abstract}{}
\addcontentsline{toc}{section}{\protect{Abstract}}

\noindent\textbf{Keywords :} anomaly detection, video, real-time, action recognition, object detection, deep learning, convolutional networks, recurrent networks, neural networks. \par\vspace{2\baselineskip}

\noindent This thesis is part of a CIFRE agreement between the company Othello and the \acrshort{LIASD} laboratory. The objective is to develop an artificial intelligence system that can detect real-time dangers in a video stream. To achieve this, a novel approach combining temporal and spatial analysis has been proposed. \\ 

\noindent Several avenues have been explored to improve anomaly detection by integrating object detection, human pose detection, and motion analysis. For result interpretability, techniques commonly used for image analysis, such as activation and saliency maps, have been extended to videos, and an original method has been proposed.  \\

\noindent The proposed architecture performs binary or multiclass classification depending on whether an alert or the cause needs to be identified. Numerous neural network models have been tested, and three of them have been selected. \acrfull{YOLO} has been used for spatial analysis, a \acrfull{CRNN} composed of \acrshort{VGG}19 and a \acrfull{GRU} for temporal analysis, and a multi-layer perceptron for classification. These models handle different types of data and can be combined in parallel or in series. Although the parallel mode is faster, the serial mode is generally more reliable.  \\ 

\noindent For training these models, supervised learning was chosen, and two proprietary datasets were created. The first dataset focuses on objects that may play a potential role in anomalies, while the second consists of videos containing anomalies or non-anomalies. This approach allows for the processing of both continuous video streams and finite videos, providing greater flexibility in detection.
\csname @openrighttrue\endcsname

\shorttableofcontents{Contents}{0}

\chapter*{Introduction}
\markboth{Introduction}{}
\addstarredchapter{Introduction}

Devices aimed at ensuring our safety (cameras, microphones, drones, etc.) are becoming increasingly widespread. However, we are still not capable of providing effective protection and assistance to citizens. These devices, intended to guarantee our safety, are, in most cases, merely used as deterrents. Currently, remote surveillance allows an operator to monitor multiple locations simultaneously using a camera system placed in public or private spaces. The images obtained from these cameras are transmitted to a series of screens for viewing and analysis, and are then archived or destroyed. This surveillance aims to monitor the safety and security conditions of these locations. Typically, these images are analyzed by human operators. As a result, surveillance is a time-consuming and costly task; furthermore, the effectiveness of such a system depends on the attention and responsiveness of the operator.
In the coming years, more and more locations will be equipped with these tools. This will increase the workload for the operators already in place and raise the amount of labor needed.
With the advancement of artificial intelligence in various fields, such as facial recognition, action recognition, and object detection and tracking, it would be more efficient to at least partially automate this analysis, in order to assist the operator in their task or even replace them in some cases.
This work is a research project carried out as part of a CIFRE agreement between the company Othello\footnote{\href{http://www.othello.group}{http://www.othello.group}} and the \acrfull{LIASD}\footnote{\href{https://www.univ-paris8.fr/UR-Laboratoire-d-Intelligence-Artificielle-et-Semantique-des-Donnees-LIASD}{LIASD Page}}. Its goal is to develop artificial intelligence methods for real-time anomaly detection in video streams and reduce intervention time.
The contributions of my work are as follows:

\begin{itemize} 
\item A problem that builds on the idea of combining temporal analysis of sequences and spatial analysis of images to improve anomaly detection, with the constraint of quickly deciding which alerts to trigger. 

\item Two datasets, one of images and one of videos; these videos represent anomalies that are relevant to our industrial needs, and the images represent objects that have a potential influence on these anomalies. In the context of this CIFRE contract, these datasets are proprietary. 

\item A state-of-the-art review including all aspects potentially useful for this objective: temporal analysis of sequences, spatial analysis of images (object detection, pose estimation, movement detection), anomaly classification, decision traceability. 

\item A set of tests on part of this state-of-the-art, particularly focusing on reference algorithms for spatial and temporal analysis of video streams, also including various video generators \citep{poirier2022detection}. 

\item The proposal of original methods for decision traceability with the contour method \citep{poirier2023visu}, for result evaluation (badBox method), and the contribution to the development of the video-generation program we eventually chose. 

\item An architecture combining image analysis to detect potentially suspicious objects or human poses, time series processing to monitor the actions of these objects, and classification to trigger alerts within a reasonable time \citep{poirier2023enhancing}.

\item The proposal of several modes at the level of spatial and temporal analysis (serial mode, parallel mode), for traceability (tracking mode), and in classification (multi-class and binary modes both by type of anomaly and globally). 

\item Numerous tests of these modes to determine which ones are better suited for specific uses of our program. 

\end{itemize}

\section*{Reading Guide}

This thesis is divided into two parts:

\begin{itemize} 
\item Part \ref{part:problematique} introduces the study's context, presents the problem, and provides a review of works on anomaly detection in video data and object detection based on artificial intelligence techniques. 

\item Part \ref{part:contribution} focuses on the system development, detailing the contributions, experiments carried out, and results obtained. 
\end{itemize}

This thesis concludes with a summary of my work, an overview of its limitations, and suggestions for future perspectives that would be interesting to explore.

\part{Problematic and State of the Art}
\parttoc
\label{part:problematique}
\chapter{Problem Statement}
\chaptermark{Problem Statement}{}
\minitoc
\newpage

\noindent The problem outlined in the general introduction is the detection of events representing anomalies in a video stream. Here, we will begin by describing anomaly detection from a general perspective before specifying the particular conditions of our problem.

\section{Anomaly Detection}

Anomaly detection is a research domain encompassing various themes. In data mining, anomaly detection involves identifying elements, events, or observations that deviate from an expected model. Anomalies are also referred to as outliers, novelties, noise, deviations, and exceptions.
Based on the work of Team Gabru\footnote{Dept. of Computer Science and Engineering, Indian Institute of Technology, Kharagpur, West Bengal, \url{https://github.com/cs60050/TeamGabru}.}, depending on the problems addressed, anomaly detection can fall into one of the following categories:

\begin{itemize}
\item \textbf{Intrusion:} Intrusion detection refers to identifying malicious activities within a network or system.

\item \textbf{Fraud:} Fraud detection involves identifying criminal activities targeting commercial organizations, such as banks, insurance agencies, etc.

\item \textbf{Medical Anomalies:} In the healthcare domain, anomaly detection is generally applied to patient records to identify certain pathologies or abnormal conditions.

\item \textbf{Industrial Damage:} This type of detection aims to identify material damages to prevent potential failures, losses, downtime, or other incidents.

\item \textbf{Textual Anomalies:} In this context, anomalies are represented by unusual sentences or by abnormal topics, articles, or documents.

\item \textbf{Visual Anomalies:} Based on images, this detection can address both static images and those derived from videos.
    \begin{itemize}
    \item For static images, various regions are observed to ensure no areas are abnormal.
    \item For videos, successive frames are analyzed to detect any unusual changes between frames (motion detection).
    \end{itemize}
\end{itemize}

Regardless of the domain, there are three main types of anomalies:

\begin{itemize}
\item \textbf{Point Anomalies:}  
An individual instance can be considered anomalous relative to the rest of the data.

\item \textbf{Contextual Anomalies:}  
A data instance is anomalous in a specific context (but not otherwise), referred to as a contextual anomaly (also known as a conditional anomaly). Each instance is defined using the following two attributes:

    \begin{enumerate}
    \item \textbf{Contextual Attributes:}  
    These attributes are used to determine the context of the instance. For example, in time-series data, time serves as a contextual attribute defining the position of an instance within the entire sequence.

    \item \textbf{Behavioral Attributes:}  
    These attributes define the non-contextual characteristics of an instance. For instance, in a spatial dataset describing global rainfall, the amount of precipitation at a specific location is a behavioral attribute.
    \end{enumerate}
\end{itemize}

\noindent An anomaly can be viewed as an event with a very low probability of occurrence. To address this type of problem, three main methods exist \citep{anomaliePDF}:

\begin{enumerate}
\item Identifying outliers without prior knowledge of the data. This is essentially clustering, an unsupervised method.

\item Modeling both normal and abnormal patterns. This approach involves a binary classification task under supervision, requiring labeled data as either normal or abnormal.

\item Modeling only normality or, in very rare cases, modeling abnormality.
\end{enumerate}

\noindent For a comprehensive overview of anomaly detection, refer to the work of \citet{chandola2009anomaly}, which provides a general introduction to anomaly detection, detailing the main algorithms for each type of data.

\section{Objective}

The anomalies of interest in this thesis are those with a direct impact on the safety and security of individuals captured in the videos. Specifically, we focus on detecting potentially dangerous incidents in visual data, corresponding to contextual anomalies. Such an anomaly can be seen as an irregular pattern (action, behavior) relative to a given situation.
The objective is therefore to classify anomalies based on their severity. For example, minor anomalies include actions that violate established societal rules, such as a cyclist riding on the sidewalk, not wearing a helmet, speeding, or jaywalking. Many of these actions are common, and you might even have committed them yourself. In this same category, we can add some so-called harmful but common anomalies, such as shoplifting, vandalism intended to damage property or locations, or an abandoned bag causing a disruption in public transportation.
On the other hand, there are anomalies that can be classified as major due to their direct impact on the safety of individuals or groups. These include accidents, fires, explosions, terrorist activities, or natural disasters such as cyclones, earthquakes, or tsunamis. Events that can be perceived as anomalies are therefore diverse and varied, with differing consequences.
This can be summarized by defining two types of behaviors: one that does not require intervention (normal case) and another that requires human intervention (abnormal case).
With advances in technology, there are now numerous tools for detecting such anomalies (smartphone-embedded cameras, public cameras, or sharing tools via social networks). The challenge lies in developing a model capable of rapidly detecting these incidents from a video and determining whether they require human intervention. Moreover, since these anomalies can have serious consequences, the ideal solution would be to detect them as early as possible.

\section{Video Analysis}

Due to its complex nature, a video can be analyzed in four different ways:
\begin{itemize}
\item Analyzing only the audio
\item Analyzing each frame independently
\item Analyzing sequences of frames
\item Analyzing both audio and frames
\end{itemize}

\noindent Since most surveillance videos do not include audio, we chose to exclude audio from our problem, leaving us with two possibilities \citep{zhu2020video}:

\begin{itemize}
\item Temporal analysis: Detecting anomalies over time (in a sequence), specifying the start and end frames of the event;
\item Spatial analysis: Detecting anomalies in space (within a frame), identifying the pixels corresponding to the event.
\end{itemize}

\noindent Our problem fundamentally falls under temporal analysis. However, it is important to note that spatial analysis is also relevant. When analyzing actions that could be categorized as anomalies, we observe that some are linked to specific objects. As humans, we can identify a fire in a scene or video by the presence of smoke or flames, recognize a fight by observing the individuals involved, detect a road accident by spotting vehicles, or assess danger when weapons like knives or firearms are visible.\label{combiner}
This leads to the idea explored in this work: combining spatial analysis, specifically object detection, with temporal analysis. Objects help humans interpret situations (e.g., a car accident cannot occur without cars). Such an approach has already been proposed by \citet{doshi2020continual}, combining \acrshort{YOLO} V3 for object detection with Flownet 2 to extract optical flow features. These features are then combined to create a feature vector processed by a \acrshort{KNN}.
In 2023, another approach was proposed by \citet{ali2023real}. Their method, called AVAD (Autoencoder-based Video Anomaly Detection), uses a convolutional autoencoder to detect anomalous frames in a video (temporal anomaly detection) combined with \acrshort{YOLO} V5 to identify the objects responsible for the detected anomaly.
Of course, not all anomalies are tied to a key object, nor are they always apparent. This is often the case in natural disaster contexts, such as cyclones, tsunamis, or earthquakes. In such situations, identifying specific objects, like a tsunami wave or a tornado, can be challenging.
Conversely, in some circumstances, key objects may be easily identifiable but not pose a danger. For example, in airports, it is common to encounter armed military personnel whose presence ensures security.
In these cases, we can often assess the level of risk by observing the surrounding objects and, more importantly, the behavior of people in the video. By analyzing individual behavior, we can determine the context and assess whether the situation constitutes an anomaly. For instance, a person with a threatening posture and another with their hands raised, or a crowd panicking, can confirm the presence of an anomaly. Background details can also help identify the anomaly type—for example, camera shaking could indicate strong winds or an earthquake, while large quantities of visible water might suggest a tsunami or a flooded river.
Both analyses are complementary and use different types of data. Spatial analysis works on static images, while temporal analysis works on videos (image sequences).  
To combine these analyses, two approaches can be considered:
\begin{enumerate}
\item Detecting various objects on-screen and verifying whether they are involved in an anomaly. This involves performing object detection to enrich the data before analyzing the video sequence.
\begin{figure}[H]
\includegraphics[width=\linewidth]{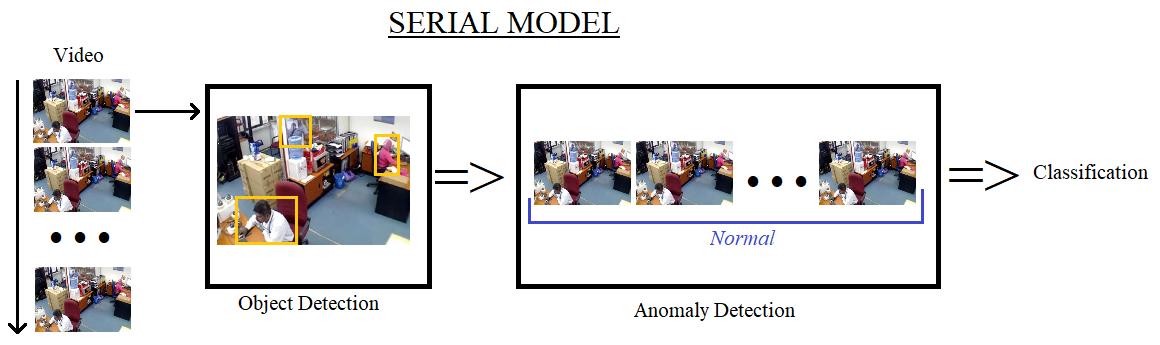}
\end{figure}

\item Analyzing the sequence and the objects present in parallel.
\begin{figure}[H]
\includegraphics[width=\linewidth]{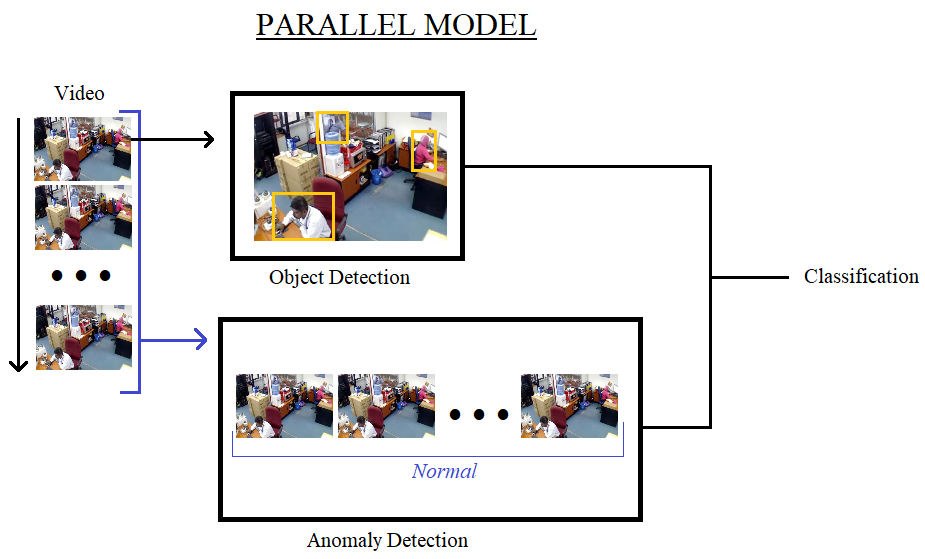}
\end{figure}
\end{enumerate}

\noindent Generally, anomaly detection can be measured in two ways:
\begin{enumerate}
\item Regression: Returning a score for each data point indicating the likelihood of it being an anomaly;
\item Classification: Assigning a normal or abnormal (or more detailed) label to each data point.
\end{enumerate}

\noindent Given that we aim to identify the type of anomaly, we will adopt the second technique, which involves labeling each element as normal or abnormal, or even assigning a specific label to each type of anomaly when possible. Regarding the learning approach, we will use supervised learning techniques because, as shown by the work of Team Gabru\footnote{Dpt of Computer Science and Engineering, Indian Institute of Technology, Kharagpur, West Bengal, \url{https://github.com/cs60050/TeamGabru}.}, they currently outperform unsupervised techniques in video data contexts.

\section{Datasets}

To distinguish between normal and abnormal actions, neural learning algorithms require training on large amounts of data. These data will have a direct impact on the features learned by the model and its decision-making. Therefore, it is necessary to use a dataset that contains a wide range of scenes representing the cases we wish to address.
In this section, we will review the different datasets available that are relevant to our problem to assess how well they meet our requirements. We will start with the summary provided by \citet{zhu2020video}, presented in Table \ref{datasets}, and complement it with others. These datasets are used for anomaly detection in video surveillance contexts.

\begin{table}[H]
\centering
\small
\caption{Datasets mentioned by \citet{zhu2020video}}
\begin{tabular}{c|c|c|c}
Dataset & \# of Videos & Average Frames & Example Anomalies \\ \hline
UCSD Pred 1 & 70 & 201 & Bikers, small carts \\ 
UCSD Pred 2 & 28 & 163 & Bikers, small carts \\ 
Subway Entrance & 1 & 121,749 & Wrong direction, no payment \\ 
Subway Exit & 1 & 64,901 & Wrong direction, no payment \\ 
Avenue & 37 & 839 & Run, throw, new object \\ 
UMN & 5 & 1,290 & Run \\ 
DAD & 1,730 & 100 & Traffic accidents \\ 
CADP & 1,416 & 366 & Traffic accidents \\ 
A3D & 1500 & 85 & Traffic accidents \\ 
DADA & 2000 & 324 & Traffic accidents \\ 
DoTA & 4677 & 156 & Traffic anomalies, e.g., collision \\ 
Iowa DOT & 200 & 27,000 & Traffic accidents \\ 
ShanghaiTech & 437 & 726 & Bikers, cars \\ 
UCF Crime & 1,900 & 7,247 & Arson, accident, burglary, fighting \\ 
Street Scene & 81 & 2509 & Jaywalking, car illegally parked \\ 
\end{tabular}
\label{datasets}
\end{table}

\begin{figure}[H]
\begin{center}
\includegraphics[width=\linewidth]{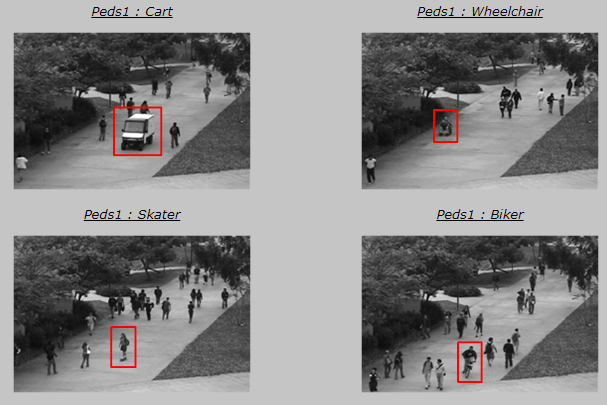}
\caption{Excerpt from the UCSD dataset} 
\label{UCSD}
\end{center}
\end{figure}

Some of these datasets focus solely on traffic issues, such as the Dashcam Accident Dataset (DAD), Car Accident Dataset (CADP), A3D, DOTA Detection of Traffic Anomaly (DADA), which we have excluded. This is because we require a dataset with the widest possible range of anomalies. Among the more diverse datasets, we find UCSD (see Figure \ref{UCSD}) and ShanghaiTech, which address anomalies that we have classified as minor, such as bikes on pedestrian-only paths. These were also excluded from our study. Of greater interest is the UCF Crime dataset.

UCF Crime is currently the benchmark dataset for anomaly detection in videos. It was created by \citet{sultani2018real} at the University of Florida and contains 1,900 untrimmed videos representing realistic anomalies across 13 categories: abuse, arrests, arson, assault, road accidents, burglary, explosions, fighting, robbery, shootings, shoplifting, stealing, and vandalism, as well as 950 normal videos. The training set consists of 800 normal and 810 abnormal videos. All videos are temporally annotated with timestamps indicating the start and end of each anomaly present in a video. These annotations are provided as a CSV file accompanying the dataset. Moreover, some videos contain multiple anomalies, either of the same type or different types. Additionally, UCF Crime encompasses various lighting conditions, image resolutions, and camera angles.
Unfortunately, while the training set has a balanced number of videos (normal/abnormal), this balance does not extend to the incidents or their durations. Table \ref{ucfvideo} illustrates the distribution of videos and anomalies for each category.

\begin{table}[H]
	\centering
	\small
	\caption{Category distribution of videos in the UCF Crime dataset}
	\begin{tabular}{l|ll|ll|ll}
		\hline
		 & \multicolumn{2}{c|}{Training} & \multicolumn{2}{c|}{Testing} & \multicolumn{2}{c}{Total} \\ 
		Class & Videos & Sequences\textsuperscript{1} & Videos & Sequences & Videos & Sequences \\ 
		\hline
		\textbf{Normal} & \textbf{800} & \textbf{800} & \textbf{150} & \textbf{150} & \textbf{950} & \textbf{950} \\ 
		Abuse & 48 & 56 & 2 & 2 & 50 & 58 \\ 
		Arrest & 45 & 41 & 5 & 5 & 50 & 46 \\ 
		Arson & 41 & 42 & 9 & 10 & 50 & 52 \\ 
		Assault & 47 & 48 & 3 & 4 & 50 & 52 \\ 
		Burglary & 87 & 87 & 13 & 15 & 100 & 102 \\ 
		Explosion & 29 & 32 & 21 & 22 & 50 & 54 \\ 
		Fighting & 45 & 67 & 5 & 5 & 50 & 72 \\ 
		Road Accident & 127 & 134 & 23 & 23 & 150 & 157 \\ 
		Robbery & 145 & 165 & 5 & 5 & 150 & 170 \\ 
		Shooting & 27 & 33 & 23 & 25 & 50 & 58 \\ 
		Shoplifting & 29 & 39 & 21 & 25 & 50 & 58 \\ 
		Stealing & 95 & 137 & 5 & 7 & 100 & 144 \\ 
		Vandalism & 45 & 59 & 5 & 8 & 50 & 67 \\ 
		\hline
		\textbf{Total anomalies} & \textbf{810} & \textbf{940} & \textbf{140} & \textbf{156} & \textbf{950} & \textbf{1096} \\ 
		\hline
		\textbf{\textcolor{red}{Total}} & \textbf{\textcolor{red}{1610}} & \textbf{\textcolor{red}{1740}} & \textbf{\textcolor{red}{290}} & \textbf{\textcolor{red}{306}} & \textbf{\textcolor{red}{1900}} & \textbf{\textcolor{red}{2046}} \\ 
		\hline
	\end{tabular}
	\vspace{0.5em} 
	\begin{minipage}{0.9\linewidth}
	\textsuperscript{1} Usable sequences representing either a normal case or an anomaly.
	\end{minipage}
	\label{ucfvideo}
\end{table}

Depending on the class, sequences representing an incident vary greatly, as shown in Tables \ref{ucfseqtrain} (for training data) and \ref{ucfseqtest} (for test data), which indicate the duration of episodes representing an anomaly. A gunshot typically lasts only a few seconds, unlike fighting or theft. The longest videos, however, are in the normal class, with durations reaching several minutes or even exceeding an hour. Finally, some abnormal videos do not contain any major anomalies, as is the case for certain videos in the training set of the Arrest or Burglary classes.

\begin{table}[H]
\centering
\caption{Duration of training anomalies in UCF Crime}
\small
\begin{tabular}{c | c | c | c}
Classes & Minimum Duration & Average Duration & Maximum Duration \\ \hline
\textbf{Normal} & 6.7s & 394.86s (6min 34) & 32,550s (9h20) \\ 
Abuse & 2.88s & 54.49s & 110.2s (1min 50) \\ 
Arrest & 6s & 64.30s (1min 4) & 255.88s (4min 15) \\ 
Arson & 4.24s & 32s & 180s (3min) \\ 
Assault & 1.92s & 27.76s & 278s (4min 38) \\ 
Burglary & 1.68s & 59.37s & 321.2s (5min 20) \\ 
Explosion & 5.2s & 15.47s & 30s \\ 
Fighting & 3.6s & 22.72s & 62s \\ 
Road Accident & 2.16s & 8.35s & 32s \\ 
Robbery & 2.52s & 36.15s & 104.76s (1min 44) \\ 
Shooting & 1.6s & 4.93s & 16.8s \\ 
Shoplifting & 3.6s & 15.5s & 70.6s (1min 10) \\ 
Stealing & 4.8s & 31.78s & 118.4s (1min 58) \\ 
Vandalism & 3.6s & 24.71s & 81.6s (1min 21) \\ 
\end{tabular}
\label{ucfseqtrain}
\end{table}

\begin{table}[H]
\centering
\caption{Duration of testing anomalies in UCF Crime}
\small
\begin{tabular}{c | c | c | c}
Classes & Minimum Duration & Average Duration & Maximum Duration \\ \hline
\textbf{Normal} & 9.96s & 144.20s (2min 24) & 3,599.9s (59min 59) \\ 
Abuse & 3s & 3.2s & 3.4s \\ 
Arrest & 12s & 62.4s (1min 2) & 124.8s (2min 4) \\ 
Arson & 3.8s & 32.63s & 138s (2min 18) \\ 
Assault & 14s & 85.05s (1min 25) & 276s (4min 36) \\ 
Burglary & 5.6s & 44.54s & 118.4s (1min 58) \\ 
Explosion & 3s & 13.86s & 64.24s (1min 4) \\ 
Fighting & 10.8s & 35.46s & 65.2s (1min 5) \\ 
Road Accident & 1.2s & 3.94s & 7.6s \\ 
Robbery & 4.8s & 30.48s & 72.6s (1min 12) \\ 
Shooting & 3.2s & 16.14s & 69.6s (1min 9) \\ 
Shoplifting & 2.4s & 12.66s & 108s (1min 48) \\ 
Stealing & 2.4s & 34.29s & 123.6s (2min 30) \\ 
Vandalism & 2.4s & 11.35s & 27.6s \\ 
\end{tabular}
\label{ucfseqtest}
\end{table}

\noindent In addition to the previously mentioned datasets, Movie Fight\footnote{\url{https://www.kaggle.com/datasets/naveenk903/movies-fight-detection-dataset}}, which contains 200 movie clips divided into two categories (Fight and Non-fight), and Hockey Fight\footnote{\url{https://www.kaggle.com/datasets/yassershrief/hockey-fight-vidoes}}, which contains 1,000 clips of fights occurring during professional hockey games in North America between the 2009–2010 and 2018–2019 NHL seasons, are also worth noting.
Upon examining these datasets, we observed that Movie Fight does not provide enough videos to effectively train a robust model, while Hockey Fight lacks diversity in its scenes. The clips all occur in similar environments, with identical camera angles and image quality, potentially limiting a model's ability to adapt to different scenarios.
As for UCF Crime, although it is currently the benchmark dataset, many studies using it report poor results and recommend creating custom datasets.
For instance, \citet{sernani2021deep} developed a public dataset called AIRTLab to test the robustness of violence detection techniques, particularly in handling false positives. It consists of 350 videos split into two categories violent and non violent including scenarios such as hugging, clapping, teasing, etc.
\citet{vrskova2022new} point out in their article that the UCF dataset suffers from insufficient, poorly cleaned, and unbalanced data. Unlike normal videos, which can last several minutes, abnormal videos (anomalies) are only a few seconds long. To address this issue, they created the Abnormal Activities dataset, comprising 1,069 videos divided into 11 classes, each containing approximately 100 videos. These classes include: knife threats, theft, fighting, pollution (videos showing people littering the environment), harassment, vandalism, kidnapping, terrorism, drunkenness, begging (videos featuring people bothering or harassing others by asking for money), and normal videos. These scenes were recorded under different lighting conditions. During the creation of this dataset, the aim was to simulate the placement of security cameras in public spaces such as buildings, parking lots, and natural settings.\label{notrejeu}
From this exploration, it became clear that no existing dataset met the diversity, representativeness, and volume criteria we sought. However, some of these datasets may complement our dataset for testing purposes.
To build a dataset addressing our requirements, several solutions are possible:
\begin{itemize}
\item Collecting New Data

Currently, the steps of collecting, cleaning, and classifying data must be performed manually, as no sufficiently advanced technology automates these processes. Furthermore, certain anomalies occur less frequently, making them harder to find. Due to their severity, some anomalies are censored. Additionally, multiple actions may occur in a single video, with some considered anomalous and others normal. As a result, each video contains either a normal sequence or one or more anomalies interspersed with normal sequences. Consequently, a large volume of videos does not necessarily translate to a large quantity of exploitable scenes.

\item Using Data Augmentation

Since videos are composed of image sequences, another solution is to apply data augmentation to each frame to artificially increase the dataset's size. This includes transformations such as brightness changes, mirroring, blurring, zooming, etc., as well as more advanced techniques that sometimes require machine learning models, like degrading or enhancing the video’s resolution or smoothness. However, in the context of anomaly detection, data augmentation poses challenges. Given the small and raw nature of the datasets, artificial transformations may alter the data's quality and relevance, potentially compromising the analysis results.

\item Generating Synthetic Data

A final option is the use of generative networks to produce synthetic data. This technique was employed by \citet{jacob2019anomaly}, who also criticized the lack of datasets with sufficient volume or quality to properly train deep learning models and proposed this solution in his thesis.
\end{itemize}

\noindent Due to budgetary constraints and limited computational resources, we decided not to explore synthetic data generation to address the dataset quantity issue. Synthetic data creation requires substantial computational power, particularly for training generative networks, which is resource-intensive. Consequently, we opted for other approaches, such as collecting new data and using realistic data augmentation techniques to increase the volume and diversity of our dataset.
Despite these efforts, action recognition remains a challenging task due to cluttered backgrounds, occlusions, lighting variations, and changes in perspective. For a model to accurately discern an action, the dataset must contain videos that are both diverse and varied. The vast redundancy of information complicates the extraction of discriminative features. According to \citet{arif20193d}, ``The success of action recognition problems depends on an appropriate feature extraction process. This extraction is critical for distinguishing variations in a scene.`` Therefore, each of our videos must be meticulously cleaned to enable the model to extract relevant features and subsequently generalize its approach effectively.

\section{Requirements}
Our model should not rely on binary classification but must instead be capable of identifying the specific type of anomaly. This approach will facilitate determining the appropriate individuals to intervene in case of a problem.
However, this choice requires selecting a representative list of anomalies for training. The model must also have high adaptability to detect anomalies that are not part of this list. It should be able to differentiate between anomalies that could have severe consequences for the individuals involved and those with less significant impact. This prioritization will help ensure safety by optimizing intervention efforts.
One unresolved question is whether it is better to issue frequent alerts, even at the risk of errors, or to alert only when certain, possibly missing some cases.
Additionally, it would be valuable to localize the anomaly within the data, making the model more transparent and enabling a human reviewer to verify the detection's accuracy. However, how can we teach the model to focus on what we expect whether on the foreground or the background ? Since the data processed is vast and complex, significant computational power is required. Moreover, given the rapid succession of images in videos, the redundancy of information should help the model determine relevant features.
A specific concern relates to the interpretability of results. Machine learning algorithms, particularly neural networks, are widely used to address various challenges. Despite their effectiveness, they are often seen as highly complex and opaque structures (black boxes) that provide solutions without explaining the reasoning behind their decisions. This lack of transparency makes them difficult to debug or improve when performance goals are not met. It also raises ethical and legal concerns. How can we trust AI systems if we cannot explain their choices ? How can we communicate these findings to non-experts if we cannot understand them ourselves ? \\

The introduction of the General Data Protection Regulation (GDPR) in May 2019, particularly Article 22-1, which states that decisions cannot be based solely on automated processing, and the human need for explicit and transparent information for decision-making, have significantly accelerated research in this area. For these reasons, interpretability and explainability have become critical issues in AI research. As a result, numerous studies are conducted annually, especially on convolutional networks for image analysis (\cite{olah2018the,olah2017feature,bau2020units,zhang2018interpretable}).
According to the French CNIL, explainability is defined as “the ability to link and make understandable the elements considered by the AI system in producing a result.” This aspect of machine learning aims to help us understand the decision-making mechanisms of a model.
Thus, as an additional requirement, we include the necessity to visualize the area where the anomaly occurs. This practice is already employed for anomaly detection in images, and we aim to integrate it into video-based anomaly detection.
Lastly, in this thesis, we aim to perform anomaly detection rather than prediction. Our analysis is conducted retrospectively, once the action has been completed. As a result, the latency time depends on the length of the sequence to be analyzed. Given the potential consequences of certain anomalies, our objective is to provide analysis in the shortest possible time, even if it is not strictly real-time.
As this thesis is a CIFRE project, it is essential to mention the constraints imposed by the company. The primary constraints are as follows:

\begin{enumerate}

\item Potential alerts must not be missed, even if this leads to an excess of alerts;

\item The final model must be compatible with both CPUs and GPUs;

\item The types of anomalies studied must align with categories recognized by the company (e.g., fights, gunshots, etc.).
\end{enumerate}

\section{Conclusion}

Our main challenges are as follows:  
\begin{itemize}  
\item Building a proprietary dataset that is representative of the various types of anomalies studied.
The core issue that initiated this challenge lies in the lack of adequate data for our project, as highlighted in the earlier section on datasets. Existing datasets, such as Movie Fight, Hockey Fight, and UCF CRIME, have exposed limitations in terms of volume and diversity, hindering our ability to develop a performant anomaly detection model. Addressing this significant gap, the need for a dataset that comprehensively covers the variety of targeted anomalies becomes imperative.

\item Labeling the data to create a ground truth clearly delineating the boundary between normal and abnormal behaviors.  
The necessity to provide a reliable and consistent reference for training our model led to this critical challenge. Since the model’s effectiveness depends on its ability to distinguish normal behaviors from anomalies, the precision of this ground truth is of utmost importance for accurate and coherent anomaly identification across diverse scenarios.

\item Segmenting labeled videos to retain only representative events.  
This challenge stems from the need to focus the model's attention on the most meaningful events. The large volume of video data creates a dilemma between retaining relevant information and eliminating redundant content. Storage and computational constraints play a significant role in the decision to segment videos, aiming to preserve key moments while avoiding an overload of irrelevant information.

\item Increasing the quantity of available data and balancing it when necessary.  
Training an effective anomaly detection model, particularly with deep learning techniques, requires a substantial amount of data. However, the available data is often insufficient and imbalanced in terms of anomaly frequency. To achieve desired performance levels, increasing data volume and balancing it to ensure adequate representativeness are essential.

\item Cleaning the data to filter out noise.
The challenge of data quality led to this task. Noisy or incorrectly recorded data can impair the model’s ability to discern meaningful patterns. However, removing noise is not without challenges. Balancing the preservation of relevant information with the elimination of unwanted elements is complex, as the quality of cleaned data is critical to ensuring the performance and reliability of the anomaly detection model.

\item Designing a model capable of performing near real-time analysis by combining spatial and temporal analysis.  
Driven by the necessity for rapid intervention in critical situations, developing a near real-time analysis model became essential. The design of this model was inspired by the aim to emulate human-like behavior, fusing spatial and temporal analysis to capture the way humans perceive their environment, focusing on salient elements and observing their evolution over time.

\item Visualizing the features used in decision-making.  
New regulations, such as those related to the GDPR, have prompted the need to make model decisions transparent and comprehensible. The challenge of visualizing the features selected by the model arose from the desire to comply with ethical and legal standards while ensuring trust in automated decisions.
\end{itemize}  

\noindent In the following sections, we will present a state of the art on anomaly detection (temporal analysis) and object detection (spatial analysis). We will then explain our contributions and responses to these challenges.

\label{part:etatLart}
\chapter{State of the Art}
\chaptermark{State of the Art}{}
\minitoc
\newpage

\section{Anomaly Detection in Videos}

The field of computer vision encompasses numerous techniques for analyzing visual data that can be used for anomaly detection, such as calculating the trajectory of objects on screen, analyzing color histograms, or performing optical flow. These technologies need to be manually calibrated and generally require adjustments depending on the scene being processed. This is the issue that deep learning technologies aim to solve by automatically extracting relevant features from data to generalize their analysis. Thus, we focused on this type of technology. For more information on standard analysis techniques, we refer to \citet{jacob2019anomaly}.

\subsection{Convolutional RNN}

\begin{figure}[H]
\includegraphics[width=\linewidth]{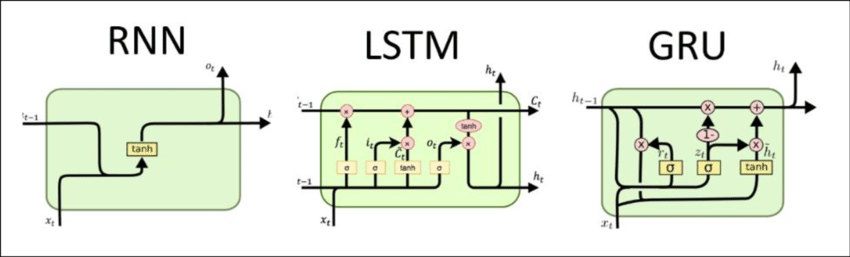}
\caption{Recurrent Architectures \citep{toharudin2020employing}}
\label{RNN architecture}
\end{figure}

Recurrent networks (\acrfull{RNN}) are considered the most suitable neural models for processing time series. They were developed by \citet{elman1990finding} and \citet{jordan1997serial}, who each proposed their own version. Unlike feed-forward networks, they include feedback loops between units, which allow them to remember dynamic time series. The most well-known of them, \acrfull{LSTM}, was proposed by \citet{hochreiter1997long} and later improved by \citet{gers2000learning} with the introduction of a forget gate. The advantage of this type of network is that certain information related to previously viewed data is retained and used in decision-making. Through this mechanism, \acrshort{LSTM} networks are considered references in time series analysis.
\acrshort{LSTM} is not the only \acrshort{RNN} that has proven effective in this context. Another model, \acrfull{GRU}, has a simpler architecture but is equally effective. While an \acrshort{LSTM} consists of a three-gate architecture, with a forget gate, an input gate, and an output gate, the \acrshort{GRU} only has two: a reset gate and an update gate. Figure \ref{RNN architecture} details these three architectures.

\begin{figure}[H]
    \includegraphics[width=\linewidth]{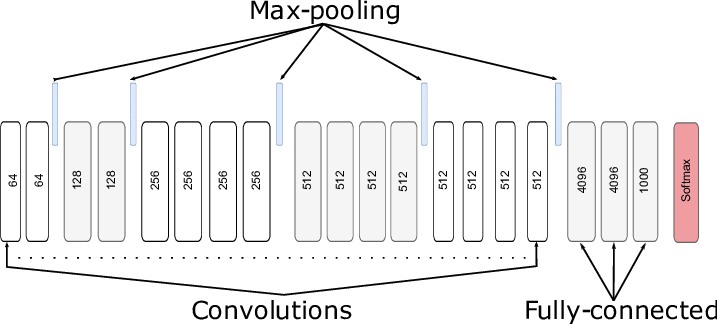}
    \caption{\acrshort{VGG}19 Architecture (\cite{lagunas2018transfer})}
    \label{vgg19}
\end{figure}

\noindent While \acrshort{RNN}s can effectively handle the temporal aspect of video data, they are not suited for processing the images that make up these videos. This is because images are complex data with crucial information encoded in pixel positions, and \acrshort{RNN}s are not designed to consider this spatial structure, making them inappropriate for directly handling images.
For image processing, the common approach is to use convolutional networks, or \acrfull{CNN}s. These networks, introduced by \citet{lecun1989backpropagation}, can extract spatial features from images through operations called convolutions. Among these architectures is \acrshort{VGG}19, a model proposed by Karen Simonyan and Andrew Zisserman (\cite{simonyan2014very}) composed of 23 layers grouped into five main blocks, each comprising 2 to 4 convolutional layers followed by a pooling layer (Figure \ref{vgg19}).

\noindent Other architectures are also used in image processing. For example, ResNet (\cite{he2016deep}) is a deep network based on the idea of ``shortcuts.`` The addition of residual connections helps prevent performance degradation due to additional layers. Inception (\cite{szegedy2015going}) uses inception blocks to extract features at different spatial scales from the same image. InceptionResNet (\cite{szegedy2016inception}) combines the benefits of ResNet and Inception. Xception (\cite{chollet2017xception}) is an improved version of Inception that uses depthwise separable convolutions by applying convolution on each channel of the input followed by spatial convolution. This method significantly reduces computational costs while improving model accuracy. EfficientNet (\cite{tan2019efficientnet}), another example, is a family of convolutional network architectures that optimize the balance between depth, width, and image resolution to maximize accuracy while minimizing computational costs.
To process video data, these networks have been combined with recurrent networks such as \acrshort{LSTM}, resulting in two types of architectures, differing in the level of integration of the two types of models: Convolutional \acrshort{LSTM} (\cite{shi2015convolutional}) (often abbreviated as \acrshort{CNN} + \acrshort{LSTM}) and ConvLSTM. In the Convolutional \acrshort{LSTM} architecture, each image passes through different convolutional layers to produce a feature vector, which is then processed by the \acrshort{LSTM}. The ConvLSTM architecture is an \acrshort{LSTM} network where the internal matrix multiplications have been replaced by convolutions.
Many works use this type of model for video analysis, whether for action recognition (\cite{shi2015convolutional}), anomaly detection (\cite{majd2019motion}), or violent action detection (\cite{vrskova2022new, vrskova2020violent, de2021temporal, jahlan2021mobile}).
Although less commonly mentioned in the literature, Convolutional \acrshort{GRU} models are generally more effective than Convolutional \acrshort{LSTM} and are also less computationally intensive (\cite{ravi2021exploring}).

\subsection{3D Convolution}

\begin{figure}[H]
\includegraphics[scale=0.6]{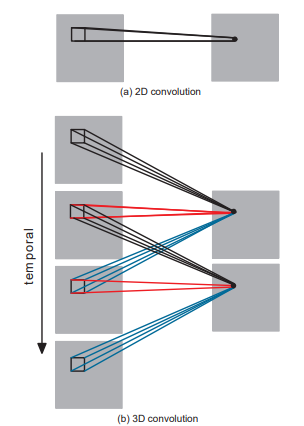}
\includegraphics[scale=0.6]{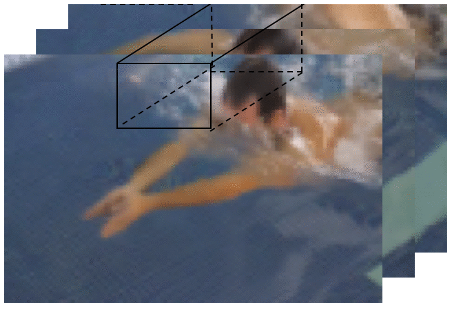}
\caption{Operation of convolution filters, \citet{ji20123d, actionDetect}}
\label{c3d}
\end{figure}

\noindent \citet{ji20123d} developed a model based on 3D convolutions (\acrshort{C3D}) for action recognition in video streams. The general idea is to observe each image while taking into account the preceding and following frames to capture the various movements made by objects on screen.
Unlike Convolutional \acrshort{LSTM} / \acrshort{GRU} approaches, which reduce each image to a 1D feature vector, 3D convolutions retain information related to depth and, in the case of videos, inter-frame changes. To achieve this, the kernels in each layer move in three directions (x, y, z) to create a 3D activation map (Figure \ref{c3d}).
Another difference is that instead of extracting information from each frame individually, as a 2D convolution would, and leaving it to a recurrent network to link them over time, \acrshort{C3D} processes packets of successive frames to analyze differences among them and thus capture motion-related information. This technique has shown excellent results in several fields, such as human pose estimation in images or videos, action recognition (\cite{ji20123d, vrskova2022human, tran2015learning}), and medical image segmentation.
Due to their effectiveness in processing videos, 3D convolutions have also been used for anomaly detection tasks \citep{mishra2021hybrid}. However, this performance comes with a significant computational load.

\subsection{Convolutional Autoencoder}

\begin{figure}[H]
\includegraphics[scale=0.5]{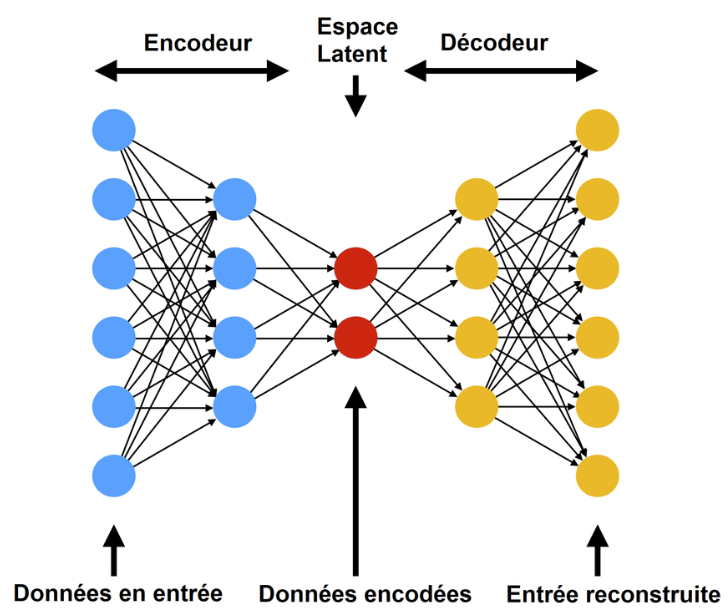}
\caption{Architecture of an autoencoder, \citet{sublime2022apprentissage}}
\label{autoencodeur}
\end{figure}

\begin{figure}[H]
\includegraphics[width=\linewidth]{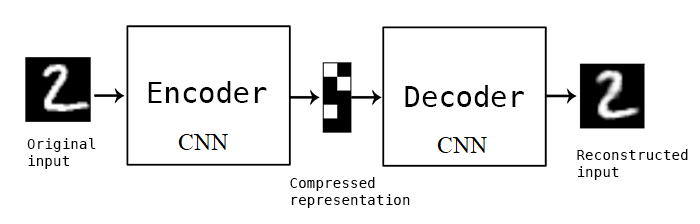}
\caption{Image autoencoder, \citet{chollet2015keras}}
\label{autoencodeurFonctionnement}
\end{figure}

Anomaly detection in video surveillance is challenging due to the variety of anomaly types (see Chapter \ref{part:problematique}), which limits the use of supervised techniques. In such cases, unsupervised approaches that do not require labeled data are preferable. Among these approaches, autoencoder models stand out.
The architecture of an autoencoder consists of two parts: the encoder and the decoder (Figure \ref{autoencodeur}). Its operation is fairly straightforward: input data are first encoded and then passed through the decoder to be reconstructed (decoded) (Figure \ref{autoencodeurFonctionnement}). The model’s performance can be measured by comparing the initial data with the reproduced data. In the context of anomaly detection, any abnormal data provided to the model will be difficult to reconstruct, making it relatively easy to identify.

\noindent For image processing, the encoder and decoder are typically \acrshort{CNN}s, which are widely used for visual anomaly detection \citep{jacob2019anomaly, ribeiro2018study, an2015variational, chen2018autoencoder, hasan2016learning}.
The compressed representation (the output of the encoder and input to the decoder) usually has a 1D structure, which may lead to a loss of potentially useful information for reconstructing the action in the video.
Autoencoders have many variants. In the context of video data processing, one variant involves replacing the \acrshort{CNN}s with convolutional \acrshort{LSTM}s to combine the benefits of \acrshort{RNN}s with autoencoders, allowing input images to be reconstructed based on previously remembered frames (\cite{rumelhart1986learning, medel2016anomaly, chong2017abnormal, deepak2021residual, wang2018abnormal, luo2017remembering}).
Despite their effectiveness in the decoding phase, autoencoders reconstruct each frame of the video, which is time- and compute-intensive, limiting their use in real-time applications where responsiveness is essential.

\subsection{Temporal Convolutional Network (TCN)}

\begin{figure}[H]
\includegraphics[width=\linewidth, height=8cm]{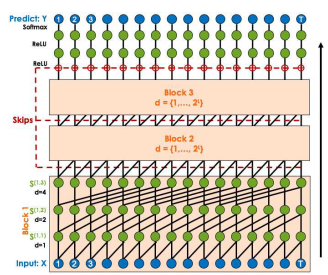}
\caption{Architecture of \acrshort{TCN} \citep{lea2017temporal}}
\label{archiTCN}
\end{figure}

Temporal Convolutional Networks (\acrshort{TCN}) are a family of convolutional neural network models designed for time-series modeling (Figure \ref{archiTCN}). Unlike traditional \acrshort{RNN}s, which use recurrent connections to capture temporal dependencies, \acrshort{TCN}s use temporal convolutions to capture patterns within sequences.

\begin{figure}[H]
\includegraphics[width=\linewidth, height=8cm]{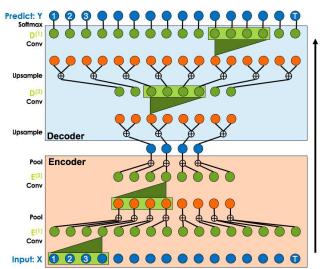}
\caption{Architecture of the TCN-based autoencoder \citep{lea2017temporal}}
\label{autoencodeurTCN}
\end{figure}

\noindent Dilated filters play a crucial role in the ability of \acrshort{TCN}s to capture temporal patterns at different scales. Dilated filters are convolutions where the indices of the values to be considered are spaced by a specific number of steps, known as the dilation rate. This enables \acrshort{TCN}s to capture patterns at various temporal scales while maintaining reasonable computational complexity. Furthermore, dilated filters can be causal or non-causal. Causal dilated filters are useful for recognition tasks where it is important that the model cannot see into the future. Non-causal dilated filters, on the other hand, can be used for classification tasks where the entire sequence is available from the start.
\citet{lea2017temporal} introduced an encoder-decoder architecture using \acrshort{TCN}s for action segmentation and detection in videos. Figure \ref{autoencodeurTCN} shows how temporal convolutions are used to encode input sequences; then, temporal deconvolutions are used to reconstruct the output sequences. Experimental results suggest that \acrshort{TCN}s are capable of capturing complex temporal patterns, such as the movements comprising an action, while being resistant to temporal variations. This means they can recognize actions even if they do not occur at the exact same time or have different durations in each occurrence.

\subsection{Transformer}

\begin{figure}[H]
\includegraphics[width=\linewidth]{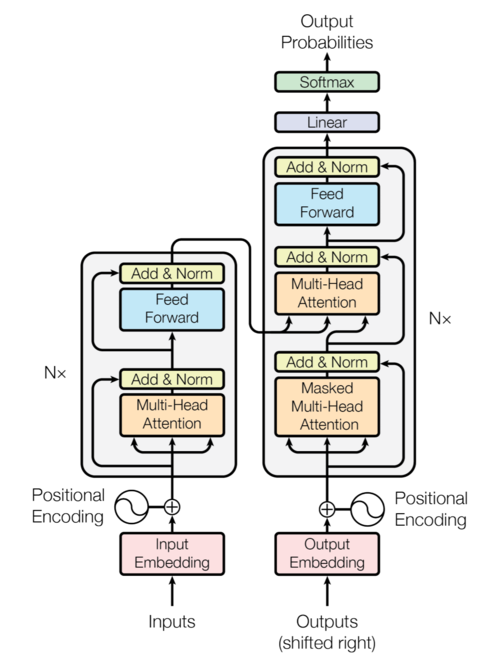}
\caption{Transformer architecture, \citet{vaswani2017attention}}
\label{transformer}
\end{figure}

Transformers are models introduced by \citet{vaswani2017attention} for textual analysis. These models can be viewed as ``sequence-to-sequence`` models, meaning they take a data sequence as input and produce another sequence of the same type as output. Like \acrshort{RNN}s, they are capable of processing sequential data and are commonly used for translation tasks.
Their architecture (Figure \ref{transformer}) is inspired by autoencoders and consists of a stack of encoders and decoders (the same number on each side). Each piece of data passes through the various encoders. Once encoded, it is sent to the decoders and goes back up through the different layers. Each decoder input contains both information from the previous decoder and that from the last encoder. A unique feature is that each network block, including each encoder and decoder, has an attention mechanism, which assesses the degree of association or dissociation between different pieces of information.

\begin{figure}[H]
\includegraphics[width=\linewidth]{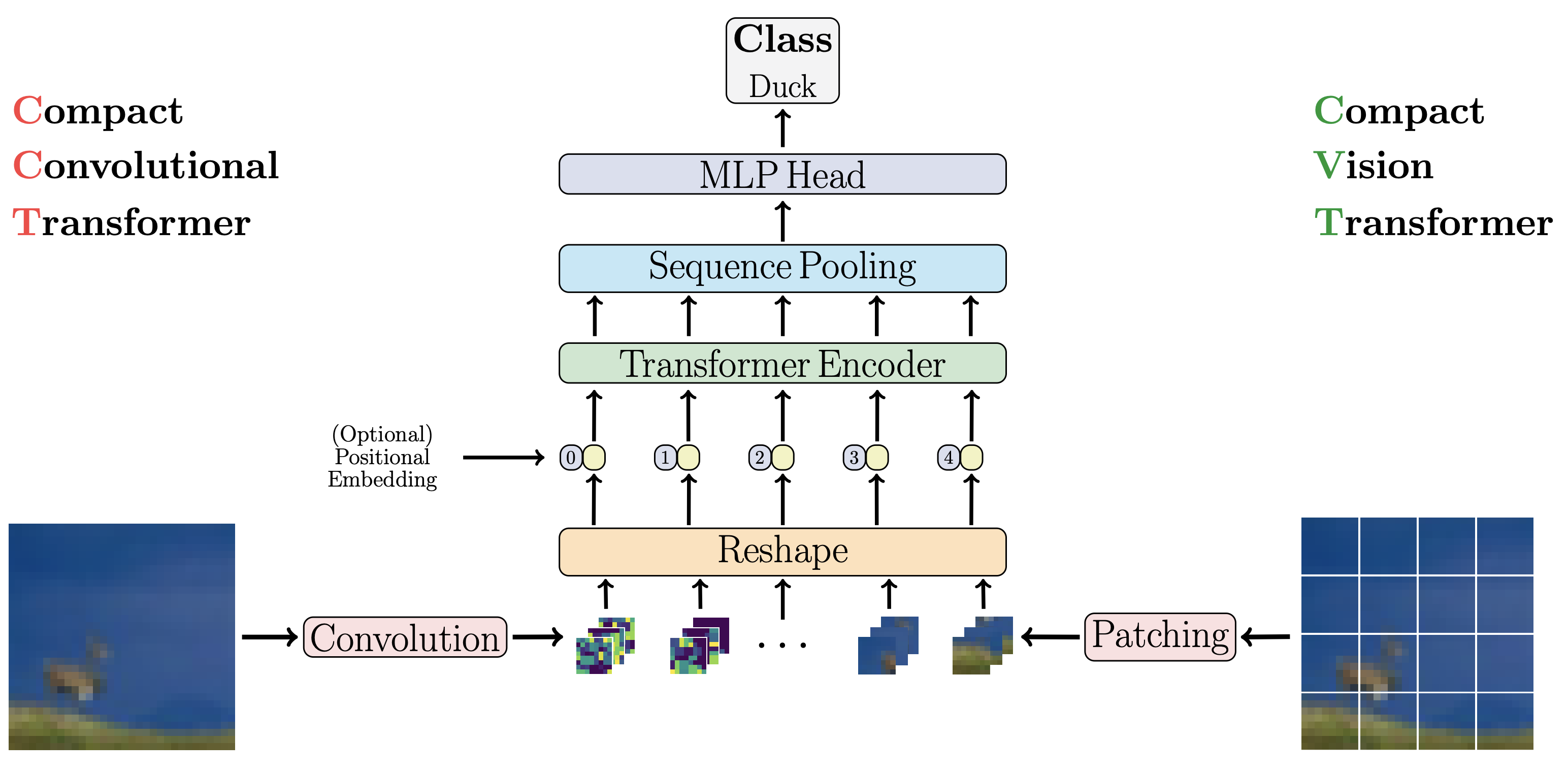}\\
\includegraphics[width=\linewidth]{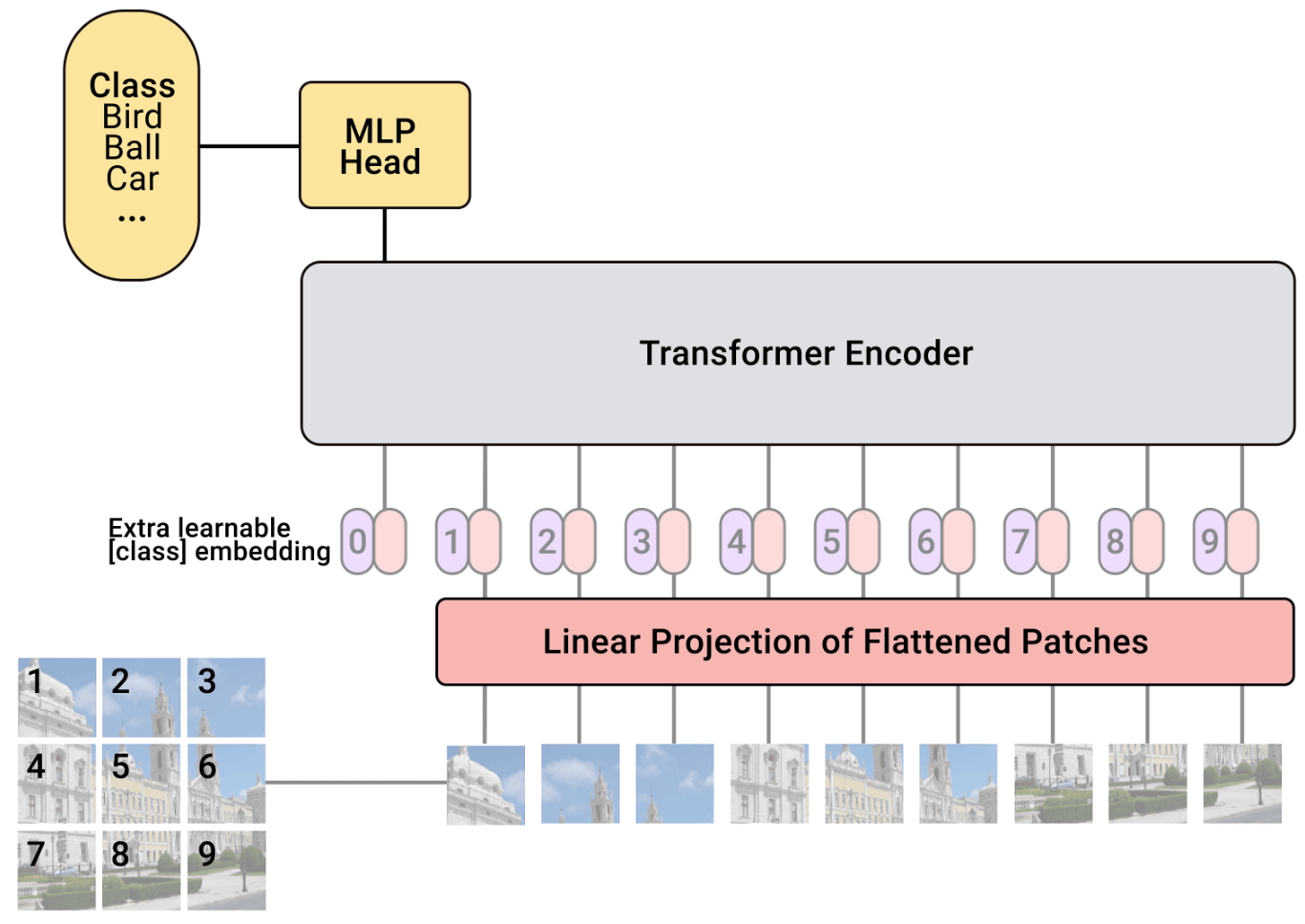}
\caption{Vision Transformer, \citet{vitGit, hassani2104escaping}}
\label{transformer1}
\end{figure}

\begin{figure}[H]
\includegraphics[width=\linewidth]{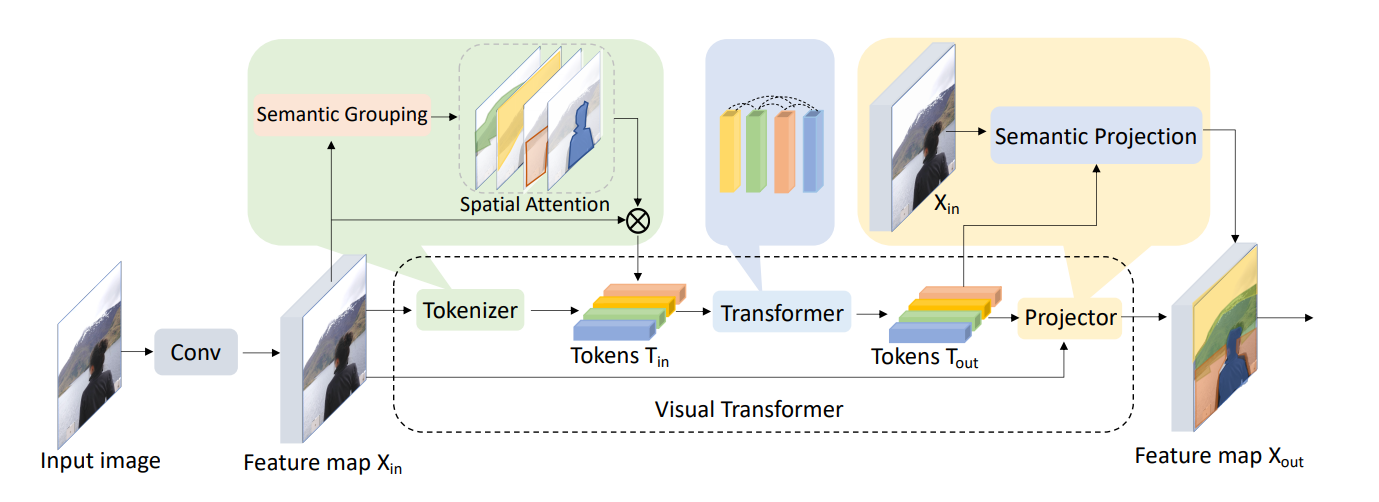}
\caption{Vision Transformer, \citet{wu2020visual}}
\label{transformer2}
\end{figure}

\begin{figure}[H]
\includegraphics[width=\linewidth]{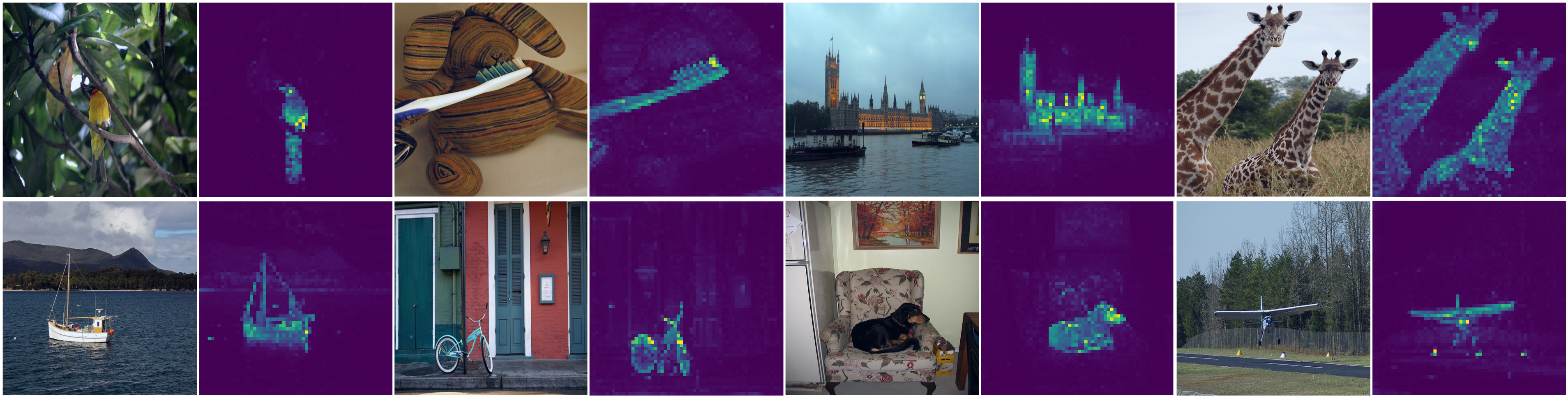}
\caption{Attention map generated by DINO, \citet{caron2021emerging}}
\label{attention_map}
\end{figure}

\noindent More recently, in 2020-2021, Google teams proposed an adaptation of transformer networks for processing images or video data (\cite{dosovitskiy2020image, arnab2021vivit}) (Figures \ref{transformer1}, \ref{transformer2}). Thanks to the attention mechanism, these networks are effective in tasks like object detection and image segmentation. These models are called \acrfull{ViT}.\label{DINO}
In 2021, researchers at Facebook (\cite{caron2021emerging}) proposed a model named \gls{DINO} based on similar technology. This model uses \acrshort{ViT} with self-supervised learning, allowing it to autonomously identify important areas in an image or video. These areas can then be visualized using attention maps (Figure \ref{attention_map}).

\noindent For more information on the various types of vision transformers available today, refer to the study by \citet{vitGit}.

\subsection{Generative models}

\begin{figure}[H]
\includegraphics[width=\linewidth]{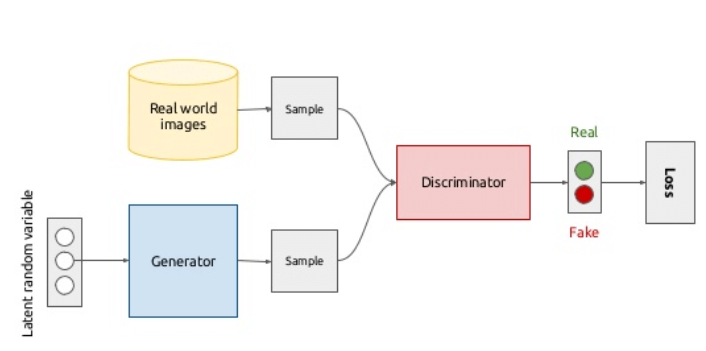}
\caption{Vanilla GAN Architecture, \citet{articleGAN}}
\label{gan}
\end{figure}

For anomaly detection in video data, some research leverages generative models such as \acrfull{GAN} (\cite{lin2021learning, dimokranitou2017adversarial, schlegl2017unsupervised}). Today, these models are widely used to generate synthetic data (image, video) or to improve the resolution of such data.
\acrshort{GAN} was proposed by \citet{goodfellow2014generative}. It consists of two sub-models: a generative model and a discriminative model (Figure \ref{gan}). The generator's task is to create natural-looking data similar to the original data, while the discriminator's task is to determine if the data appears natural or has been artificially generated. As training progresses, both models improve through what is called adversarial optimization until the discriminator can no longer distinguish between synthetic (fake) and authentic data. The idea is to train a model on video data so that it can generate synthetic videos and then use the discriminator as a classifier to identify abnormal data specifically, any video containing an anomalous action.

\noindent\citet{zhu2020video} provide a comprehensive discussion on supervised and unsupervised deep learning methods for anomaly detection in surveillance videos. Additionally, \citet{actionDetect} explore action recognition in video data and mention certain techniques or models that were not covered in our study.

\section{Object Detection}

\begin{figure}[H]
\centering
\includegraphics[]{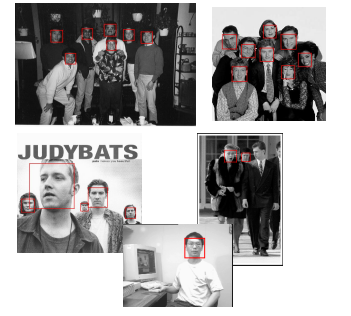}
\caption{Boost Cascade \citep{viola2001rapid}}
\label{boost}
\end{figure}

Object detection was first introduced by \citet{viola2001rapid}. Their initial goal was to detect faces, and to achieve this, they developed a machine learning model known as ``boost cascade`` (Figure \ref{boost}).

\noindent Later, other technologies emerged, such as \gls{HOG} and \gls{DPM}, all relying on manually selected feature extraction techniques, including edges, corners, and gradients in images, coupled with non-neural machine learning algorithms.
It was not until September 2012 that the field took a major leap forward with the work of \citet{krizhevsky2012imagenet}, who won the ImageNet LSVRC-2012 competition with their deep convolutional network, AlexNet, capable of recognizing a large number of objects with an error rate of 15.3\%. In addition to winning the ImageNet competition, Alex and his team were the first to use deep learning to perform image classification.
However, object detection is a more complex problem than image classification because it not only requires recognizing objects but also locating them within the image.

\subsection{RCNN / Fast RCNN and Faster RCNN}

\begin{center}
\begin{figure}[H]
\includegraphics[width=\linewidth]{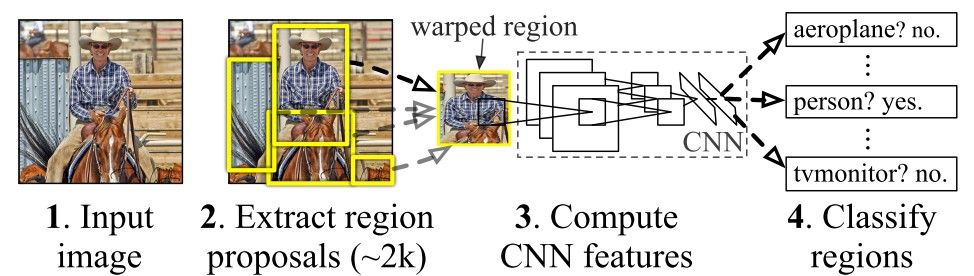}
\caption{RCNN operation, \citet{girshick2014rich}}
\label{RCNN}
\end{figure}
\end{center}

To address the localization problem, \citet{girshick2014rich} proposed the \acrfull{RCNN} model, which is capable of recognizing 80 different types of objects using a new technique that works as follows: fragmenting the image into thousands of regions (sub-images), then performing classification for each of them to detect the objects present (see Figure \ref{RCNN}).

\noindent First, the input image is divided into 2000 sub-images called ``regions of interest`` using the method known as \enquote{Selective Search} \citep{uijlings2013selective}. This method divides the input image into several parts, and each neighboring part is compared. The parts with similar characteristics in terms of color, texture, or shape are merged to form these regions. Each of these regions is then analyzed by \acrshort{RCNN}. First, they pass through a convolutional network to extract key features. Then, these features pass through an \acrshort{SVM} and a regression model to identify the objects present in these different regions and design bounding boxes.
Although this technique is slow, as each image is divided into 2000 regions before being analyzed by \acrshort{RCNN}, it remains effective and will be adopted by all subsequent object detection models, including Fast-\acrshort{RCNN} and Faster-\acrshort{RCNN}, which improve on the base model (\cite{girshick2015fast, ren2015faster}).
The main improvement of Fast-\acrshort{RCNN} is that selective search is no longer performed on the image beforehand but rather on the feature maps using a layer called the \enquote{\acrshort{RoI} pooling layer}. 
Each image passes through the convolutional layers to form feature maps. These maps are then sent to this new layer to extract regions of interest. Unlike the base model, the number of proposed regions is no longer fixed but varies depending on the input image. Then, a \acrshort{RoI} pooling layer is used to resize all the proposed regions so that they can be fed into a fully connected network for classification and detection of bounding boxes for each object.
Faster-\acrshort{RCNN}, on the other hand, replaces this selection method with a specific layer called the ``region proposal network`` (RPN), which directly generates regions of interest. This layer uses a sliding window approach to detect regions of interest, allowing it to detect objects at different scales and sizes. The regions of interest are then classified by a fully connected network. This approach is faster and more efficient than the selection methods used in previous versions.

\subsection{YOLO}

\subsubsection{YOLOV1}

\begin{figure}[H]
\begin{center}
\includegraphics[scale=0.4]{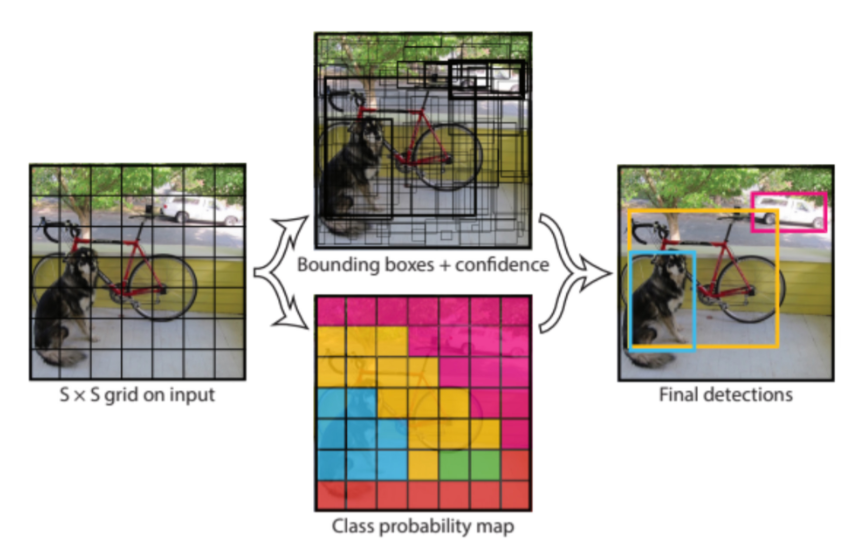}
\caption{How YOLO works, \citet{redmon2016you}}
\label{yolo}
\end{center}
\end{figure}

\begin{figure}[H]
\begin{center}
\includegraphics[scale=0.5]{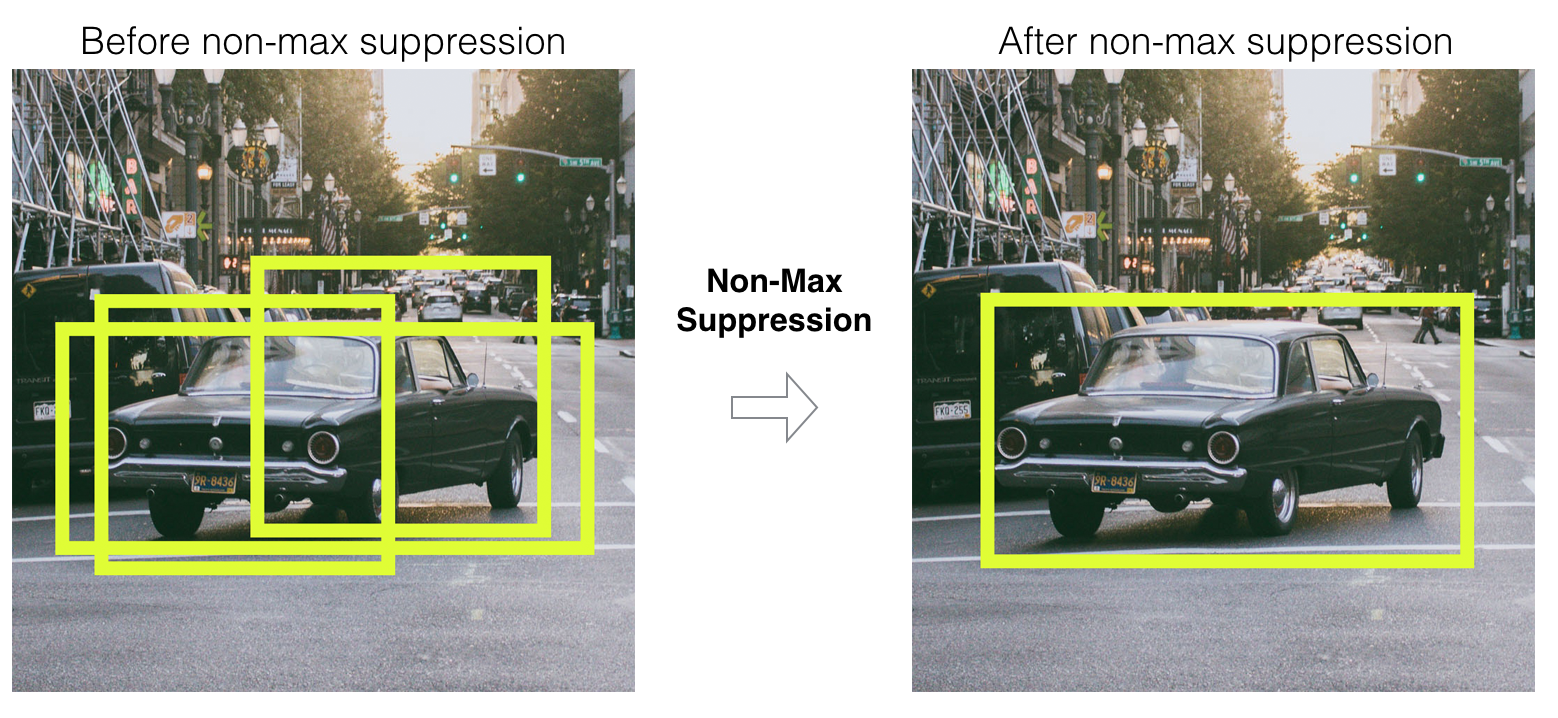}
\caption{Non-Maximum Suppression (NMS), \citet{jain2019incremental}}
\label{NMS}
\end{center}
\end{figure}

In 2015-2016, \citet{redmon2016you} published the \acrshort{YOLO} algorithm, a new object detection model that marked a turning point in the field. Unlike \acrshort{RCNN} and other models that treat detection as a special case of classification, \acrshort{YOLO} was designed as a regression task.
As the name suggests, You Only Look Once. \acrshort{YOLO} begins by dividing the image into an $x * x$ grid, then for each of these regions, a random number of anchor boxes are predicted (see Figure \ref{yolo}). These boxes can have different sizes and positions depending on the types of objects being detected. The downside is that this technique can result in the same object being detected by multiple boxes. To avoid this, a \gls{NMS} calculation is performed to keep only the bounding boxes that maximize the probability of each object detected (see Figure \ref{NMS}).
For each bounding box in the list, we retrieve the one with the highest score (confidence rate), then compare it to all other predicted boxes for the same class using the \gls{IoU} calculation. This calculation measures the degree of similarity/overlap between two boxes. Any boxes with a similarity exceeding the threshold set by the user will be considered as representing the same object and consequently removed. As a result, \acrshort{YOLO} is very fast and has become a reference model.

\subsubsection{YOLOV2}

Later, \citet{redmon2017yolo9000} introduced a new version of \acrshort{YOLO}, called \acrshort{YOLO}9000 or \acrshort{YOLO}V2. This model is capable of recognizing up to 9000 different objects in real-time thanks to a new architecture, Darknet-19. It is a convolutional network composed of 19 convolutional layers and 5 max-pooling layers, primarily using 3x3 filters. In the first layer, there are 32 filters, which are then doubled after each max-pooling layer. This means the number of features extracted increases progressively as the layers are stacked, allowing the network to capture more complex information as it advances through the model. Aside from the architectural change, the innovation here is using anchor boxes defined according to the reference dataset rather than arbitrarily delimiting them. This allows the model to perform better, as generally, an object fits a similar pattern regardless of the context.

\subsubsection{YOLOV3}

\cite{redmon2018yolov3} introduced their latest version, \acrshort{YOLO}V3. This new version changes the base architecture once again by integrating a convolutional network composed of 53 layers trained on IMAGENET and called Darknet-53. The activation function has also been modified. It shifts from a softmax activation where classes are treated dependently, with a total of 100\% for all probabilities, to a logistic classifier for each class, allowing for multi-class labeling. In addition to this, this version performs three detection steps, making the model more accurate, better at recognizing small objects, and allowing it to adapt better to various image sizes. Despite a slight loss in speed, this new version is more efficient and has become a reference for real-time object detection. The multi-class aspect, which is present in all later versions, is interesting in itself but is not particularly relevant to this thesis.

\subsubsection{YOLOV4}

\begin{figure}[H]
\includegraphics[width=\linewidth]{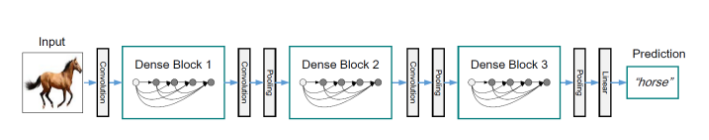}
\caption{DenseNet architecture, \citet{Huang_2017_CVPR}}
\label{denseNet}
\end{figure}

\begin{figure}[H]
\begin{center}
\includegraphics[scale=0.25]{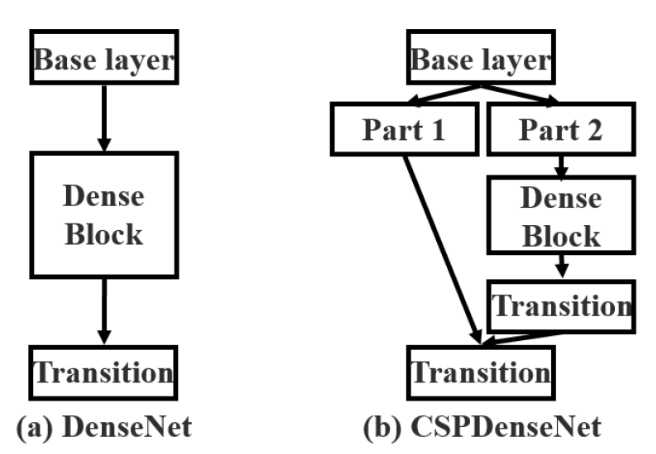}
\caption{DenseNet vs CSPDenseNet, \citet{wang2020cspnet}}
\label{denseVsCSP}
\end{center}
\end{figure}

\begin{figure}[H]
\includegraphics[scale=0.5]{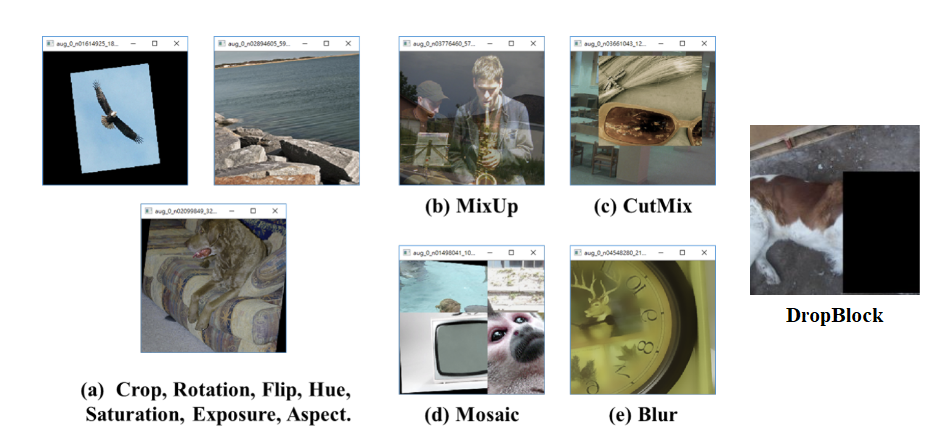}
\caption{YOLOV4 data augmentation, \citet{bochkovskiy2020yolov4}}
\label{yolov4DataAug}
\end{figure}

\citet{bochkovskiy2020yolov4}, a new team, proposed \acrshort{YOLO}V4 as a continuation of the previous version. This version is considered by the original team to be the logical next step in their work. The architecture is now based on a CSPDarknet53 model inspired by DenseNet networks. In DenseNet networks introduced by \citet{Huang_2017_CVPR}, each layer within a block transmits its output to all following layers within the block, which facilitates error backpropagation (see Figure \ref{denseNet}).
The CSPDarknet53 network follows a similar principle but uses CSPDenseNet layers that send only a part of the features into a dense block, while adding the rest directly to the output (see Figure \ref{denseVsCSP}). A second notable change is the loss function, now called \gls{CIoU}, which takes \gls{IoU} into account when calculating the error. The final change concerns the \gls{NMS} calculation, which filters the bounding boxes to avoid detecting the same object multiple times. This calculation has been replaced by \gls{DIoU-NMS}, which takes into account the centers of the compared boxes, so that two similar and close objects are not considered as the same object. Unlike the previous method, which deleted predicted bounding boxes and kept only the best, this new technique adjusts the confidence score for each bounding box. Among the additions, there are various data augmentation techniques performed before training the model (see Figure \ref{yolov4DataAug}), as well as a DropBlock method used during training, which removes the most representative area of an image to force the network to learn other features it might otherwise have ignored.

\subsubsection{YOLOF, YOLOP, YOLOR, YOLOS, YOLOX}

Later, in 2021, many variants of \acrshort{YOLO} were released, each addressing a different issue:
\begin{itemize}
\item YoloF (You Only Look One-level Feature) is an architecture based on the Resnet model combined with encoders [see Figure \ref{YoloF}].
\begin{figure}[!ht]
\includegraphics[width=\linewidth]{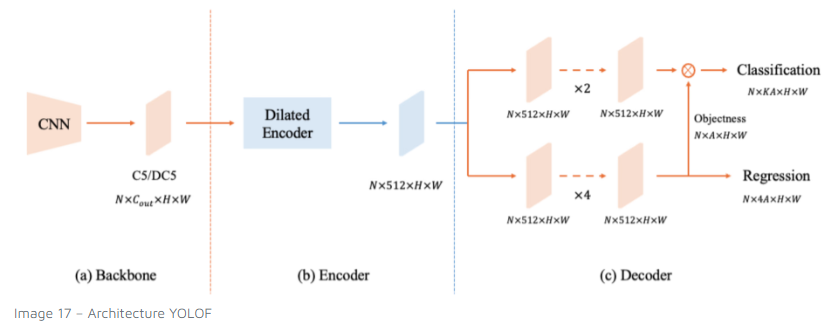}
\caption{YOLOF architecture, \citet{chen2021you}}
\label{YoloF}
\end{figure}

\item YOLOP (You Only Look Once for Panoptic Driving Perception) is specialized for autonomous driving (see Figure \ref{YoloP}).
\begin{figure}[H]
\includegraphics[width=\linewidth]{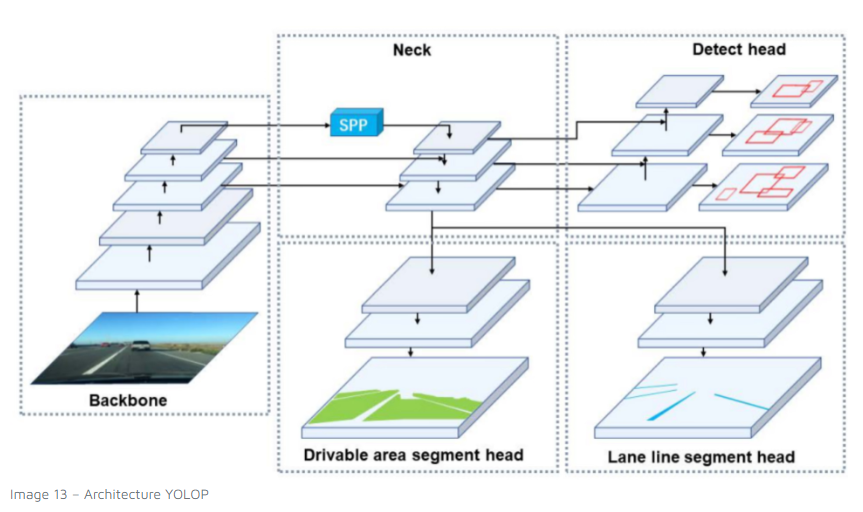}
\caption{YOLOP architecture, \citet{krishna2022you}}
\label{YoloP}
\end{figure}

\item YOLOR (You Only Learn One Representation) uses both explicit knowledge (from its learning) and implicit knowledge. Thanks to the implicit knowledge acquired, this model performs very well for multitask learning, and can therefore be used for other image analysis tasks [see Figure \ref{YoloR}]. 
\begin{figure}[H]
\includegraphics[width=\linewidth]{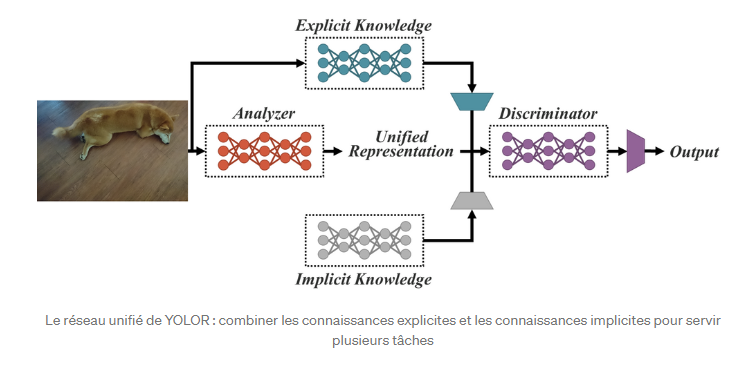}
\caption{YOLOR architecture, \citet{wang2021you}}
\label{YoloR}
\end{figure}

\item YOLOS (You Only Look at One Sequence) \citep{fang2021you} is a YOLO variant that uses Transformers instead of convolutional networks (see Figure \ref{YoloS}). Transformers are networks that have the advantage of possessing an attention mechanism, allowing them to detect important areas of an image that can be represented using attention maps. Currently, this variant is not optimized. The authors’ goal was simply to show that it is possible to use transformers for object detection.

\begin{figure}[H]
\includegraphics[width=\linewidth]{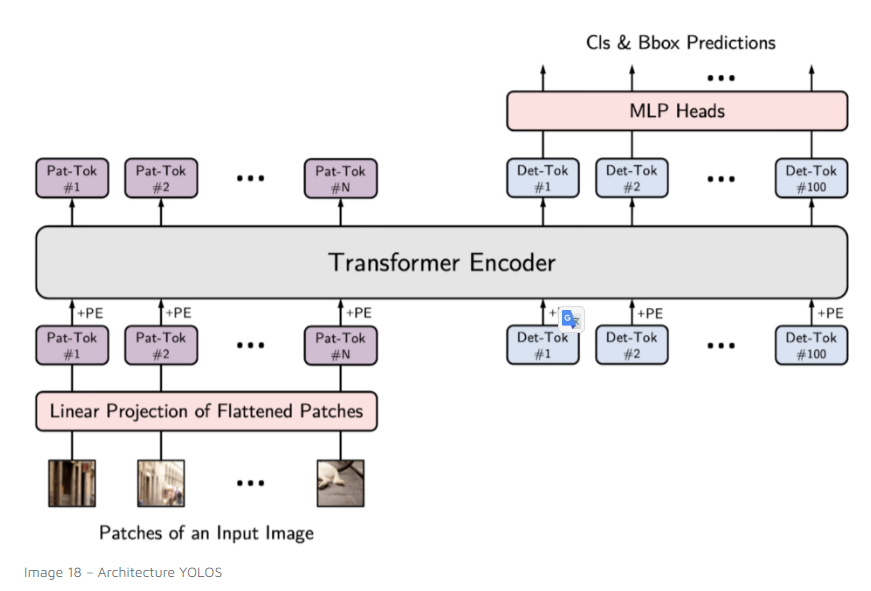}
\caption{YOLOS architecture \citep{fang2021you}}
\label{YoloS}
\end{figure}

\item \citep{ge2021yolox} introduced YOLOX, a variant of \acrshort{YOLO}V3 that does not use any anchor boxes, resulting in faster and more accurate performance.
\end{itemize}

\subsubsection{YOLOV5}

\begin{figure}[H]
\includegraphics[width=\linewidth]{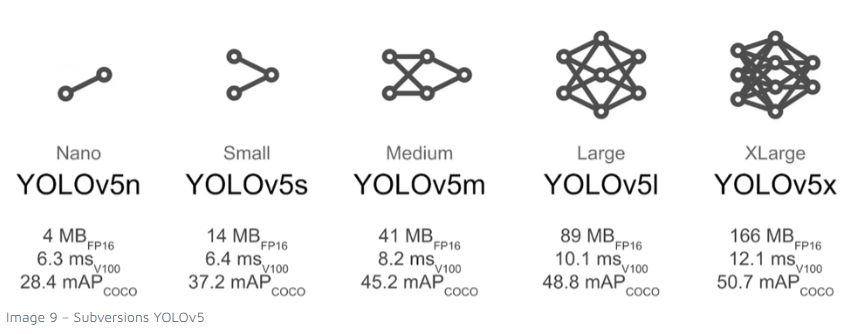}
\caption{Architectures proposed by \acrshort{YOLO}V5}
\label{Yolov5}
\end{figure}

A few months after the release of \acrshort{YOLO}V4, version 5 was released\footnote{\href{https://github.com/ultralytics/yolov5}{https://github.com/ultralytics/yolov5}}; however, it has not been subject to any official publication. This version does not implement or invent new techniques; it is simply the PyTorch extension of \acrshort{YOLO}V3 and not a continuation of the original code. Its main goal is to make \acrshort{YOLO} compatible with macOS and to offer five different versions of the model [see Figure \ref{Yolov5}].

\subsubsection{YOLOV6}

Recently, the Chinese company Meituan released MT-YOLOV6\footnote{\href{https://github.com/meituan/YOLOv6}{https://github.com/meituan/YOLOv6}}, an algorithm that is not officially part of the \acrshort{YOLO} series but is heavily inspired by the original \acrshort{YOLO}. It has no relation to version 5, but it surpasses it significantly in terms of accuracy and speed.

\subsubsection{YOLOV7}

\begin{figure}[H]
	\begin{center}
		\includegraphics[scale=0.7]{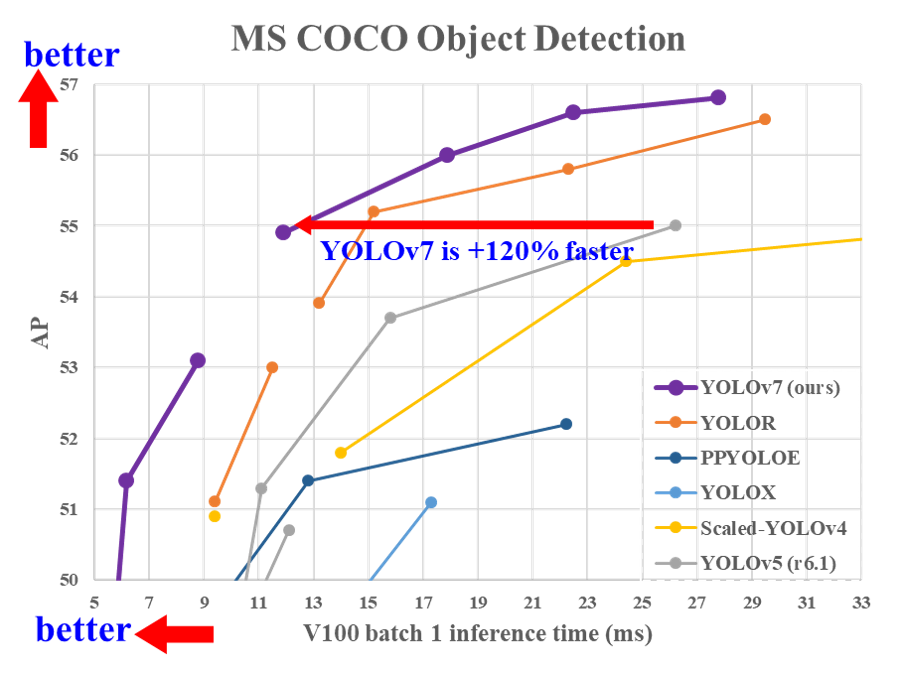}
		\caption{Performance of \acrshort{YOLO}V7, \citet{wang2022yolov7}}
		\label{yolov7Perf}
	\end{center}
\end{figure}

\begin{figure}[H]
	\begin{center}
		\includegraphics[width=\linewidth]{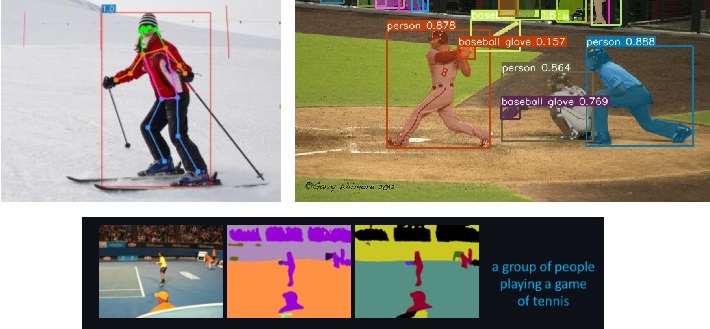}
		\caption{Applications of \acrshort{YOLO}V7, \citet{wang2022yolov7}}
		\label{yolov7Appli}
	\end{center}
\end{figure}

\cite{wang2022yolov7} proposed version 7, developed by the authors of version 4, which is based on YOLOR. It is said to be 120\% faster than \acrshort{YOLO}V5 and more efficient than all other available architectures, including the previously most performant YOLOR version (see Figure \ref{yolov7Perf}). Figure \ref{yolov7Appli} shows different applications enabled by this version.

\begin{figure}[H]
	\centering
	\includegraphics[width=\linewidth]{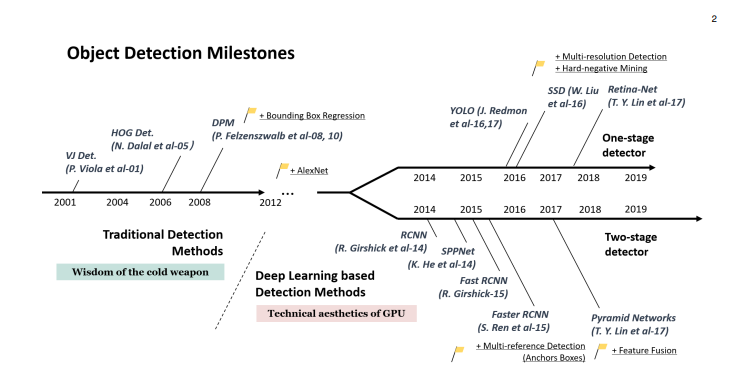}
	\caption{Evolution of object detection (\cite{zou2019object})}
	\label{zou}
\end{figure}

\noindent More details about the functioning of these models can be found in \citet{PulkitSharma1, PulkitSharma2, PulkitSharma3}. \citet{zou2019object} review the evolution of the field over the last twenty years, summarized in Figure \ref{zou}.

\section{Explainability}

Techniques that improve explainability in the visual domain fall into two categories: \\

\noindent On one hand, there are model-agnostic technologies, such as LIME (Local Interpretable Model-Agnostic Explanations) proposed by \citet{ribeiro2016should}, and SHAP (SHapley Additive exPlanations) proposed by \citet{lundberg2017unified}.\\

\noindent On the other hand, there are technologies specific to certain types of models, such as convolutional neural networks, for which we find techniques like convolutional filter visualization (\cite{jiang2021layercam}), saliency maps (\cite{smilkov2017smoothgrad, simonyan2014deep}), activation maps (\cite{selvaraju2017grad, aditya1710grad, wang2020score}), etc. \\

\noindent To visualize these features, there are many available libraries.
\citet{raghakotkerasvis} proposed Keras-vis, a public library that allows the visualization of convolutional neural network features. It allows for the visualization of convolutional filters in each layer, which are the weight matrices that detect patterns in an image. These filters can be visualized as pixel matrices, helping to better understand the patterns each filter is sensitive to.
Moreover, Keras-vis also allows the visualization of how these filters evolve throughout training. This can be useful to understand how the network learns to extract increasingly complex features over time.
Finally, the library also enables the visualization of activation maps, which show the regions of an image that were activated by each filter. These maps can help to understand what the network ``sees`` in the image at each processing step, which can be useful for diagnosing potential issues with the network's operation.
Later, \citet{Keract} developed the Keract library with similar features. In the same year, \citet{gotkowski2020m3d} proposed another library that allows the visualization of both 2D and 3D attention maps. The Keras library, developed by \citet{chollet2015keras}, also includes some of these visualization techniques.
For a more in-depth study of explainability techniques, refer to \citet{molnar2019}.

\section{Conclusion}

Throughout this study, various technologies have been identified for anomaly detection, including Convolutional \acrshort{RNN}, ConvRNN, autoencoders, Transformers, \acrshort{TCN}, and \acrshort{GAN}. In order to highlight our strategic choices and provide a solid foundation for our approach, a comparison outlining the advantages and disadvantages of each option is presented in Table \ref{tabDetectionAnomalie}.

\begin{table}[H] 
	\centering 
	\small 
	\caption{Comparison of different anomaly detection techniques} 
	\resizebox{\columnwidth}{!}{
		 \begin{tabular}{c | c | c | c | c | c | c | c} 
	& \acrshort{CNN} + \acrshort{RNN} & ConvRNN & \acrshort{C3D} & \acrshort{TCN} & \acrshort{GAN} & Auto-encoder & Transformer \\ \hline 
	Documentation & High & High & High & Low & High & High & High \\ 
	Implementation & Simple & Simple & Medium & Difficult & Difficult & Difficult & Difficult \\
	Computational Power & Low & Low & Moderate & Moderate & Huge & Variable & Huge \\ 
	Mode & Supervised & Supervised & Supervised & Supervised & Unsupervised & Unsupervised & Supervised \\ 
	Speed & Fast & Fast & Fast & Slow & Slow & Slow & Slow \\ 
	\end{tabular}
	} 
	\label{tabDetectionAnomalie}
\end{table}

\noindent Due to our constraints in computational power, caused by the unavailability of GPU servers, we were forced to rule out \acrshort{GAN} and Transformers. Additionally, due to our real-time requirements, autoencoders were also excluded. Their need to reconstruct each frame of our videos would have introduced significant delays. Because of their lack of documentation and perceived inefficiency for our specific anomaly detection needs in videos, \acrshort{TCN} were also discarded.
\label{choixCGRU} After analyzing the available datasets for addressing our problem, we found that the vast majority of them are oriented towards supervised approaches, contrasting normal and anomalous classes. In this regard, Convolutional \acrshort{RNN}, ConvRNN (which can be combined with \acrshort{GRU} or \acrshort{LSTM} for the \acrshort{RNN} part), and 3D convolutions emerged as the most suitable and appropriate choices for our approach.
\label{choixYolo} Regarding object detection, several models can be considered, such as \acrshort{RCNN}, Fast-\acrshort{RCNN}, Faster-\acrshort{RCNN}, and \acrshort{YOLO}. The specifics of each model are summarized in Table \ref{tabDetectionAnomalie}, allowing for an in-depth comparison of their respective characteristics.

\begin{table}[H] 
	\centering 
	\small 
	\caption{Comparison of various object detection techniques discussed} 
	\begin{tabular}{c | c | c | c | c } 
		& \acrshort{RCNN} & Fast-\acrshort{RCNN} & Faster-\acrshort{RCNN} & \acrshort{YOLO} \\ \hline
		Documentation & High & High & High & High \\ 
		Implementation & Simple & Simple & Simple & Simple \\ 
		Computational Power & Moderate & Moderate & Moderate & Moderate \\ 
		Speed & Slow & Slow & Fast & Very Fast \\ 
	\end{tabular} 
	\label{tabDetectionObjet} 
\end{table}

\noindent However, given the real-time constraint due to the impact that the anomalies we are interested in can have, the fastest models should be prioritized. While Faster-\acrshort{RCNN} is quite fast and yields very good results, for real-time problems, models that perform single-step analysis, such as \acrshort{YOLO}, are better candidates (\cite{RohithGandhi}). According to \citet{doshi2020continual}, \acrshort{YOLO} would be the best algorithm in the group. To quote them: ``Compared to other state-of-the-art models like SSD and ResNet, \acrshort{YOLO} offers a higher images-per-second (\acrshort{FPS}) rate while achieving better accuracy. For online anomaly detection, speed is a critical factor, and therefore, we currently prefer using \acrshort{YOLO} V7.`` Additionally, \acrshort{YOLO} has seen significant improvements in recent years, especially in terms of accuracy and processing speed.
With its various versions, \acrshort{YOLO} remains the fastest model available today. At the time of finalizing this thesis, version 8 had appeared, which we did not have the opportunity to test.

\part{Implemented System}
\parttoc
\label{part:contribution}
\chapter{System}  
\chaptermark{System}{}  
\minitoc  
\newpage  

\section{Introduction}  

In this chapter, we will present our work and contributions. We will begin by introducing the overall architecture of our system, its components, and the datasets created for the training phase. We will then explain how the experiments conducted and the results obtained influenced our choices and approaches. Finally, we will conclude this chapter with a synthesis of our work.  

\section{Overall Architecture}  

\begin{figure}[H]  
    \centering  
    \includegraphics[width=\linewidth]{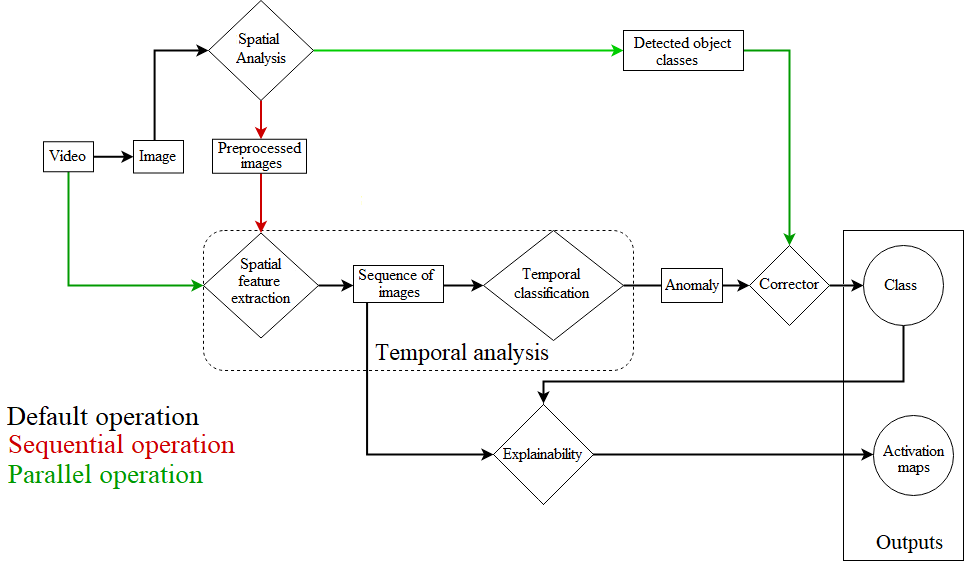}  
    \caption{Overall Architecture}  
    \label{architectureGlobale}  
\end{figure}

In Section \ref{combiner}, we introduced our goal of developing an anomaly detection system capable of identifying risks to individuals by combining spatial analysis performed on images (including object detection) with temporal analysis of video streams.  
Figure \ref{architectureGlobale} illustrates the general principle of our system, which follows two possible workflows. The first workflow, shown in red, adopts a sequential configuration where the temporal component processes the output of the spatial component. The second workflow, shown in green, follows a parallel configuration where each image sequence is processed simultaneously by both components before being classified by combining their detections. Lastly, our system includes an explainability component that highlights the areas of attention leading to the model’s detection. This component can be disabled to enhance processing speed.  

\noindent In the following sections, we will explain each component of the figure above (the diamonds), before summarizing the overall functioning of the system.

\section{Spatial Analysis Component}  

The first element of our system performs a spatial analysis of each image included in the video to be processed. It receives an image as input and outputs it to two different components depending on the chosen operating mode: in parallel mode, it simply transmits the class of the detected objects to the Correction component; in serial mode, it transmits the pre-processed image to the Spatial Feature Extraction component. Depending on the chosen model, this preprocessing may include object detection, image segmentation, or human pose analysis.  
We chose \acrshort{YOLO} for the reasons discussed in the problem statement (processing speed and performance, see section \ref{choixYolo}), as well as because this family of models is constantly being researched and improved by the scientific community. In our preliminary tests, we compared \acrshort{YOLO} (at the time, version 3) with Faster-\acrshort{RCNN}, the fastest of the other models, and found that they had similar execution times. However, version 7 of \acrshort{YOLO} outperforms them significantly in speed. In this study, we compared the performance of versions 3, 4, and 7. The details of the tests and results can be found in section \ref{sec:AnalyseSpatiale}.  
This model is pre-trained on the Pascal-VOC\footnote{\url{https://pjreddie.com/projects/pascal-voc-dataset-mirror/}} and COCO\footnote{\url{https://cocodataset.org/\#home}} datasets, enabling it to recognize 80 types of objects, including: humans, vehicles (truck, bicycle, motorcycle, bus, car, airplanes, train, boats), animals (dog, cat, giraffe...), food (pizza, banana...), and everyday objects (phone, television, chair…). Among all these classes, very few are useful for our problem. Only humans and vehicles can be exploited in cases of accidents or fights. Moreover, crucial objects for our problem, such as knives, guns, or flames, which allow us to detect potential shootings or fires, are not included.  
It was therefore essential to build our own dataset in order to re-train \acrshort{YOLO}, which is described in the next subsection.  

\subsection{Images}  

Our dataset is proprietary and combines images extracted from videos used for our temporal analysis with images that we have collected and labeled ourselves. Contrary to what one might think, the images used do not necessarily need to be realistic, and it is not mandatory for them to come exclusively from surveillance videos.  
We collected over ten thousand images representing firearms. We initially divided them into two classes: ``gun`` for small firearms and ``weapons`` for larger ones. However, any weapon represents a potential risk regardless of its size or appearance. Therefore, we ultimately decided to combine them into a single class, which resulted in improved performance for our system.  
We also created a class of images to identify flames, which is important in the case of fires. We had also created a class representing knives, but ultimately decided not to include it in our final dataset. This class contained very different objects, and the action itself was more akin to an assault or a fight.  
Currently, our dataset contains three classes:  

\begin{enumerate}  
\item Firearms: about 10,000 images,  
\item Flames: more than 2,000 images,  
\item Humans (representing each individual present in our images).  
\end{enumerate}  

\noindent Images in which none of these classes are present have also been added to our dataset for both training and testing.  
This dataset is augmented using the data augmentation techniques included in \acrshort{YOLO}, as previously mentioned (figure \ref{yolov4DataAug}). We used more than ten different techniques applicable to each of our images, resulting in nearly 150,000 images in total.

\section{Temporal Analysis}  

\begin{figure}[H]  
   \includegraphics[width=\linewidth]{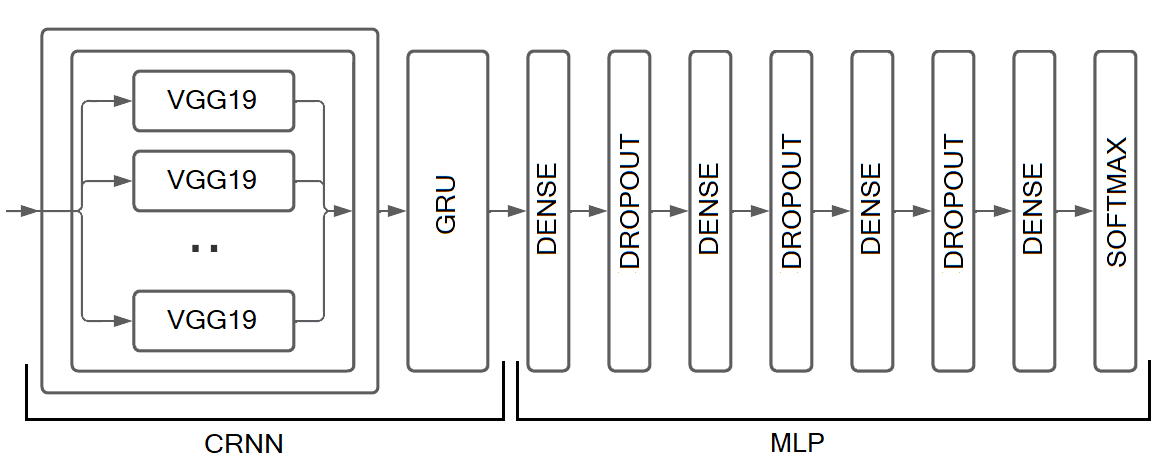}  
   \caption{Architecture of the temporal analysis module (CGRU)}  
    \label{archiCGRU}  
\end{figure}

For the reasons outlined in section \ref{choixCGRU} of the state of the art, we chose well-documented, easy-to-implement, and resource-efficient approaches for temporal analysis: Convolutional \acrshort{RNN}s, ConvRNN, and 3D convolutions. After conducting detailed tests in section \ref{sec:AnalyseTemporelle}, we ultimately selected an architecture based on Convolutional \acrshort{GRU} (\acrshort{CGRU}), as detailed in figure \ref{archiCGRU}.  
This processing is performed by two distinct components: a convolution (spatial feature extractor) and a \acrshort{GRU} (temporal classifier), which we will examine below.  

\subsection{Spatial Feature Extraction}  

The second component of our system is dedicated to spatial feature extraction; it receives a video sequence and returns the sequence after highlighting its features. In serial mode, the sequence it receives as input corresponds to the original images modified by \acrshort{YOLO}; in parallel mode, it simply receives the input video. The output is then passed to the component that performs temporal feature extraction and (optionally) to the explainability module.  
For spatial feature extraction, we chose \acrshort{VGG}19. However, since this convolution is 2D, it is not suitable for processing image sequences. If we apply it directly to our data, we will lose information because all the images in our sequence will need to be merged to fit a 2D format. This is why we included it in a time-distributed layer provided by the Keras library. This layer allows us to apply one or more layers to each time slice of our input.  
In our case, applying \acrshort{VGG} to each image in our sequence helps accumulate features from different images and form the sequence for analysis. This time-distributed layer receives a video sequence as input, extracts each image, and passes it to the \acrshort{VGG} network for extracting relevant features. Once these features are extracted for each image, we obtain a new time series output that will be passed to the next component.

\subsection{Detection}  

The next component consists of two parts: a temporal feature extraction component, applied to the output of the time-distributed layer, and a classification component, which performs the detection of the alert type to be associated with the sequence.  
Temporal feature extraction is carried out by a \acrshort{GRU} followed by a few fully connected layers and some forget layers to prevent overfitting.  
Each sequence passed by the time-distributed layer goes through the forget gate of our \acrshort{GRU}, which controls how much information should be forgotten. The information retained, deemed relevant, will be sent to the update gate for learning. This gate's role is to concatenate the new data with that from the previous state and pass them through a sigmoid function to detect the important features.  
After passing through the update gate, the relevant information is then processed by the fully connected layers of our classifier, a \acrshort{MLP}, to establish the final detection.

\subsection{Videos}

\begin{figure}[H]
\includegraphics[width=\linewidth]{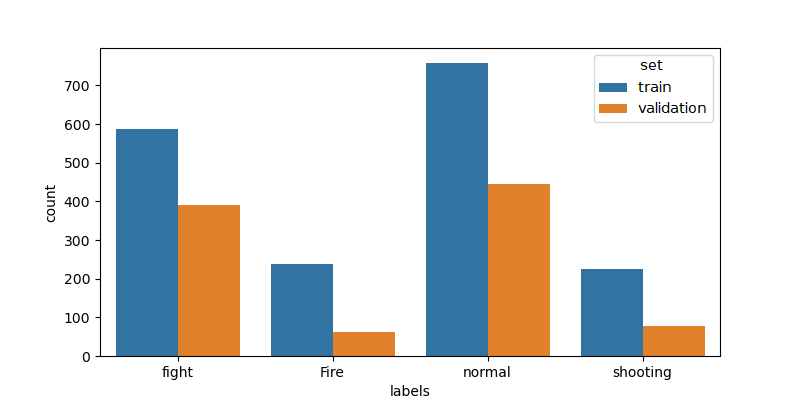}
\caption{Data Distribution}
\label{fullDataVideo}
\end{figure}

For the reasons outlined in the problem statement (see page \pageref{notrejeu}), we decided to create our own video dataset to train our \acrshort{CGRU}. This dataset is independent of the image dataset we created for object learning.
This dataset is proprietary and follows the model established by UCF Crimes, meaning we oppose anomalies to the normal class. However, since data collection is a time-consuming task, we limited ourselves to three main classes: fights, gunshots, and fires. Additionally, some of these categories were more difficult to collect than others due to the rarity of these events or censorship. Unlike the normal class, where any daily video can be considered representative, anomalies have different frequencies of occurrence, which can affect the amount of data available. To ensure maximum balance, we collected as many videos representing an anomaly as we did normal cases.
To assess whether our dataset can be considered reliable, we need to check if our system can achieve good performance after training. Therefore, we divided our dataset into binary sub-categories, each time opposing an anomaly to the normal class, allowing us to train models using only certain classes and evaluate the reliability of each one.
Our dataset, described in tables \ref{videoDataRepartition} and \ref{videoDataStats}, consists of three classes representing a wide selection of videos from surveillance cameras or smartphones. These videos have varying qualities and resolutions, cover different angles of view, and were recorded at different times of the day. Figure \ref{fullDataVideo} shows their distribution by class.

\begin{table}[H]
\caption{Video Distribution in Our Dataset} 
\begin{center}
\begin{tabular}{c | c | c | c  }
 Classes & Train & Validation & Total \\ \hline
 Fight & 587 & 391 & 978 \\ 
 Gunshot & 247 & 64 & 311 \\ 
 Fire & 237 & 61 & 298 \\ 
\end{tabular}
\label{videoDataRepartition}
\end{center}
\end{table}

\begin{table}[H]
\caption{Duration of Anomalies in Our Dataset} 
\begin{tabular}{c|cccc|cccc}
 \hline
 & \multicolumn{4}{c|}{Train} & \multicolumn{4}{c}{Validation} \\
Classes & Min & Average & Max & Total & Min & Average & Max & Total \\ \hline
 Fight & 1.2s & 5.1s & 140s & 50.05min & 1s & 4.8s & 139s & 31.50min \\ 
 Gunshot & 1s & 4.5s & 16s & 17.40min & 1s & 36.5s & 36s & 8min \\ 
 Fire & 1.3s & 51.3s & 504s & 3h40min & 1.4s & 49.8s & 211s & 46min \\ 
\end{tabular}
\label{videoDataStats}
\end{table}

\section{Correction Component}
\label{ComposantCorrection}

The fourth component of our system is used to combine, when operating in parallel mode, the object detection results with those of the temporal analysis. Its role is to correct the predicted class based on the results of the spatial analysis performed by \acrshort{YOLO}.
To achieve this, we have created a dictionary where each object type learned by \acrshort{YOLO} is linked to an anomaly: flames with the ``fire`` anomaly and firearms with the ``gunshot`` anomaly.
Once the class has been predicted by our \acrshort{CGRU} model, we analyze the objects detected by \acrshort{YOLO} to check if it has detected a flame or a firearm with a probability exceeding the set confidence threshold.
If a flame is detected and the \acrshort{CGRU} model has predicted the ``normal`` class, we change this to the``fire`` class based on the detection probability. In the case where a firearm is detected, we concluded that it only represents a risk when it can be handled by a person. Therefore, we calculate the \gls{IoU} between this firearm and all the persons detected whose score exceeds the set confidence threshold to determine if a person is in contact with the firearm. If this is the case, we correct the predicted class by changing it to the ``gunshot`` class.

\section{Explainability}
\label{explicabilite}

\begin{figure}[H]
\centering
\includegraphics[width=\linewidth]{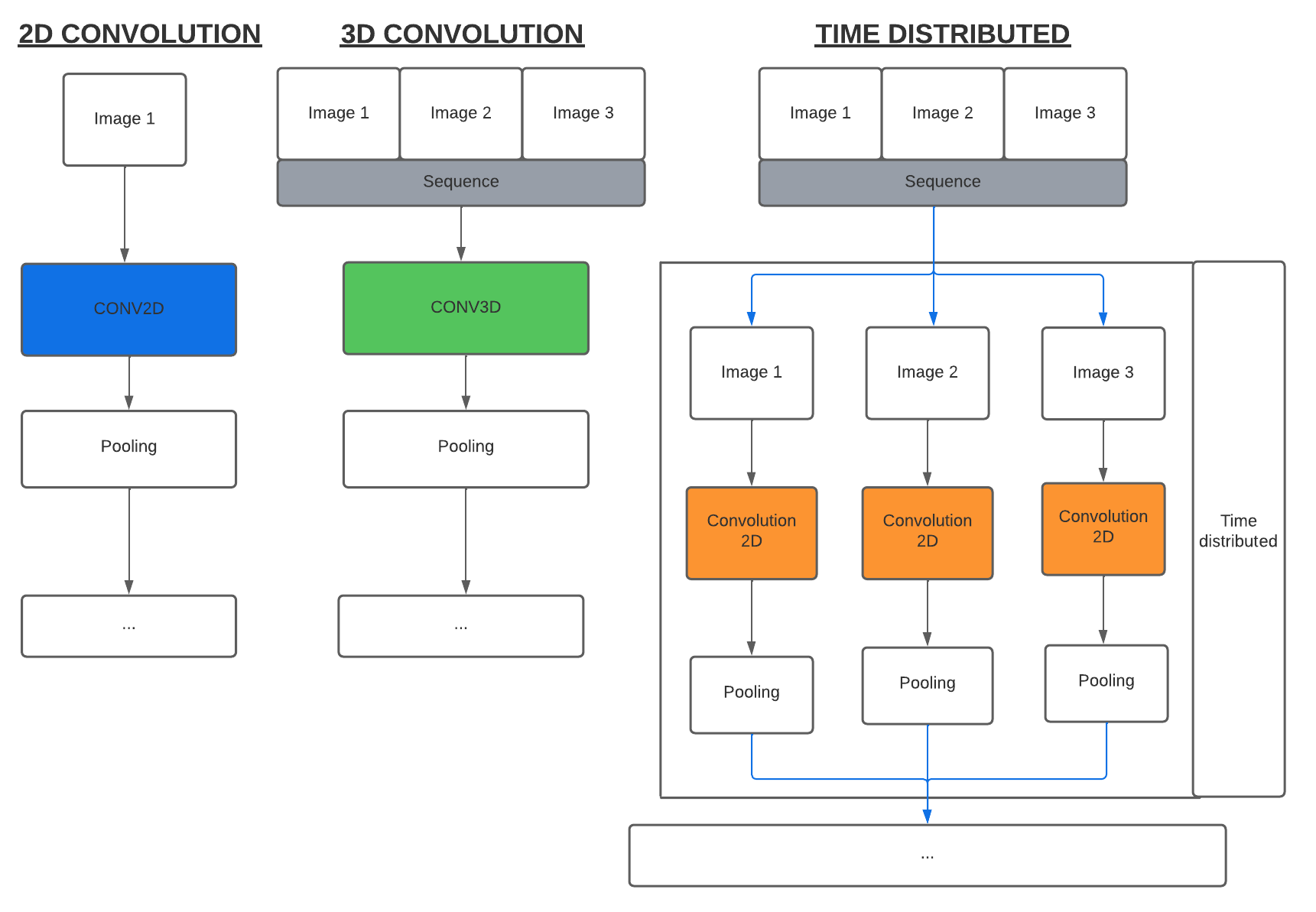}
\caption{2D Convolution / 3D Convolution / Convolution embedded in a Time Distributed Layer}
\label{ComparaisonConvolution}
\end{figure}

The European Union defines explainability as the ability to understand and explain how a decision was made by an automated system. In this context, our explainability module aims to justify the decisions made by the components of our anomaly detection system, particularly to facilitate understanding of the decisions made by our temporal analysis model (\acrshort{CGRU}), and more specifically, our \acrshort{VGG} convolution working with images.
In terms of explainability, it is much easier to interpret the features learned by our convolutions since they are visual, unlike those of our \acrshort{RNN}. At first glance, it seems that none of the visualization libraries specific to \acrshort{CNN} are suitable for our type of network, for the reasons detailed below (see figure \ref{ComparaisonConvolution}).
We have thus drawn inspiration from techniques used to explain image classification. The principle is to visually highlight the part of the image that caused the decision-making process, adapted to video.

In the case of a model using 2D or 3D convolution layers, all layers of the network are directly connected. However, in our architecture, we use a convolution that is accessed through a time distributed layer, which is not directly connected to the other layers of the network, thus blocking the flow of information to the other layers.
The second issue concerns the third dimension added by the time factor. Unlike models that handle images or 3D objects where we would have a single input and therefore a single output visualization, here we are dealing with videos. Thus, we have multiple images as input but only one visualization as output, representing the entire sequence.
We therefore performed a visualization for each of our images to understand the processing done by our sub-model. The goal is to visualize the areas on which the model relies to detect an anomaly. For this, we will primarily use saliency maps as well as activation maps.
To perform this type of visualization, we will propagate information through our network to obtain the final activation. This will allow us to generate saliency maps by calculating the gradient of this activation with respect to our input data, and activation maps by calculating the gradient of this activation with respect to the output of the layer we wish to visualize.
For the saliency maps, we will need to calculate our gradient from a sequence rather than an image. Fortunately, the result will be the same size as the input data, meaning we will get a list of gradients equal to the size of our sequence. We will then display these gradients to create a saliency map for each image.
Regarding the activation maps, our only solution will be to use the output of the time distributed layer. As explained earlier, its purpose is to add the time factor to our data by allowing us to perform the same processing for each image in our sequence, in this case by applying \acrshort{VGG}19 to each image. The output of this layer can then be viewed as a list of results, containing an output for each of our images. Our gradient will have the same dimension as the output of this layer. We can then apply each gradient to its corresponding output to obtain our activation map and project it onto the relevant image.
We have recorded the results of our visualizations in parallel with the visualizations intended for the operator. This approach facilitates debugging and allows us to deactivate the generated visualizations when used in production, speeding up the process. This method also ensures a consistent visualization for each image rather than having changing attention areas, which would disturb people analyzing the videos. This is because convolutional neural networks do not have a fixed attention area, unlike vision transformers. Instead, they scan the images, looking for features, hence attention areas that move.

\section{Operation}
\label{Fonctionnement}

\begin{figure}[H]
    \centering
    \includegraphics[width=\linewidth, height=20cm]{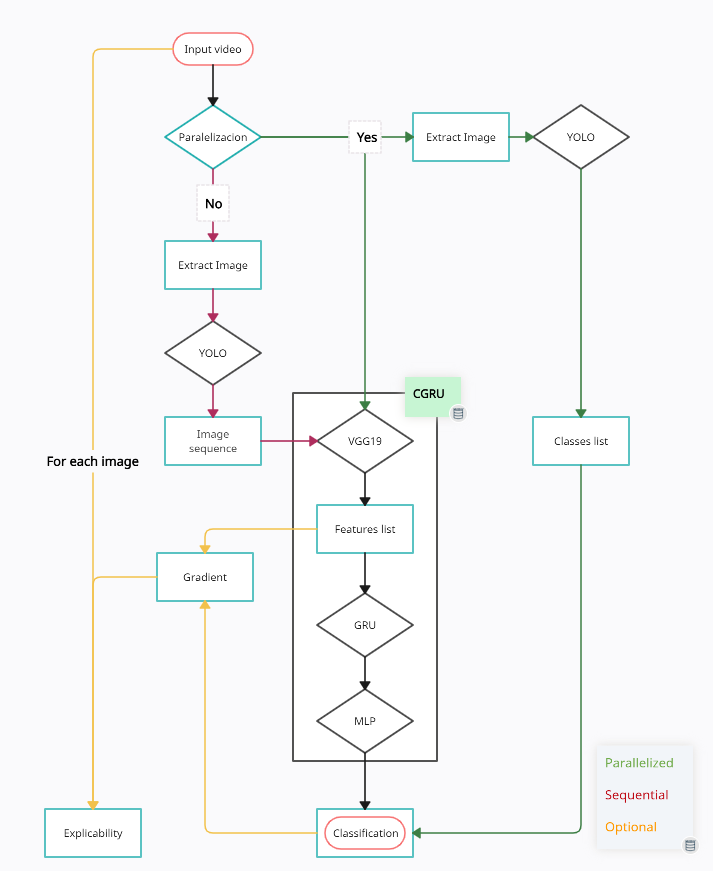}
    \caption{Overall Operation}
    \label{architectureGlobaleFonction}
\end{figure}

The operation of our anomaly detection model is presented in figure \ref{architectureGlobaleFonction}. It relies on a sophisticated architecture that integrates the two types of analysis mentioned: temporal analysis and spatial analysis. For temporal analysis, we adopted a Convolutional \acrshort{GRU} (\acrshort{CGRU}) approach, which consists of a combination of \acrshort{VGG}19, \acrshort{GRU}, and an \acrshort{MLP}. This temporal analysis module is capable of processing sequences of 20 consecutive images, which allows capturing temporal evolutions of anomalies in continuous video streams or finite videos. The \acrshort{CGRU} used can be multi-class or binary, depending on the issue being addressed, but it does not operate in multi-label mode, thus simplifying the classification process.
Regarding spatial analysis, our model leverages \acrshort{YOLO}v7. In sequential mode, the preprocessing applied to the images varies depending on the specific \acrshort{YOLO} architecture chosen. Preprocessing operations can include tasks such as image segmentation, object detection, or even human pose analysis. In parallel mode, the system identifies the objects present in a sequence of images and then uses this information to perform the final detection. This detection is performed according to a set of rules defined as follows:

\begin{itemize}
        \item [$\Rightarrow$] Let A be the set of anomalies known by the model. 
        \item [$\Rightarrow$] Let O be the set of key objects known by the model. 
\end{itemize}

\begin{enumerate}
    \item $\forall \text{ anomaly} \in A : \text{class} = \text{anomaly}.$         
    \item $\neg (\exists \text{ anomaly} \in A) \land (\exists \text{ key object} \in O) : \text{class} = \text{anomaly} \cap \text{object}$.
    \item \begin{align*}
        \text{class} = \begin{cases}
        \text{gunshot} & \text{if } (\text{key object} = \text{firearm}) \\ 
        & \text{and } (\text{IOU between this object and a person} > 0) \\
        \text{normal} & \text{otherwise}
        \end{cases}
    \end{align*}
\end{enumerate}

In summary:
\begin{itemize}
    \item For any detected anomaly, the class is defined as that anomaly.
    \item If no anomaly is detected but a key object known by the model is present in the sequence, then the class is defined as the anomaly associated with that key object.
    \item If the detected key object is a firearm, the system checks if the \gls{IoU} between this firearm and a detected person is greater than 0. If true, it means a person is near the firearm, and the class is defined as ``gunshot``. If the \gls{IoU} is less than or equal to 0, the class is defined as ``normal``.
\end{itemize}

\noindent Thus, the system takes into account the presence of anomalies, key objects, and the relationships between them to assign the appropriate class to each detected situation in the image sequence.
A fundamental feature of our system is its explainability module, which can be used independently of the chosen operational mode. This module allows explaining the decisions made by the various components of our model, especially the \acrshort{CGRU}, by highlighting the attention areas that influenced the detections. This functionality is crucial for improving the transparency and understanding of the model's results, thereby contributing to its usefulness and adoption.
In summary, our anomaly detection model cleverly combines temporal and spatial analysis to identify abnormal behaviors in either finite videos or continuous streams. By utilizing models such as \acrshort{CGRU} and \acrshort{YOLO}v7, while offering customization possibilities depending on the specific application needs. In the next section, we will delve into the experiments and results that played a crucial role in our choice and configuration of these models.

\section{Conclusion}

Throughout this chapter, we have presented in detail each component of our architecture, whether it is the spatial analysis utilizing \acrshort{YOLO}v7 or the temporal analysis via \acrshort{CGRU}, while highlighting the essential role they play in the overall anomaly detection process. Additionally, we have also discussed the datasets used to train each of these models, as well as the preprocessing methods applied to them. Now, we will explore the results of the experiments we conducted, which guided our decision to adopt these models for our setup.

\chapter{Experiments and Results} 
\chaptermark{Experiments and Results}{} 
\minitoc 
\newpage

\section{Introduction}

Since I was unable to access real surveillance cameras during my PhD, all the videos used in the experiments will represent finite information streams (videos downloaded locally). Video data requires significant computational power for processing, and since I did not have access to a GPU server either at my company or research lab, all the experiments presented in this thesis were carried out on a computer equipped with 32 GB of RAM, an Intel Core i9 processor with 16 cores clocked at 2.3 GHz, and an Nvidia GeForce RTX2080 graphics card with 8 GB of dedicated RAM.
In this chapter, we will discuss the experimental phase during which we combined our two models to evaluate their performance in anomaly detection. We will examine how the results vary depending on whether the mode is chosen in parallel or serial. After that, we will present the experiments carried out in object detection and anomaly detection, which allowed us to determine which models to choose for each of our approaches. Finally, we will conclude this section with a global summary of our work.

\section{Data Preprocessing}

\begin{figure}[H] \includegraphics[width=\linewidth]{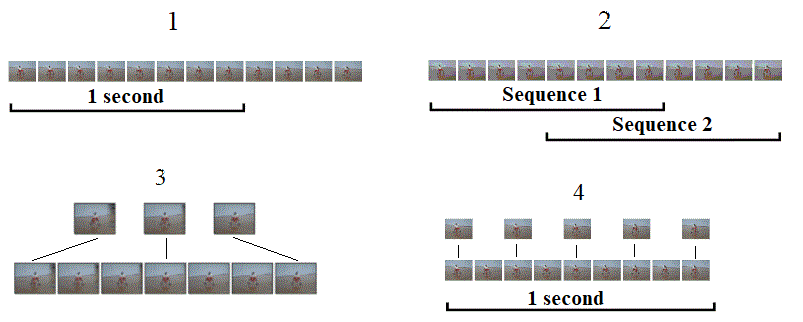} \caption{Functioning of the tested generators \citep{patriceFerlet}} \label{generator} \end{figure}

In this section, we detail the preparation techniques applied to our data.
As we saw in Section \ref{problematique}, a video can contain one or more anomalies bounded by normal sequences. Because of this, we had to split videos containing anomalies to isolate the anomalies and facilitate their analysis. Despite this, managing the data remains a real challenge for this project, particularly due to the large amount of video data to be processed.
Loading the entire dataset at once is simply not feasible due to the memory limitations of our system. To solve this problem and load our data incrementally into memory, we opted to use a generator\footnote{\href{https://github.com/metal3d/keras-video-generators}{Source code of the generator}}.
Our first experiment, therefore, was to compare the effect of different generators. We primarily tested four (see Figure \ref{generator}):

\begin{enumerate} \item With a sliding window to form successive sequences. \item With a sliding window and overlap between sequences. \item With a dynamic step, collecting $x$ images per video. \item With a sliding window and dynamic step. \end{enumerate}

\noindent The generator that gave us the best results was the third. This is because it addresses the two main disadvantages of generators using a sliding window (generators 1, 2, 4). Indeed, models trained with one of these generators have a learning time that depends on the length of the videos: the longer the videos to be processed, the longer the learning time. Their second disadvantage is that the size of the window can be quite difficult to define, as for each video, it must contain the entirety of the action we want the model to learn, otherwise, it could affect its performance.
Despite its advantages, the third generator still has a significant drawback, also related to the sequence configuration. For each video processed, it will calculate a dynamic step based on the number of FPS of the video and the number of images set for forming our sequences. The problem here is particularly apparent for datasets with actions of different durations, as in this case, the time between images can vary greatly. For short actions, only a few seconds will separate the images forming our sequence, while in long videos, there may be several minutes between each image. Therefore, the shorter the video, the more detailed it will be, compared to a much less detailed long video.
Due to its ability to generate sequences containing non-successive images, the third generator gives us the possibility, by changing the step, to set the desired detection interval: ranging from detection per sequence to detection per video. This allows us, in the end, to handle both finite videos and continuous video streams.
Once our generator was selected, we focused on the size of our data. We varied the size of our images; Table \ref{tailleIMG} shows that the optimal size is $112 \times 112$.

\begin{table}[H]
	\begin{center} 
	\caption{Variation in the size of our images} 
	\begin{tabular}{c|c|c|c} Image Size & Accuracy & Recall & F1-Score \\ \hline 
		80*80 & 20.2\% & 85.8\% & 32.8\% \\
		112*112 & \textbf{22.9\%} & \textbf{89.6\%} & \textbf{36.5\%} \\
		140*140 & 22.1\% & 89.1\% & 35.4\% \\ 
	\end{tabular} 
	\label{tailleIMG}
	 \end{center} 
\end{table}

\noindent For the sequence size, we tested sequences from 15 to 30 images, as the minimum duration of our videos is one second, which, according to current standards, corresponds to 30 \acrshort{FPS}. Table \ref{tailleSeq} shows that the optimal size is 20 images.

\begin{table}[H] 
	\begin{center}
	 \caption{Variation in the size of our sequences} 
	\begin{tabular}{c|c|c|c} 
		Sequence Size & Accuracy & Recall & F1-Score \\ \hline 
		15 & 22.3\% & \textbf{89.8\%} & 35.7\% \\
		20 & \textbf{33\%} & 83.1\% & \textbf{47.3\%} \\ 
		30 & 22.9\% & 89.6\% & 36.5\% \\ 
	\end{tabular}
	 \label{tailleSeq}
	\end{center}
\end{table}

\begin{figure}[H] 
	\begin{center}
		 \includegraphics[scale=0.9]{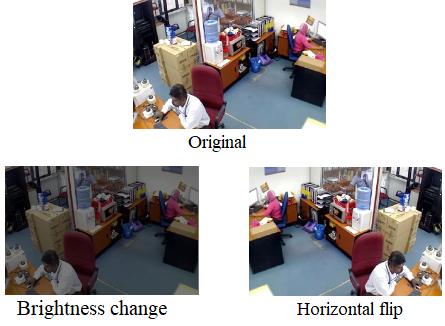} 
		\caption{Data augmentation}
		 \label{dataAug} 
	\end{center} 
\end{figure}

\begin{figure}[H]
	 \begin{center} 
		\includegraphics[scale=0.8]{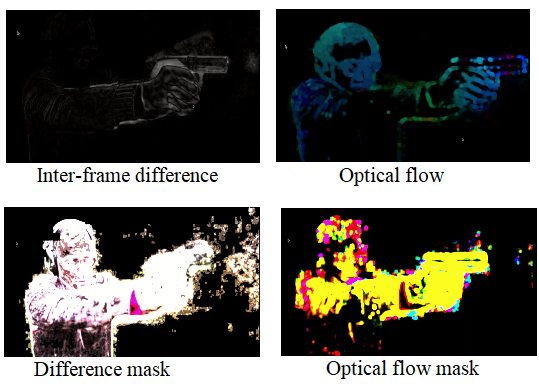} 
		\caption{Preprocessing applied to an image representing a gunshot} 
		\label{preprocess}
	 \end{center}
 \end{figure}

\noindent To evaluate the performance of each of our experiments, we developed our own algorithm to specify the desired step when extracting images. This allows us to evaluate our models under the same conditions as during the training phase, i.e., performing detection per video or detection per sequence (to assess its performance on continuous streams). Our ultimate goal is to assist surveillance operators in their task of monitoring continuous streams, so performance related to sequence detection will be prioritized.
After configuring our sequences, we enhanced our data by applying data augmentation techniques specific to this type of data, such as mirror effects, zooms, and changes in brightness, to enrich our data (see Figure \ref{dataAug}), multiplying it by three while remaining true to realistic scenarios.
Next, we tested various preprocessing techniques, starting with methods commonly used in the field of computer vision: optical flow calculation, inter-image difference, and mask application (see Figure \ref{preprocess}).

\noindent The optical flow does not provide usable results, with accuracy remaining stable (at 50\%) throughout the training. The inter-image difference yields slightly more interesting results, which are improved by data augmentation and degraded by the use of a mask. But to our great surprise, we obtained better results when we limited ourselves to data augmentation alone. The results are summarized in Table \ref{perfTraitement} (except for the first line, all results include data augmentation). Without any preprocessing, the accuracy is poor (and thus the F1-score), while recall is excellent. Data augmentation without further preprocessing gives the best accuracy and a good recall. The inter-image difference degrades the results, with or without a mask.

\begin{table}[H] 
	\begin{center}
		 \caption{Techniques applied to the Fight model} 
		\begin{tabular}{c|c|c|c|c} Technique & Accuracy & Precision & Recall & F1-Score \\ \hline 
		None & \textbf{83.4\%} & 33\% & \textbf{83.1\%} & 47.3\% \\
		Data augmentation & 63.16\% & \textbf{93.62\%} & 60.31\% & \textbf{73.36\%} \\
		Inter-image difference & 59.27\% & 92.19\% & 56.35\% & 69.99\% \\
		Optical Flow & X & X & X & X \\ 
		Mask difference & 23.10\% & 82.24\% & 10.80\% & 19.24\% \\ 
		Mask Optical Flow & X & X & X & X \\
		\end{tabular} 
		\label{perfTraitement} 
	\end{center}
 \end{table}

\begin{figure}[H] 
	\begin{center} 
		\includegraphics[width=\linewidth]{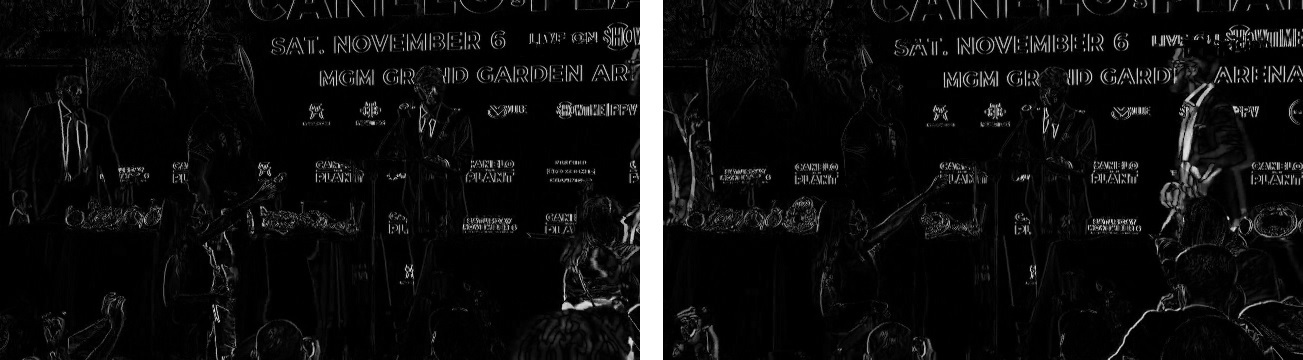} 
		\caption{Images representing a fight} 
		\label{bagarreDiff} 
	\end{center} 
\end{figure}

\noindent These results can be explained by the fact that we have many videos filmed with smartphones, mostly by amateurs. In such videos, the camera is usually mobile, and as a result, the background is also moving. Consequently, the inter-image difference forces the model to extract irrelevant features, as shown in Figure \ref{bagarreDiff}. This figure shows the difference between two successive images extracted from a video representing an interview between two boxers before their fight, where it can be seen that the text on the posters in the background stands out more than the people in the foreground.

\begin{figure}[H] 
	\begin{center} 
		\includegraphics[scale=0.5]{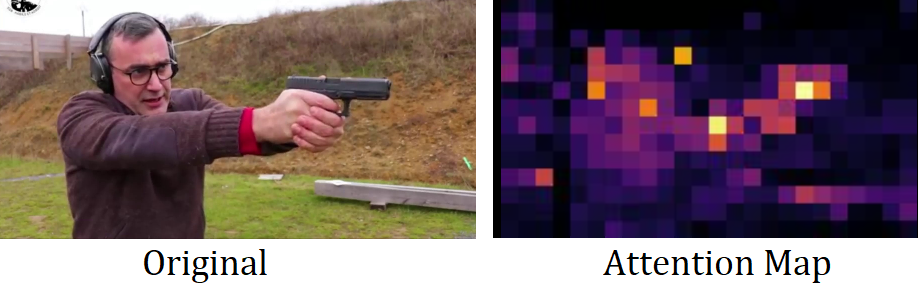} 
		\caption{Attention map for a gunshot video class} 
		\label{dinoAtt} 
	\end{center}
\end{figure}

\noindent We then employed more sophisticated preprocessing methods by submitting our data to preprocessing by specialized models. The results generated were then used for anomaly detection.
Our first idea was to use \gls{DINO} (see Figure \ref{dinoAtt}), presented in the \ref{DINO} section of the state of the art.

\begin{figure}[H] 
	\begin{center} 
		\includegraphics[width=\linewidth]{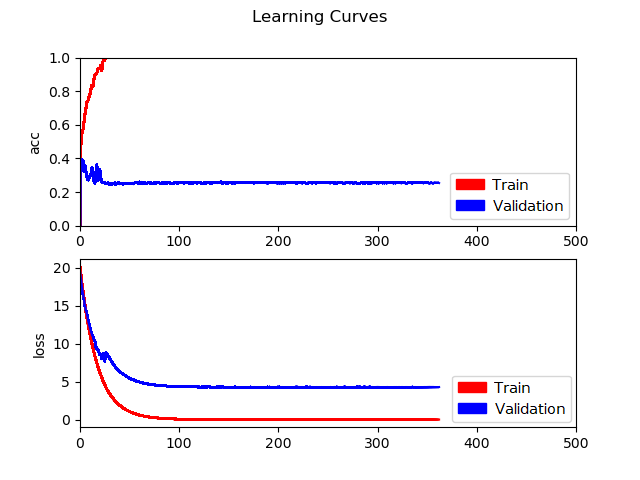} 
		\caption{Overfitting for the Fight class with {DINO} preprocessing} 
		\label{transTrain} 
	\end{center} 
\end{figure}

\noindent After testing DINO on our various datasets, we found that this method requires significant computational time, making it unsuitable for our near-real-time problem.
However, the results obtained on gunshot videos were encouraging, as evidenced by Tables \ref{perfTrans} and \ref{matriceTrans}. The model is able to extract relevant features for this class of events without requiring prior re-training.
Unfortunately, this is not the case for fights, as indicated by the learning curve presented in Figure \ref{transTrain}. These results reaffirm our decision to exclude Vision Transformers from our study in favor of methods better suited to our problem.

\begin{table}[H] 
	\captionof{table}{Performance on the gunshot dataset with {DINO} preprocessing} 
	\label{perfTrans} 
	\begin{center}
		\begin{tabular}{c|c|c|c|c} 
			Class & Accuracy & Precision & Recall & F1-Score \\ \hline 
			Gunshot & 99.7\% & 99.2\% & 96\% & 97.6\% \\
		\end{tabular} 
	\end{center} 
\end{table}

\begin{table}[H]
	\captionof{table}{Confusion matrix on the gunshot dataset with {DINO} preprocessing} 
	\label{matriceTrans} 
	\begin{center} 
		\begin{tabular}{c|cc}
			 \diagbox{Truth}{Predicted} & Gunshot & Normal \\ \hline 
			Gunshot & \textbf{99.96\%} & 0.04\% \\ 
			Normal & 4\% &  \textbf{96\%} \\ 
		\end{tabular} 
	\end{center}
\end{table}

\noindent We also used \acrshort{YOLO} to detect objects and entities in our videos and to estimate the pose of people, with success. This is why we integrated it into the proposed architecture. We detail the results in the next section.

\section{Experiments in Series Mode} 
\label{Serie}

In this section, we will evaluate the performance of our models arranged in series, and show which parameters, techniques, and strategies help improve them. We begin by breaking down the videos by extracting each of their frames and passing them to \acrshort{YOLO} to detect the various objects present in the images and integrate the results of this detection (the bounding boxes around the objects). Then, the video sequence is reassembled to be passed to \acrshort{CGRU} for anomaly detection. We compared the performance of this approach with that of the \acrshort{CGRU} model alone, trained on the same videos. The results can be found in Table \ref{tabPerfSerie}.

\begin{table}[H] 
	\centering 
	\caption{Performance of CGRU vs YOLO+CRU on the gunshot / normal classes} 
	\label{tabPerfSerie} 
	\begin{tabular}{c|c|c|c|c} Model & Accuracy & Precision & Recall & F1-Score \\ \hline 
		\acrshort{CGRU} & 91.89\% & 35.14\% & 84.86\% & 49.70\% \\
		 \acrshort{YOLO} and \acrshort{CGRU} & 91.89\% & 35.14\% & 84.86\% & 49.70\% \\
	 \end{tabular} 
\end{table}

\begin{figure}[H] 
	\centering 
	\includegraphics[scale=0.3]{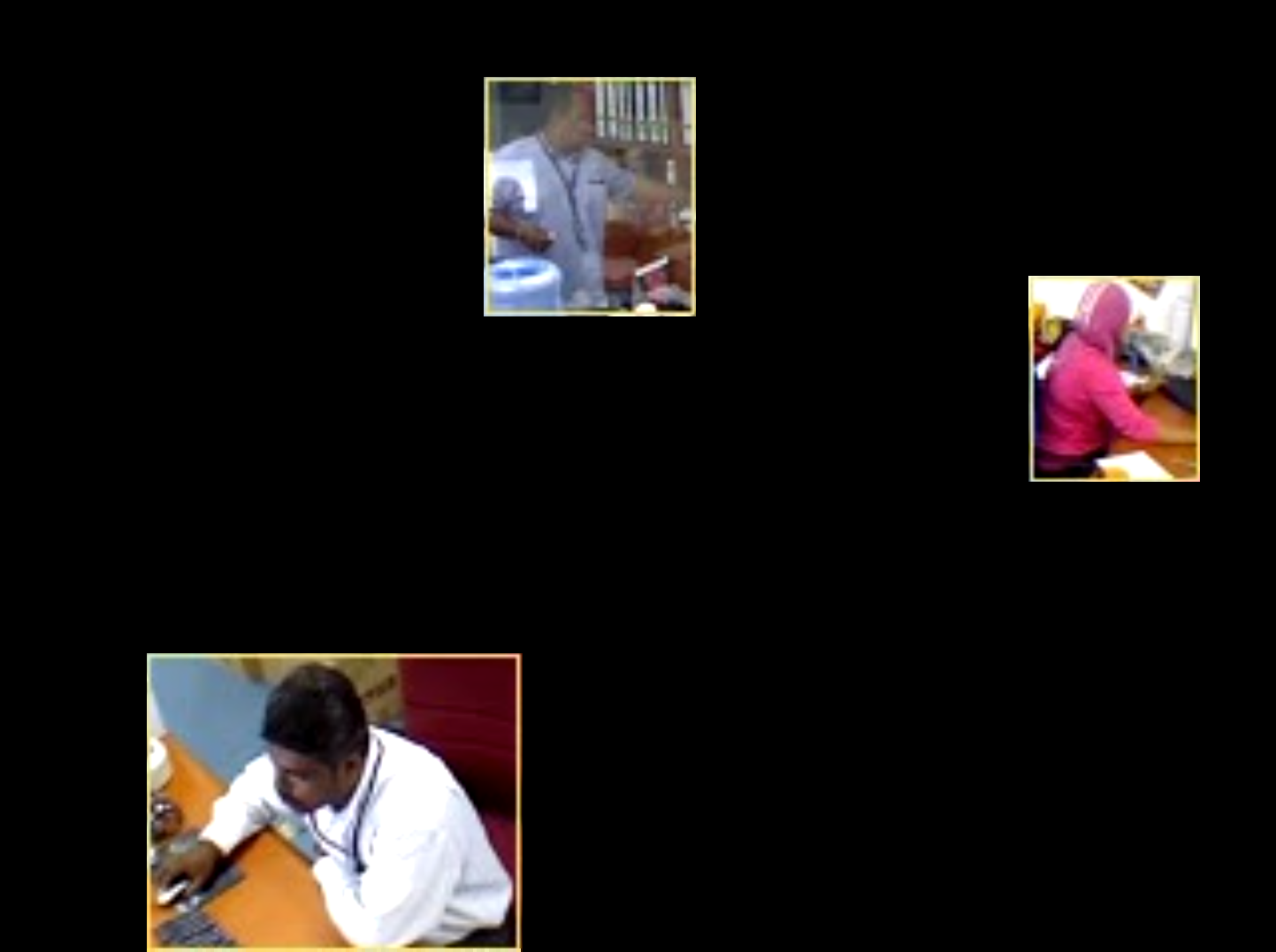}
	 \caption{Mask created with YOLO} 
	\label{yoloMask} 
\end{figure}

\noindent It can be noted that displaying the bounding boxes around objects has no impact on the performance of our model. This means that the model does not take them into consideration during its training. We further enhanced our preprocessing by using these bounding boxes to create masks that keep only the parts of the image with objects, thereby removing as much of the background as possible (Figure \ref{yoloMask}).

\noindent For any image where no object is detected, we had two options: either keep the original image or replace it with a black image. Since this preprocessing is crucial for the reliability of our anomaly detection model, we decided to test both possibilities. Based on the results obtained (Tables \ref{fondNoir}, \ref{fondNoirMatrice}, \ref{fondBlanc}, \ref{fondBlancMatrice}), it seems that the performance is quite similar; however, we observed several differences. Using a black background seems to improve the perception of the normal class at the expense of other classes (fight, fire), which could be explained by the following hypothesis: in some cases, key objects such as flames or people are present, but \acrshort{YOLO} did not detect them, or the detections did not meet the required reliability and overlap thresholds. As a result, this information was lost due to the black background.
Confidence and overlap thresholds are important parameters in object detection using \acrshort{YOLO}, as they determine the level of confidence needed to consider an object as detected correctly. These parameters are manually set before each analysis. In our case, the confidence threshold was set at 55

\begin{table}[H] 
	\begin{center} 
	\caption{Performance for a mask with black background} 
	\label{fondNoir} 
		\begin{tabular}{c|c|c|c} Accuracy & Precision & Recall & F1-Score \\ \hline 
			75.58\% & 79.34\% & 75.58\% & 76.50\% \\ 
		\end{tabular} 
	\end{center}
 \end{table}

\begin{table}[H] 
	\begin{center} 
	\caption{Confusion matrix for a mask with black background} 
	\label{fondNoirMatrice} 
		\begin{tabular}{c|cccc} 
			\diagbox{Truth}{Predicted} & Fight & Gunshot & Fire & Normal \\ 
			\hline Fight & \textbf{55.52\%} & 4.43\% & 1.76\% & 38.29\% \\ 
			Gunshot & 16.20\% &\textbf{41.44\%} & 1.53\% & 40.83\% \\
			Fire & 14.09\% & 12.49\% & \textbf{20.59\%} & 52.83\% \\ 
			Normal & 10.55\% & 2.98\% & 2.09\% & \textbf{84.38\%} \\ 
		\end{tabular} 
	\end{center} 
\end{table}

\begin{table}[H] 
	\begin{center} 
		\caption{Performance for a mask without black background} 
		\label{fondBlanc} 
		\begin{tabular}{c|c|c|c} 
			Accuracy & Precision & Recall & F1-Score \\ \hline 
			72.22\% & 82.37\% & 72.22\% & 75.68\% \\ 
		\end{tabular} 
	\end{center}
 \end{table}

\begin{table}[H] 
	\begin{center} 
		\caption{Confusion matrix in percentage for a mask without black background} 
		\label{fondBlancMatrice} 
		\begin{tabular}{c|cccc} 
			\diagbox{Truth}{Predicted} & Fight & Gunshot & Fire & Normal \\ 
			\hline Fight & \textbf{63.06\%} & 9.33\% & 1.96\% & 25.65\% \\ 
			Gunshot & 25.99\% & \textbf{41.45\%} & 2.29\% & 30.27\% \\
			 Fire & 19.23\% & 15.98\% & \textbf{32.44\%} & 32.35\% \\ 
			Normal & 15.33\% & 4.41\% & 2.07\% & \textbf{78.19\%} \\ 
		\end{tabular} 
	\end{center} 
\end{table}

\begin{figure}[H] 
	\centering \includegraphics[width=\linewidth]{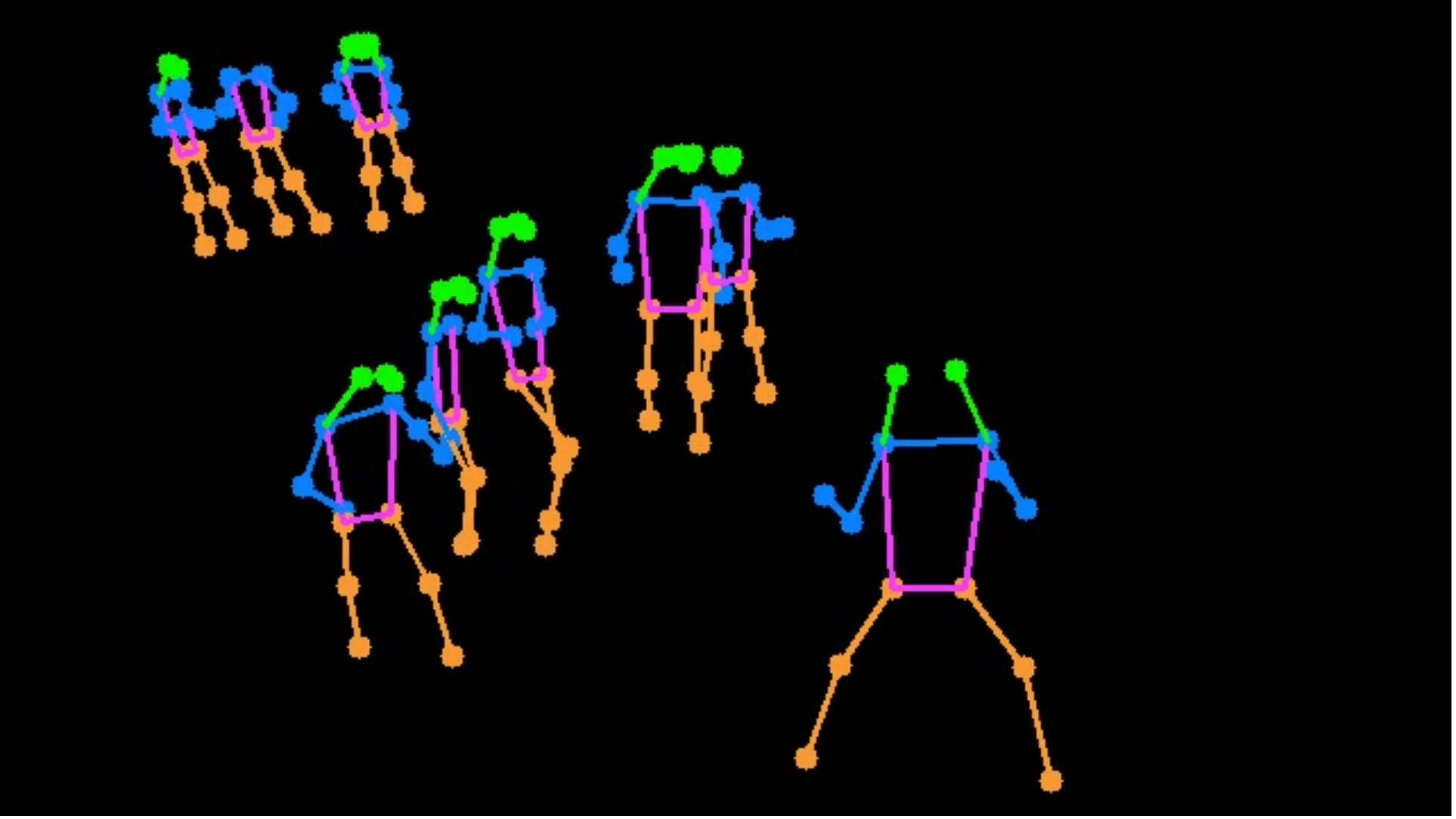} 
	\caption{Pose estimation by \acrshort{YOLO}V7 for a fight, without background}
	 \label{poseNoFond} 
\end{figure}

\noindent In some cases, our anomalies represent specific actions performed by physical persons, actions that we can recognize by the pose that these people take. With version 7 of \acrshort{YOLO}, pose detection can be performed by tracing the ``skeleton`` of each person present on screen, similar to what technologies like OpenPose do (see the blog by \citet{HumanPose}).
We then performed pose estimation during preprocessing to see if it could improve anomaly detection related to behaviors. Initially, since not all our anomalies are related to physical people, we restricted our dataset to the Fight and Gunshot classes. Additionally, we also removed the background from the videos to avoid extracting non-significant features [Figure \ref{poseNoFond}].

\begin{table}[H] 
	\begin{center} 
	\caption{Sequence evaluation (3 classes), without background}
	 \label{PoseFondNoir} 
		\begin{tabular}{c|c|c|c} {Accuracy} & {Precision} & {Recall} & {F1-Score} \\ \hline 
			87.2\% & 90.9\% & 87.8\% & 89\% \\
		 \end{tabular} 
	\end{center} 
\end{table}

\begin{table}[H] 
	\begin{center} 
		\caption{Confusion matrix for sequence evaluation (3 classes), without background} 
		\label{PoseFondNoirMatrice} 
		\begin{tabular}{c|cccc} 
			\diagbox{Truth}{Predicted} & Fight & Gunshot & Normal \\ \hline
			 Fight & \textbf{64.5\%} & 1\% & 34.5\%\\ 
			Gunshot & 11.7\% &\textbf{ 70.7\%} & 17.6\% \\ 
			Normal & 9.5\% & 0\% & \textbf{90.5\%} \\ 
		\end{tabular} 
	\end{center} 
\end{table}

\begin{figure}[H] 
	\centering \includegraphics[width=\linewidth]{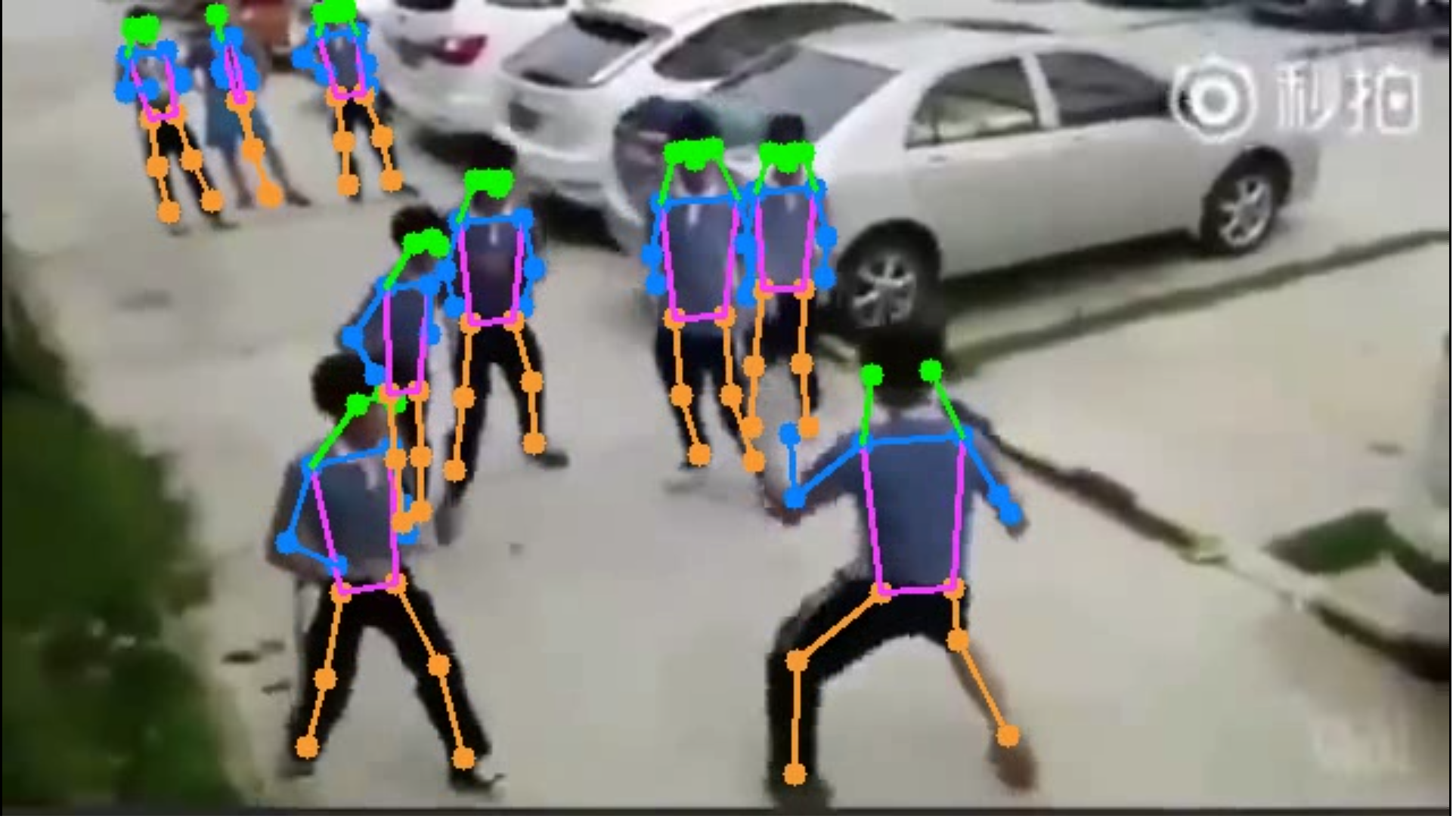} 
	\caption{Pose estimation by \acrshort{YOLO}V7 for a fight, with background} 
	\label{poseFond} 
\end{figure}

\noindent Despite good performance as indicated in Tables \ref{PoseFondNoir} and \ref{PoseFondNoirMatrice}, without the background, it is impossible to detect anomalies that are not related to human behavior. In the next step, we reintroduced the background to our images [Figure \ref{poseFond}, \ref{PoseFondBlancMatrice}] along with all the classes in our dataset to train our model and see if it can detect anomalies such as fires using data with this type of preprocessing [See Table \ref{PoseFondBlanc}, \ref{PoseFondBlancMatrice}].

\begin{table}[H] 
	\begin{center} 
		\caption{Sequence evaluation (4 classes)} 
		\label{PoseFondBlanc} 
		\begin{tabular}{c|c|c|c} {Accuracy} & {Precision} & {Recall} & {F1-Score} \\ \hline
			 87.3\% & 87.6\% & 87.3\% & 87.1\% \\ 
		\end{tabular} 
	\end{center} 
\end{table}

\begin{table}[H] 
	\begin{center} 
		\caption{Confusion matrix in percentage for sequence evaluation (4 classes)} 
		\label{PoseFondBlancMatrice} 
		\begin{tabular}{c|cccc} 
			\diagbox{Truth}{Predicted} & Fight & Gunshot & Fire & Normal \\ \hline
				 Fight & \textbf{60.5\%} & 2.4\% & 1.3\% & 35.8\% \\ 
				Gunshot & 10\% & \textbf{55.6\%} & 14.8\% & 19.6\% \\ 
				Fire & 15.5\% & 10.6\% & \textbf{48\%} & 25.9\% \\ 
				Normal & 3.4\% & 0.6\% & 1\% & \textbf{95\%} \\ 
		\end{tabular}
	 \end{center} 
\end{table}

\noindent In terms of overall performance, it seems that adding the fire class doesn't have a significant impact, but a closer look at the results shows otherwise. The confusion matrices reveal a significant decrease in the reliability of detecting the gunshot class, due to the model labeling some gunshots as fires. Finally, we decided to replace our multi-class model with a normal/abnormal model to assess the impact of \acrshort{YOLO}. We trained two new models: one using images without the background and the other keeping it.

\begin{table}[H] 
	\begin{center} 
		\caption{Normal / abnormal detection without background (with fight and gunshot)} 
		\label{normalAnormalNoFond}
		\begin{tabular}{c|c|c|c} {Accuracy} & {Precision} & {Recall} & {F1-Score} \\ \hline 
			89.18\% & 95.16\% & 92.25\% & 93.83\% \\ 
		\end{tabular} 
	\end{center}
\end{table}

\begin{table}[H] 
	\begin{center} 
		\caption{Confusion matrix for normal / abnormal detection without background \ based on 2 anomalies: fight and gunshot} 
		\label{normalAnormalNoFondMatrice} 
		\begin{tabular}{c|cccc} 
			\diagbox{Truth}{Predicted} & Abnormal & Normal \\ 
			\hline Abnormal & \textbf{64.10\%} & 35.90\% \\ 
			Normal & 7.75\% & \textbf{92.25\%} \\ 
		\end{tabular}
	 \end{center} 
\end{table}

\begin{table}[H] 
	\begin{center} 
		\caption{Normal / abnormal detection with background (with 3 anomalies: fight, gunshot, and fire)} 
		\label{normalAnormalFond} 
		\begin{tabular}{c|c|c|c} {Accuracy} & {Precision} & {Recall} & {F1-Score} \\ \hline 
			84.46\% & 95.52\% & 84.84\% & 89.87\% \\ 
		\end{tabular} 
	\end{center} 
\end{table}

\begin{table}[H] 
	\caption{Confusion matrix for normal / abnormal detection with background \ based on 3 anomalies: fight, gunshot, and fire} 
	\label{normalAnormalFondMatrice} 
		\begin{center} 
			\begin{tabular}{c|cccc} 
				\diagbox{Truth}{Predicted} & Abnormal & Normal \\ \hline 
					Abnormal & \textbf{82.78\%} & 17.22\% \\ 
					Normal & 15.15\% & \textbf{84.85\%} \\ 
			\end{tabular} 
		\end{center} 
\end{table}

\noindent Regardless of the preprocessing used, combining the different types of anomalies improves the performance of our models, with the exception that including anomalies unrelated to the posture of the people present, such as fires, degrades performance.
On the other hand, it can be observed that in multi-class processing cases (tables \ref{normalAnormalNoFond}, \ref{normalAnormalNoFondMatrice}, and \ref{normalAnormalFond}, \ref{normalAnormalFondMatrice}), using \acrshort{YOLO} to preprocess our data significantly improves performance.

\section{Parallel Mode Processing}
\label{Parallele}

Next, we tested our models in parallel mode. The idea behind this approach is to use object detection in parallel with temporal analysis and then combine the results to output a detection, reducing false positive and false negative rates.
Our first experiment was to combine our \acrshort{CGRU} trained on the ``gunshot`` class with \acrshort{YOLO}V4 trained to recognize firearms using a simple rule: if the ``normal`` class is predicted but a firearm is detected with a confidence score exceeding a certain threshold, then we replace the ``normal`` class with the ``gunshot`` class. As our \acrshort{CGRU} is trained to detect fires and \acrshort{YOLO}V4 is trained to recognize flames, we also combined and compared them following the same idea.
For each combination, we evaluated their performance on detections made for each sequence of 20 frames in our videos, using a confidence threshold set at 55\%.

\begin{table}[H] 
	\begin{center} 
		\caption{Performance Related to Fire Detection} 
		\label{paralleleFire} 
		\begin{tabular}{c|c|c|c|c} 
		Model & {Accuracy} & {Precision} & {Recall} & {F1-Score} \\ \hline 
		\acrshort{CGRU} & 86\% & 85.8\% & \textbf{96.2\%} & 90.7\% \\ 
		\acrshort{CGRU} + \acrshort{YOLO} & \textbf{93.2\%} & \textbf{94.6\%} & 95.9\% & \textbf{95.2\%} \\ 
		\end{tabular} 
	\end{center}
	
	\begin{center}
		 \begin{tabular}{c|cc}
			\multicolumn{3}{c}{Confusion Matrix for \acrshort{CGRU}} \\ 
			\diagbox{Truth}{Predicted} & Fire & Normal \\ \hline 
			Fire & \textbf{60.5\%} & 39.5\% \\
			Normal & 3.7 \% & \textbf{96.3\%} \\ 
		\end{tabular} 
	\end{center}
	
	\begin{center} 
		\begin{tabular}{c|cc} 
			\multicolumn{3}{c}{Confusion Matrix for \acrshort{CGRU} + \acrshort{YOLO}} \\ 
			\diagbox{Truth}{Predicted} & Fire & Normal \\ \hline 
			Fire & \textbf{86.4\%} & 13.6\% \\ 
			Normal & 4\% & \textbf{96\%} \\ 
		\end{tabular}
	 \end{center} 
\end{table}

\begin{table}[H] 
	\begin{center} 
		\caption{Performance Related to Gunshot Detection (already presented in Table \ref{tabPerfSerie})} 
		\begin{tabular}{c|c|c|c|c} Model & {Accuracy} & {Precision} & {Recall} & {F1-Score} \\ \hline 
			\acrshort{CGRU} & 91.89\% & 35.14\% & 84.86\% & 49.70\% \\ 
			\acrshort{YOLO} + \acrshort{CGRU} & 91.89\% & 35.14\% & 84.86\% & 49.70\% \\ 
		\end{tabular} 
	\end{center} 
\end{table}

\begin{figure}[H] 
	\centering 
	\includegraphics[width=\linewidth]{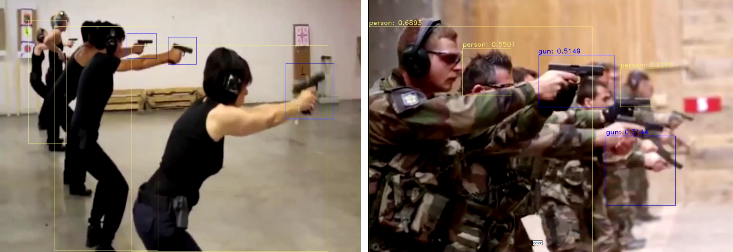} 
	\caption{Result of YOLO V4 trained on the classes gun + person} 
	\label{resultatYolov4GunPerson} 
\end{figure}

\noindent Although the false negatives remain high, as shown in Table \ref{paralleleFire}, there is a clear improvement in fire detection, while for the gunshot class, there is no change. At first glance, it seems that our \acrshort{CGRU} and \acrshort{YOLO} share the same features for gunshot detection. These performances appear to depend on the precision of \acrshort{YOLO} as well as the condition set in the correction component (see \ref{ComposantCorrection}).
We then decided to refine our condition for the gunshot class to reduce incorrect detections. Since \acrshort{YOLO} uses an \gls{IoU} calculation to filter detected objects, we chose to do the same and consider that a gunshot anomaly could only occur if the \gls{IoU} between a weapon and a person was greater than zero. This means that the two bounding boxes must touch, overlap, or one must be contained within the other, as shown in Figure \ref{resultatYolov4GunPerson}.

\noindent Starting from the idea that a firearm only poses a risk when it is being used by a person, we trained a new model on a dataset including firearms and some images from the COCO dataset. For person detection, Table \ref{YoloPerson} shows that this new model outperforms \acrshort{YOLO} version 4. However, for firearm detection, the model trained exclusively on firearms remains the best.

\begin{table}[H] 
	\begin{center} 
		\caption{Detection of the person class on the COCO dataset + our dataset} 
		\label{YoloPerson} 
		\begin{tabular}{c|c|c|c} 
			Model & True Positive & False Positive & False Negative \\ \hline 
			\acrshort{YOLO}v4 & 37.57\% & 12.82\% & 62.43\% \\ 
			\acrshort{YOLO} (gun/person) & \textbf{45.27\%} & \textbf{0.62\%} & \textbf{54.73\%} \\ 
		\end{tabular} 
	\end{center}

	\begin{center} 
		\begin{tabular}{c|cccc} 
			Model & \gls{IoU} & {Precision} & {Recall} & {F1-Score} \\ \hline 
			\acrshort{YOLO}v4 & 60.68\% & 74.56\% & 37.57\% & 49.69\% \\ 
			\acrshort{YOLO} (gun/person) & \textbf{80.76\%} & \textbf{98.66\%} & \textbf{46.32\%} & \textbf{61.18\%} \\ 
		\end{tabular} 
	\end{center} 
\end{table}

\begin{table}[H] 
	\caption{Performance Comparison for Each of Our Models} 
	\begin{tabular}{c|c|c|c|c} 
		Model & {Accuracy} & {Precision} & {Recall} & {F1-Score} \\ \hline 
		\acrshort{CGRU} & 91.89\% & 35.14\% & 84.86\% & 49.70\% \\ 
		\acrshort{CGRU} + \acrshort{YOLO} (gun) & 91.89\% & 35.14\% & 84.86\% & 49.70\% \\ 
		\acrshort{CGRU} + \acrshort{YOLO} (gun/person) & \textbf{92.03\%} & \textbf{36\%} & \textbf{88.73\%} & \textbf{51.15\%} \\ 
	\end{tabular} \\

	\begin{center} 
		\begin{tabular}{c|cc} 
			\multicolumn{3}{c}{Confusion Matrix for \acrshort{CGRU} + \acrshort{YOLO} (Person + Gun)} \\ 
			\diagbox{Truth}{Predicted} & Gunshot & Normal \\ \hline 
			Gunshot & \textbf{88.38\%} & 11.62\% \\
			 Normal & 7.75\% & \textbf{92.25\%} \\ 
		\end{tabular} 
	\end{center} 
	\label{perfModelParalleleReducFauxNeg} 
\end{table}

\noindent Based on the results in Table \ref{perfModelParalleleReducFauxNeg}, coupling the \acrshort{CGRU} with \acrshort{YOLO} by performing an \gls{IoU} calculation between detected persons and firearms allows for maintaining good performance on the normal class while improving the gunshot class by approximately 4\%.
In the same vein, we wanted to see if it was possible to reduce false positives once the two models were combined. We conducted a new evaluation of the models while maintaining the previously mentioned conditions to reduce false negatives, but also false positives. For each fire detection, if no flame is detected by \acrshort{YOLO}, the sequence is considered normal; similarly, for a gunshot, if no firearm is detected, the sequence is considered normal. After evaluation, we observed a significant reduction in the false positive rate. However, this was accompanied by a decrease in the true positive rate, highlighting a lack of precision in \acrshort{YOLO} (Table \ref{perfModelParalleleReducFauxPositif}).

\begin{table}[H] 
	\caption{Performance of the \acrshort{CGRU} + \acrshort{YOLO} Model} 
	\begin{center} 
		\begin{tabular}{c|c|c|c|c} 
			Model & {Accuracy} & {Precision} & {Recall} & {F1-Score} \\ \hline 
			Gunshot & 96.91\% & 95.58\% & 36.39\% & 52.71\% \\ 
			Fire & 91.73\% & 90\% & 98.83\% & 94.24\% \\ 
		\end{tabular} 
	\end{center}

	\begin{center} 
		\begin{tabular}{c|cc} 
			\multicolumn{3}{c}{Confusion Matrix for \acrshort{RCNN} + \acrshort{YOLO} (Reducing False Positives)} \\ 
			\diagbox{Truth}{Predicted} & Gunshot & Normal \\ \hline 
			Gunshot & \textbf{36.40\%} & 63.60\% \\
			Normal & 0.9\% & \textbf{99.1\%} \\ 
		\end{tabular} 
	\end{center}
	
	\begin{center} 
		\begin{tabular}{c|cc} 
			\multicolumn{3}{c}{Confusion Matrix for \acrshort{RCNN} + \acrshort{YOLO} (Reducing False Positives)} \\ 
			\diagbox{Truth}{Predicted} & Fire & Normal \\ \hline 
			Fire & \textbf{72.70\%} & 27.30\% \\
			Normal & 1.16\% & \textbf{98.84\%} \\ 
		\end{tabular}
	 \end{center} 
	\label{perfModelParalleleReducFauxPositif} 
\end{table}

\noindent Since the anomalies we focus on can have severe consequences for future events, we decided to retain the model that minimizes the false negative rate, allowing us to miss as few alerts as possible, even at the risk of generating false alarms. These false alarms can easily be identified by a human.
Moreover, as it is simpler to use a single object detection model for all our anomalies, we also decided to train \acrshort{YOLO} on our three types of objects: persons, firearms, and flames. Finally, to conclude the experimentation with our parallel models, we replaced our multi-class model with a normal/abnormal model to compare the performance of these two architectures.

\begin{table}[H] 
	\caption{Performance of the \acrshort{CGRU} + \acrshort{YOLO} Model}
	 \begin{center} 
		\begin{tabular}{c|c|c|c|c} 
			Model & {Accuracy} & {Precision} & {Recall} & {F1-Score} \\ \hline 
			Gunshot & 96.91\% & 95.58\% & 36.39\% & 52.71\% \\ 
			Fire & 91.73\% & 90\% & 98.83\% & 94.24\% \\ 
		\end{tabular} 
	\end{center}

	\begin{center} 
		\begin{tabular}{c|cc} 
			\multicolumn{3}{c}{Confusion Matrix for \acrshort{RCNN} + \acrshort{YOLO} (Reducing False Positives)} \\ 
			\diagbox{Truth}{Predicted} & Gunshot & Normal \\ \hline 
			Gunshot & \textbf{36.40\%} & 63.60\% \\ 
			Normal & 0.9\% & \textbf{99.1\%} \\ 
		\end{tabular} 
	\end{center}

	\begin{center} 
		\begin{tabular}{c|cc} 
			\multicolumn{3}{c}{Confusion Matrix for \acrshort{RCNN} + \acrshort{YOLO} (Reducing False Positives)} \\ 
			\diagbox{Truth}{Predicted} & Fire & Normal \\ \hline 
			Fire & \textbf{72.70\%} & 27.30\% \\
			Normal & 1.16\% & \textbf{98.84\%} \\ 
		\end{tabular}
	\end{center} 
	\label{perfModelParalleleReducFauxPositif} 
\end{table}

\noindent Since the anomalies we focus on can have severe consequences for future events, we decided to retain the model that minimizes the false negative rate, allowing us to miss as few alerts as possible, even at the risk of generating false alarms. These false alarms can easily be identified by a human.
Moreover, as it is simpler to use a single object detection model for all our anomalies, we also decided to train \acrshort{YOLO} on our three types of objects: persons, firearms, and flames. Finally, to conclude the experimentation with our parallel models, we replaced our multi-class model with a normal/abnormal model to compare the performance of these two architectures.

\begin{figure}[H] 
	\centering 
	\includegraphics[width=\linewidth]{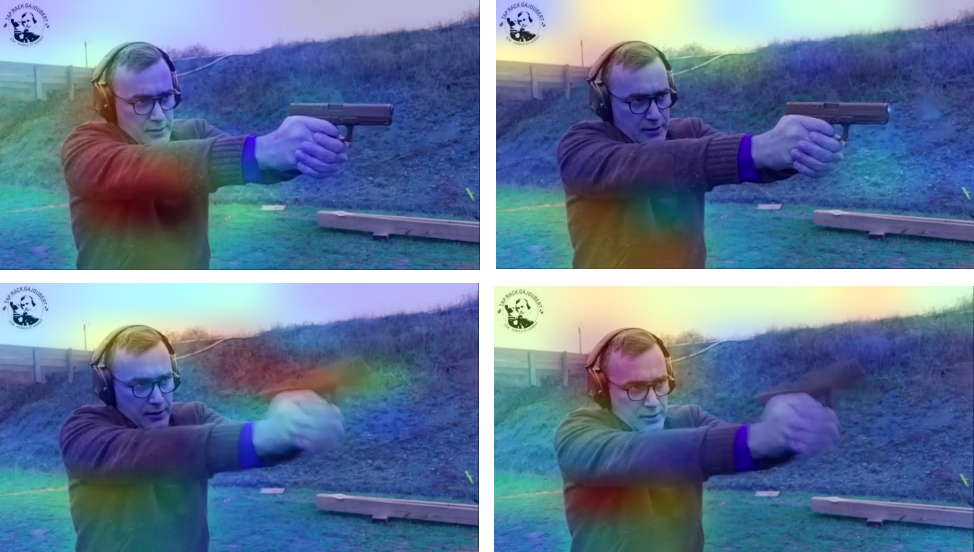} 
	\caption{Activation maps for a video from the gunshot class extracted from the test set} 
	\label{activationMap} 
\end{figure}

\begin{figure}[H] 
	\centering 
	\includegraphics[width=\linewidth]{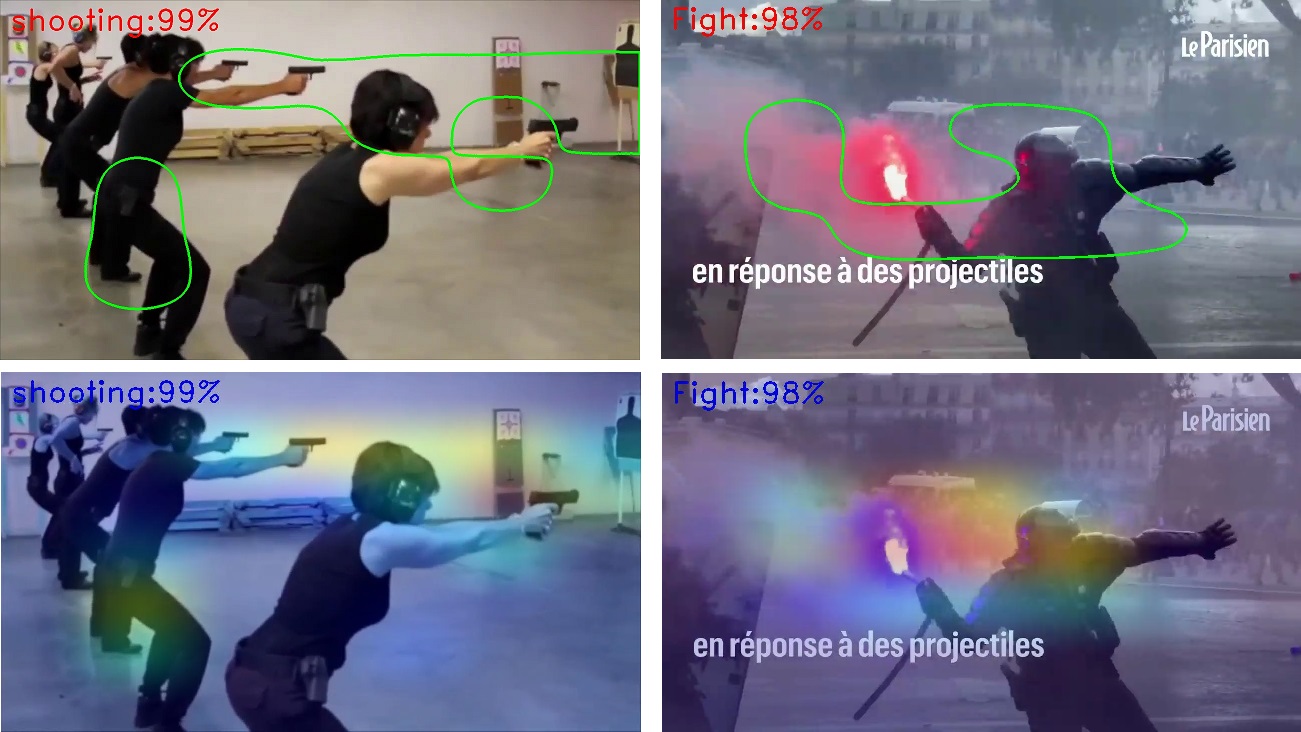} 
	\caption{Example visualization for the classes: gunshot and fight} 
	\label{Contours} 
\end{figure}

\noindent This contour visualization also allowed us to identify low-activation areas that are difficult to perceive in the activation map. However, contour visualization has some drawbacks: contour detection is not very precise, and there may be overlapping contours when a major activation zone is surrounded by a minor one. Furthermore, these contours currently do not indicate the intensity of the activations.

\begin{figure}[H] 
	\centering 
	\includegraphics[width=\linewidth]{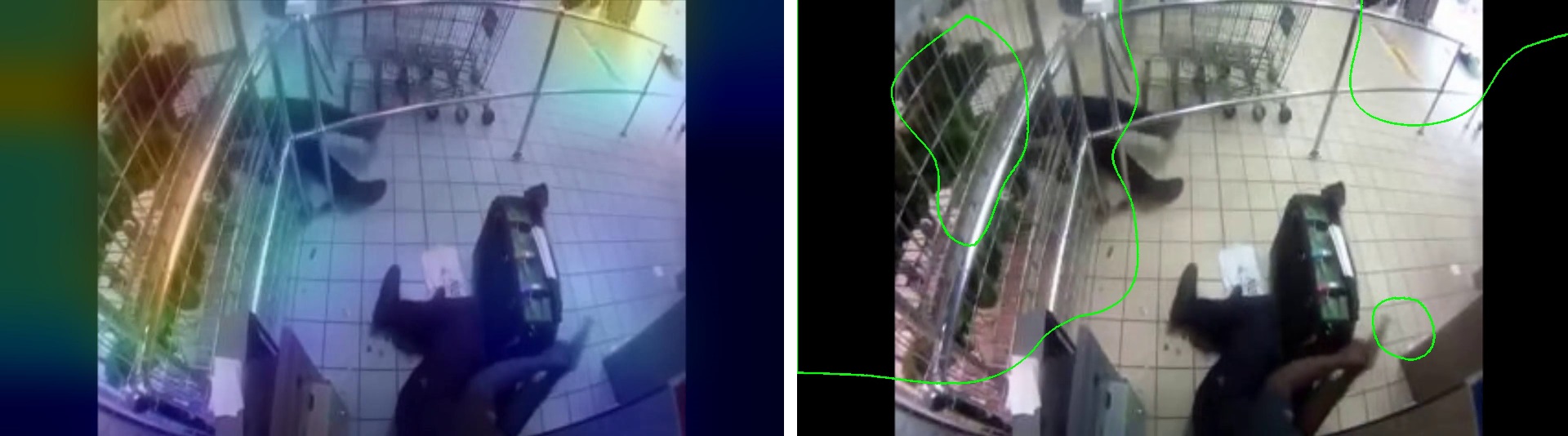} 
	\caption{Activation map and contour visualization on an image from the gunshot class} 
	\label{avantage1} 
\end{figure}

\begin{figure}[H] 
	\centering 
	\includegraphics[width=\linewidth]{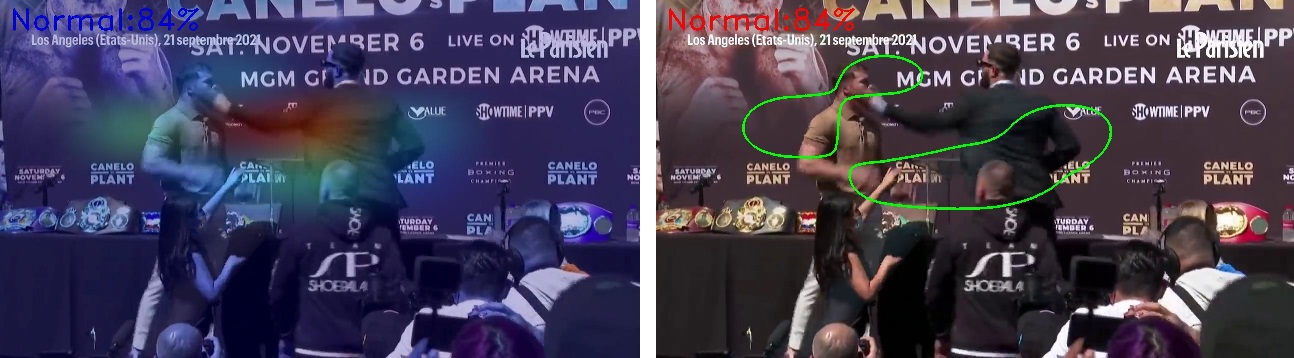} 
	\caption{Activation map and contour visualization on an image from the fight class} 
	\label{avantage2} 
\end{figure}

\begin{figure}[H] 
	\centering 
	\includegraphics[width=\linewidth]{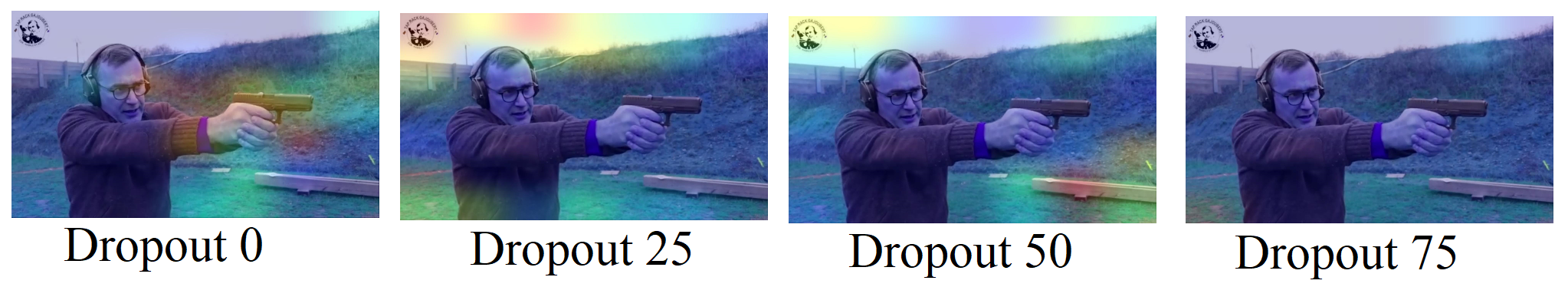} 
	\caption{Activation maps showing the variation of dropout using 64 GRU neurons}
	 \label{dropout} 
\end{figure}

\begin{figure}[H] 
	\centering 
	\includegraphics[width=\linewidth]{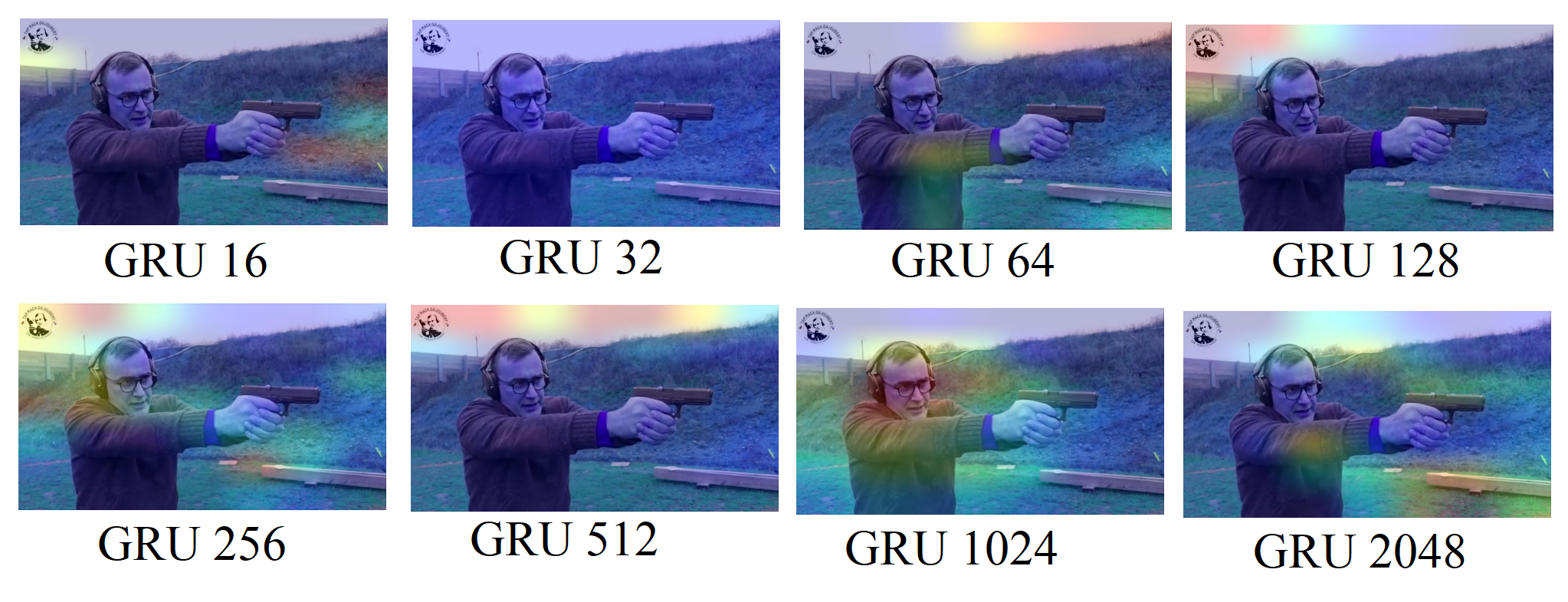} 
	\caption{Activation maps for a variation in the number of GRU neurons using a 50\% dropout} 
	\label{nbNeurone} 
\end{figure}

\begin{figure}[H] 
	\centering 
	\includegraphics[width=\linewidth]{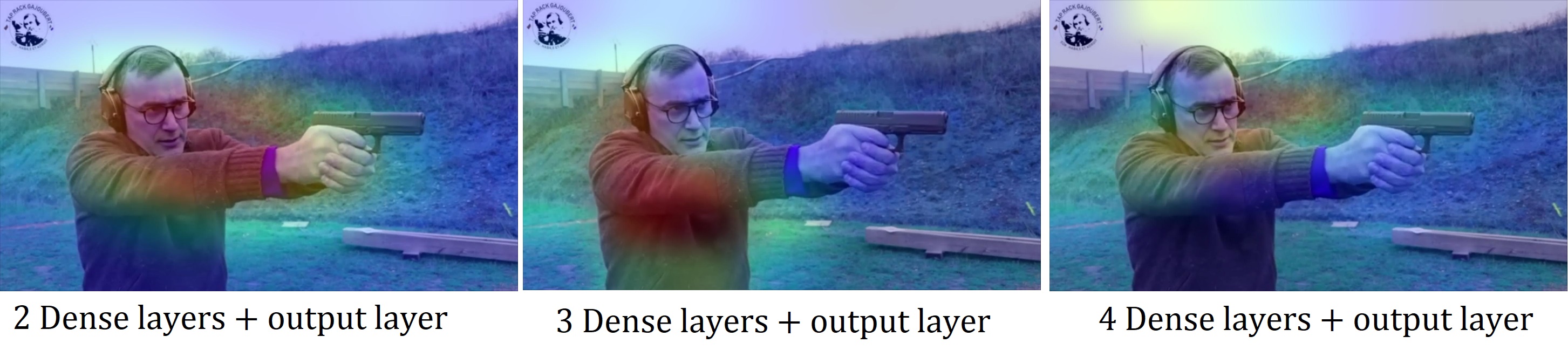} 
	\caption{Activation maps for a variation in the number of classifier layers} 
	\label{coucheDense} 
\end{figure}

\begin{figure}[H] 
	\centering 
	\includegraphics[width=\linewidth]{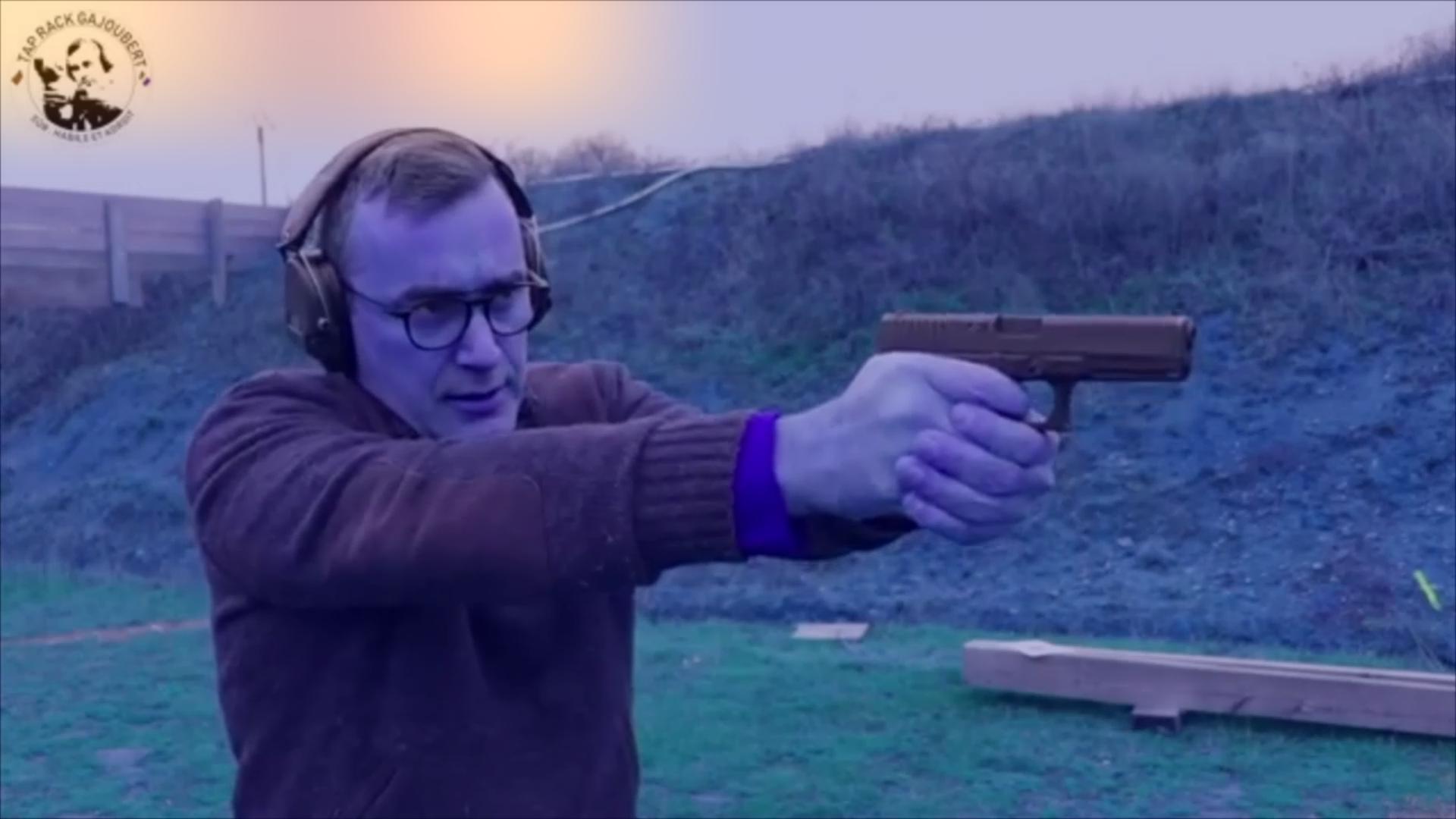} 
	\caption{Activation maps of our model with 1024 GRU neurons and 0\% dropout} 
	\label{bestNoDrop} 
\end{figure}

\noindent The images presented in Figures \ref{avantage1} and \ref{avantage2} perfectly illustrate the advantages and drawbacks of contour visualization. In Figure \ref{avantage1}, the gun is perceived as a low-activation area, hard to spot on the activation map but very clear in the contours. Additionally, a major activation zone surrounded by a minor one can be observed to the left of the image, near the victim. Figure \ref{avantage2} depicts a person striking another. In the activation map on the left, it seems the model recognized the action but predicted it as a normal action instead of a fight. The contour visualization on the right shows that it did not detect the action at all.
The activation maps also allowed us to observe the influence of other layers (\acrshort{RNN}, Dense, Dropout) on the features learned by our convolutional layers due to backpropagation, as well as to detect cases of overfitting. These are visible when the activation zones (colored spots) focus on background elements rather than the essential features of the image (Figure \ref{bestNoDrop}). This insight enabled us to better tune the parameters of these layers, modifying them until they targeted the relevant parts of the images, as shown in Figures \ref{dropout}, \ref{nbNeurone}, and \ref{coucheDense}.

\begin{figure}[H] 
	\centering 
	\includegraphics[width=\linewidth]{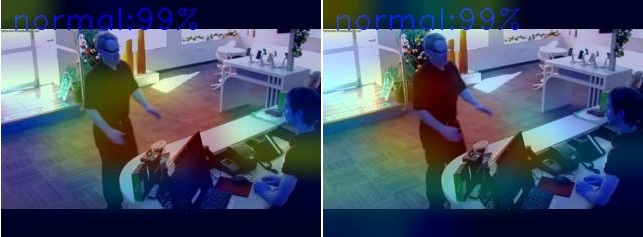} 
	\caption{Visualization of a normal video from the test set representing a rental location.} 
	\label{activationNormal} 
\end{figure}

\begin{figure}[H] 
	\centering 
	\includegraphics[width=\linewidth]{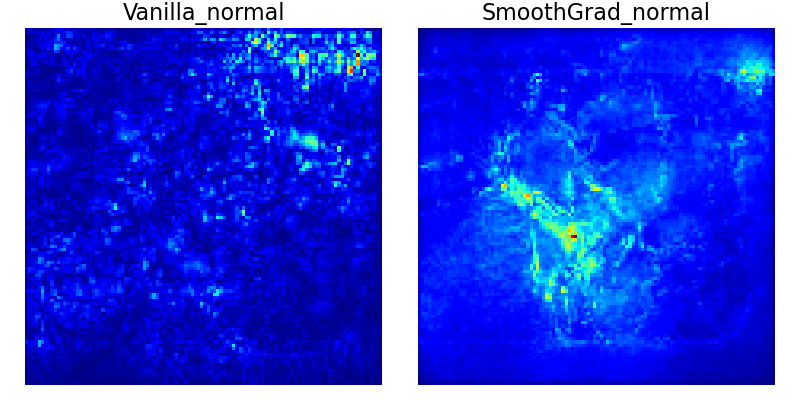} 
	\caption{Saliency map for a normal video.} 
	\label{saliencyNormal} 
\end{figure}

\begin{figure}[H] 
	\centering 
	\includegraphics[width=\linewidth]{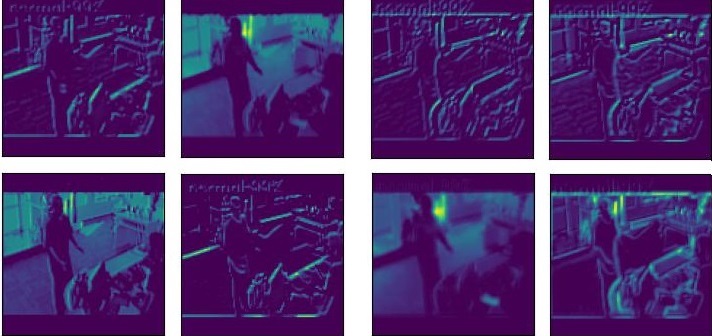} 
	\caption{4 convolution filters for layers 1 and 2.} 
	\label{filter} 
\end{figure}

\noindent Using these two techniques, it is also possible to visualize the characteristics of the normal class. This class includes a wide range of actions such as working, walking, exercising, and more. For this class, the movements are generally slow, unlike the anomalies, which are typically abrupt and fast. For a human observer, the normal class is chosen if none of the characteristics corresponding to an anomaly are found. However, this is not the behavior of our model. To detect the normal class, the model must identify characteristics that represent it. Based on these visualizations, including activation maps, convolution filters, and saliency maps [Figures \ref{activationNormal}, \ref{saliencyNormal}, \ref{filter}], it appears that for our example video, the model relies on the posture and gestures of the individuals in the frame to determine whether or not an anomaly is present.

\section{Summary}

For our near-real-time problem, we conclude by calculating the execution time of these different operational modes. To facilitate comparisons, we use the same videos. The code currently in use has not been optimized due to time constraints, but this does not hinder the comparison. In the case where \acrshort{YOLO} runs in parallel with our \acrshort{CGRU}, the two models are executed sequentially. They are not yet managed in a multithreaded way. Furthermore, since \acrshort{YOLO}V7 is faster than version 4 (used in our initial object detection tests), we use YOLOv7 here.

\begin{table}[H] 
	\begin{center} 
		\caption{Execution time for \acrshort{YOLO} and \acrshort{CGRU} in parallel.} 
		\label{vitesseParallele} 
		\resizebox{\columnwidth}{!}{ 
			\begin{tabular}{c|c|c|c} Video duration & Video FPS & Average detection time & Processing time \\ \hline 
				16s & 33 & 601ms & 15s \\ 
				44s & 30 & 533ms & 35s \\ 
				9s & 30 & 994ms & 12s \\ 
				35s & 30 & 1.1s & 57s \\ 
				23s & 30 & 1.06s & 35s \\ 
				1min 43 & 30 & 758ms & 116s (1min 56) \\ 
				50s & 30 & 826ms & 61s \\ 
				1min 05 & 30 & 886ms & 83s (1min 23) \\ 
				2s & 30 & 847ms & 847ms \\ 
				9s & 30 & 870ms & 11s \\
				2s & 30 & 1s & 1s \\
			\end{tabular}} 
	\end{center}
\end{table}

\begin{table}[H]
	\begin{center} 
		\caption{Execution time for \acrshort{YOLO} and \acrshort{CGRU} in series.} 
		\label{vitesseSerie} 
		\resizebox{\columnwidth}{!}{ 
			\begin{tabular}{c|c|c|c} Video duration & Video FPS & Average detection time & Processing time \\ \hline 
				16s & 33 & 1s & 26s \\ 
				44s & 30 & 1s & 71s (1min 11) \\ 
				9s & 30 & 1.5s & 20s \\ 
				35s & 30 & 1.5s & 81s (1min 21) \\ 
				23s & 30 & 1.5s & 48s \\ 
				1min 43 & 30 & 1.2s & 193s (3min 13) \\
				50s & 30 & 1.3s & 102s (1min 42) \\ 
				1min 05 & 30 & 1.4s & 134s (2min 14) \\ 
				2s & 30 & 1.3s & 1.3s \\ 
				9s & 30 & 1.3s & 17s \\ 
				2s & 30 & 1.5s & 1.5s \\ 
			\end{tabular}} 
	\end{center} 
\end{table}

\noindent As expected, the serial mode is slower than the parallel mode (Tables \ref{vitesseParallele}, \ref{vitesseSerie}). This is because \acrshort{YOLO} needs to prepare each image before the sequence can be analyzed. Consequently, the higher the video's \acrshort{FPS}, the longer the processing time. On the other hand, it can also be observed that our two parallel models are not yet capable of processing videos in real time. Based on the average detection time, it cannot strictly be considered real-time, although the speeds presented in the table are relatively fast.
For the parallel mode, execution time could be further reduced. In addition to managing both models in multithreaded execution, it is also possible to configure the number of images \acrshort{YOLO} needs to analyze for each sequence. The smaller this number, the faster the processing speed. For instance, choosing 1 means every frame will be analyzed, while choosing 2 means every other frame will be analyzed, and so on.

\section{Temporal Analysis: Anomaly Detection} 
\label{sec:AnalyseTemporelle}
\subsection{Model Selection}

To select our temporal analysis model, we tested several architectures derived from well-established models. This approach not only provides access to pre-trained weights but also ensures that we start with architectures that have already proven effective in other, somewhat related domains. Among the models tested were \acrshort{C3D}, convLSTM, and convolutional \acrshort{RNN}s. For these, we compared the effects of \acrshort{LSTM} against those of \acrshort{GRU} and experimented with convolution by selecting \acrshort{CNN}s that have performed well on the IMAGENET dataset (\acrshort{VGG}19, ResNet, Inception, etc.). See Section \ref{part:etatLart}.

We began by testing four variations: 
\begin{enumerate} 
	\item Do not load associated weights. 
	\item Load the weights without retraining the model. 
	\item Load the weights and retrain all layers. 
	\item Load the weights and retrain only a few layers. 
\end{enumerate}

\noindent According to our initial results shown in Table \ref{resultatModeleTemporel}, transfer learning (variations 3 and 4) provided the best performance. Therefore, we decided to continue our experiments using this approach for all our models.

\begin{table}[H]
\caption{Results of our experiments}
\label{resultatModeleTemporel}
\resizebox{\columnwidth}{!}{
	\begin{tabular}{c | c | c | c | c | c}
	Model & Configuration & Accuracy & Precision & Recall & F1-Score \\ \hline
	\acrshort{C3D} & Without weights & 82.6\% & 17.6\% & 73.3\% & 28.5\% \\ \hline

	\acrshort{C3D} & UCF Sport weights & 82.9\% & 21.4\% & \textbf{98.4\%} & 35.2\% \\ 
	& Full block retraining & & & & \\ \hline
	
	\acrshort{C3D} & UCF Sport weights & 83.2\% & 21.8\% & 98.3\% & 35.7\% \\ 
	& Last block retraining & & & & \\ \hline
	
	\acrshort{C3D} + \acrshort{SVM} & UCF Sport weights & 67\% & 13.3\% & 98.00\% & 21.9\% \\ 
	& Last block retraining & & & & \\ \hline
	
	\acrshort{C3D} + ConvLSTM & UCF Sport weights & 83.1\% & 21.6\% & 98.1\% & 35.5\% \\ 
	& Last block retraining & & & & \\ \hline
	
	\acrshort{CNN} + \acrshort{RNN} & \acrshort{VGG}19 without weights & 83.3\% & 19.6\% & 84.2\% & 31.6\% \\ 
	& + \acrshort{GRU} & & & & \\ \hline
	
	\acrshort{CNN} + \acrshort{RNN} & \acrshort{VGG}19 with weights & 83.3\% & 19.9\% & 85.6\% & 32.3\% \\ 
	& IMAGENET + \acrshort{GRU} & & & & \\ \hline
	
	\acrshort{CNN} + \acrshort{RNN} & \acrshort{VGG}19 with weights & \textbf{85.6\%} & \textbf{22.9\%} & 89.6\% & \textbf{36.5\%} \\ 
	& IMAGENET weights, last block retraining & & & & \\ 
	& + \acrshort{GRU} & & & & \\
	\end{tabular} 
}
\end{table}

\noindent According to our experiments, \acrshort{C3D}-type networks have performance levels comparable to those of \acrshort{CNN} + \acrshort{RNN} networks. Despite this, \acrshort{CNN} + \acrshort{RNN} networks demonstrated a better F1-score, indicating a better balance between alert rates and false alarms. As a result, we decided to choose this type of architecture. In addition to achieving a better F1-score on our dataset, these networks offer greater modularity in their architecture, being composed of two separate networks a \acrshort{CNN} and a \acrshort{RNN} that can be easily replaced by another model of the same category. Subsequently, we tested several architectures for our \acrshort{CNN}, always using \acrshort{GRU} for the \acrshort{RNN} part.
For the EfficientNet convolution, we tested several architectures, such as B1, B2, and B7, but none yielded usable results. The performances of the different convolution architectures tested, such as Inception, Inception-ResNet, ResNet, \acrshort{VGG}19, \acrshort{VGG}Place325, and Xception, are listed in Table \ref{testConv}. In this table, we report results only for configurations with some layers retrained.

\begin{table}[H]
\begin{center}
\small
\captionof{table}{Performance of Various Convolutional Networks} 
\label{testConv}
\begin{tabular}{ c | c | c | c | c | c }
Model & Retrained Layer & Accuracy & Precision & Recall & F1-Score \\ \hline
\multirow{2}{*}{Inception} & 0 & 83.4\% & 19.8\% & 82.2\% & 31.9\% \\ 
& 12 & 83\% & 18.1\% & 73.3\% & 29\% \\ \hline

\multirow{4}{*}{Inception-ResNet} & 0 & 79.8\% & 16.8\% & 82.7\% & 28\% \\ 
& 3 & 82.7\% & 18.8\% & 79.9\% & 30.4\% \\ 
& 5 & 82.3\% & 18.1\% & 78.1\% & 29.4\% \\ 
& 12 & 81.9\% & 17.3\% & 76.1\% & 28.2\% \\ \hline

\multirow{6}{*}{ResNet} & 0 & 82.7\% & 19.9\% & 87.1\% & 32.5\% \\ 
& 10 & 89.4\% & 27\% & 73.2\% & 39.5\% \\ 
& 16 & 79.8\% & 15.7\% & 75.3\% & 26\% \\ 
& 19 & 91.2\% & 31\% & 79.6\% & 44.7\% \\ 
& 25 & 85.7\% & 21.4\% & 78.7\% & 33.7\% \\ 
& 31 & 87.1\% & 23.6\% & 77.5\% & 36.2\% \\ \hline

\multirow{7}{*}{\acrshort{VGG}19} & 0 & 83.3\% & 19.9\% & 85.6\% & 32.3\% \\ 
& 3 & 85\% & 20.6\% & 80.4\% & 32.8\% \\ 
& 6 & 85.6\% & 22.9\% & 89.6\% & 36.5\% \\ 
& 9 & 91.6\% & 33\% & 83.1\% & 47.3\% \\ 
& 11 & \textbf{92.2\%} & \textbf{35.1\%} & 84.8\% & \textbf{49.7\%} \\ 
& 13 & 84.6\% & 21.4\% & 86.8\% & 34.4\% \\ 
& 18 & 80.5\% & 17.1\% & 83.5\% & 28.4\% \\ \hline

\multirow{5}{*}{\acrshort{VGG}-Place325} & 0 & 86.9\% & 22.7\% & 79.5\% & 35.3\% \\ 
& 4 & 86.6\% & 23.1\% & 81\% & 36\% \\ 
& 8 & 91\% & 30.9\% & 80.4\% & 44.6\% \\ 
& 10 & 86.5\% & 24.2\% & 86.8\% & 37.9\% \\ 
& 12 & 86.2\% & 22.5\% & 80.8\% & 35.2\% \\ \hline

\multirow{4}{*}{Xception} & 0 & 79.4\% & 18.1\% & \textbf{95.8\%} & 30.5\% \\ 
& 6 & 81.4\% & 17.3\% & 77.9\% & 28.3\% \\ 
& 9 & 87.7\% & 23\% & 75.7\% & 35.3\% \\ 
& 16 & 85.1\% & 21.1\% & 80.2\% & 33.4\% \\ 
\end{tabular}
\end{center}
\end{table}

\noindent Table \ref{recapCNN} summarizes the best performance of each model; the retrained layer number listed corresponds to the one yielding the best results. This table shows that \acrshort{VGG} is the convolutional network with which we obtained the best F1-score.

\begin{table}[H]
\begin{center}
\captionof{table}{Summary} 
\label{recapCNN}
\resizebox{\columnwidth}{!}{
\begin{tabular}{c|c|c|c|c|c}
Model & Retrained Layer & Accuracy & Precision & Recall & F1-Score \\ \hline
Inception & 0 & 83.4\% & 19.8\% & 82.2\% & 31.9\% \\
Inception-ResNet & 3 & 82.7\% & 18.8\% & 79.9\% & 30.4\% \\ 
ResNet & 19 & 91.2\% & 31\% & 79.6\% & 44.7\% \\ 
VGG19 & 11 & \textbf{92.2\%} & \textbf{35.1\%} & \textbf{84.8\%} & \textbf{49.7\%} \\
VGG-PLACE325 & 8 & 91\% & 30.9\% & 80.4\% & 44.6\% \\ 
Xception & 9 & 87.7\% & 23\% & 75.7\% & 35.3\% \\ 
\end{tabular} }
\end{center}
\end{table}

\noindent After selecting \acrshort{VGG}19, we also explored the possibility of removing one convolution to see how it would affect its performance. The results of this experiment are presented in Table \ref{removedConv}, and they clearly show that reducing the number of convolutions is not beneficial.

\begin{table}[H]
\begin{center}
\captionof{table}{Modification of the \acrshort{VGG}19 Architecture}
\label{removedConv}
\begin{tabular}{c|c|c|c|c}
Model & Accuracy & Precision & Recall & F1-Score \\ \hline
\acrshort{VGG}19  & \textbf{92.2\%} & \textbf{35.1\%} & \textbf{84.8\%} & \textbf{49.7\%} \\
\acrshort{VGG}19 (remove last convolution) & 83.6\% & 20.3\% & 84.6\% & 32.8\% \\
\end{tabular}
\end{center}
\end{table}

\noindent We then used this architecture as a base to compare \acrshort{LSTM} and \acrshort{GRU}, as shown in Table \ref{GRUvsLSTM}, which clearly highlights a distinct advantage for \acrshort{GRU}.

\begin{table}[H]
\begin{center}
\captionof{table}{Comparison of Performance Between \acrshort{LSTM} and \acrshort{GRU}}
\label{GRUvsLSTM}
\begin{tabular}{c|c|c|c|c}
Model & Accuracy & Precision & Recall & F1-Score \\ \hline
\acrshort{VGG}19 + \acrshort{GRU} & \textbf{92.2\%} & \textbf{35.1\%} & \textbf{84.8\%} & \textbf{49.7\%} \\ 
\acrshort{VGG}19 + \acrshort{LSTM} & 87.3\% & 24.8\% & 83\% & 38.2\% \\ 
\end{tabular} \newline
\end{center}
\end{table}

\noindent Table \ref{fullyConnect} shows the consequences of choosing the number of fully connected layers needed to train our classifier. In the last row, X represents results that are not usable due to overfitting. The best choice is therefore with three hidden layers.

\begin{table}[H]
\begin{center}
\captionof{table}{Search for the Number of Fully Connected Layers in the Classifier \\ }
\label{fullyConnect}
\begin{tabular}{c|c|c|c|c}
Fully Connected Layer & Accuracy & Precision & Recall & F1-Score \\ \hline
2 + output & 89.8\% & 29.1\% & 84.2\% & 43.2\% \\
3 + output & \textbf{92.2\%} & \textbf{35.1\%} & \textbf{84.8\%} & \textbf{49.7\%} \\
4 + output & 91.2\% & 33\% & 83.1\% & 47.3\% \\ 
5 + output & X\% & X\% & X\% & X\% \\
\end{tabular} \newline
\end{center}
\end{table}

\noindent Finally, we used the results obtained along with the explainability techniques mentioned earlier (section \ref{explicabilite}) to determine the number of neurons required for our \acrshort{GRU} and the dropout rate for our dropout layers.

\subsection{Results}

In this section, we will present our results in anomaly detection. We will first discuss the performance achieved for each of our models. Currently, we have three distinct models for detecting fights, gunshots, and fires. We will evaluate each of them in two aspects: their ability to classify videos (one detection per video) and their ability to detect anomalies in a continuous stream (one detection per sequence). Three tables will be presented for each: one table of metrics and two confusion matrices.

\begin{table}[H]
\captionof{table}{\textbf{Fight Detection Performance}}
\label{bagarrePerf}
\resizebox{\columnwidth}{!}{
\begin{tabular}{c|c|c|c|c|c}
Evaluation Mode & Accuracy & Precision & Recall & F1-Score & Elements \\ \hline
Video Detection & 85.6\% & 86\% & 84.9\% & 85.5\% & 773 \\
Sequence Detection & 63.1\% & 93.6\% & 60.3\% & 73.3\% & 9865 \\
\end{tabular}} \vspace{1\baselineskip}

\begin{center}
\begin{tabular}{c|cc|}
\multicolumn{3}{c}{\shortstack[l]{Confusion Matrix: One Video Detection}} \\
\diagbox{Truth}{Predicted} & Fight & Normal \\ \hline
Fight & \textbf{86.4\%} & 13.6\% \\
Normal & 15.0\% & \textbf{85.0\%} \\
\end{tabular} 
\end{center}\vspace{1\baselineskip}

\begin{center}
\begin{tabular}{c|cc}
\multicolumn{3}{c}{\shortstack[l]{Confusion Matrix: One  Sequence Detection}} \\
\diagbox{Truth}{Predicted} & Fight & Normal \\ \hline
Fight & \textbf{78.3\%} & 21.7\% \\
Normal & 39.8\% & \textbf{60.2\%} \\
\end{tabular}
\end{center}
\end{table}

\begin{table}[H]
\captionof{table}{\textbf{Performance of Gunshot Detection}}
\label{ShootingPerf}
\resizebox{\columnwidth}{!}{
\begin{tabular}{c|c|c|c|c|c}
Evaluation Mode & Accuracy & Precision & Recall & F1-Score & Elements \\ \hline
Video Detection & 86.5\% & 95.3\% & 80.5\% & 87.3\% & 134 \\
Sequence Detection & 91.8\% & 35.1\% & 84.8\% & 49.7\% & 13856 \\
\end{tabular}} \vspace{1\baselineskip}

\begin{center}
\begin{tabular}{c|cc}
\multicolumn{3}{c}{\shortstack[l]{Confusion Matrix: One  Video Detection}} \\
\diagbox{Truth}{Predicted} & Gunshot & Normal \\ \hline
Gunshot & \textbf{80.5\%} & 19.5\% \\
Normal & 5.3\% & \textbf{94.7\%} \\
\end{tabular}
\end{center}

\begin{center}
\begin{tabular}{c|cc}
\multicolumn{3}{c}{\shortstack[l]{Confusion Matrix: One  Sequence Detection}} \\
\diagbox{Truth}{Predicted} & Gunshot & Normal \\ \hline
Gunshot & \textbf{84.8\%} & 15.2\% \\
Normal & 7.7\% & \textbf{92.3\%} \\ 
\end{tabular}
\end{center}
\end{table}

\begin{table}[H]
\captionof{table}{\textbf{Performance of Fire Detection}}
\label{FeuPerf}
\resizebox{\columnwidth}{!}{
\begin{tabular}{c|c|c|c|c|c}
Evaluation Mode & Accuracy & Precision & Recall & F1-Score & Elements \\ \hline
Video Detection & 83.8\% & 70.4\% & 93.9\% & 80.5\% & 93 \\
Sequence Detection & 86.0\% & 85.8\% & 96.2\% & 90.7\% & 9718 \\
\end{tabular}} \vspace{1\baselineskip}

\begin{center}
\begin{tabular}{c|cc}
\multicolumn{3}{c}{\shortstack[l]{Confusion Matrix: One Video Detection}} \\
\diagbox{Truth}{Predicted} & Fire & Normal \\ \hline
Fire & \textbf{78.4\%} & 21.6\% \\
Normal & 6.0\% & \textbf{94.0\%} \\ 
\end{tabular} 
\end{center}

\begin{center}
\begin{tabular}{c|cc}
\multicolumn{3}{c}{\shortstack[l]{Confusion Matrix: One Sequence Detection}} \\
\diagbox{Truth}{Predicted} & Fire & Normal \\ \hline
Fire & \textbf{60.5\%} & 39.5\% \\
Normal & 3.7\% & \textbf{96.3\%} \\ 
\end{tabular} 
\end{center}
\end{table}

\noindent By observing these three tables (\ref{ShootingPerf}, \ref{bagarrePerf}, \ref{FeuPerf}), we can see that the detection of fights (table \ref{bagarrePerf}) is less effective in continuous streams than fire detection (table \ref{FeuPerf}). Based solely on performance, gunshot detection (table \ref{ShootingPerf}) is much less effective when dealing with continuous streams. However, if we look at the confusion matrices, we can observe a slight increase in reliability of about 2\%. In fact, gunshot detection has the best false positive / false negative ratio.
Next, we combined all our datasets to form a single multi-class model and compare its performance with each of the specialized models presented earlier. Our new dataset contains four classes (fire, fight, gunshot, normal) with an imbalanced distribution, as shown in figure \ref{fullDataVideo} from the previous chapter.

\begin{table}[H] 
	\captionof{table}{\textbf{Overall Performance of the Multi-Class Model}} 
	\label{multiClassPerf} 
	\resizebox{\columnwidth}{!}{ 
		\begin{tabular}{c|c|c|c|c|c} 
			Evaluation Mode & Accuracy & Precision & Recall & F1-Score & Elements \\ \hline 
			Video Detection & 81.6\% & 81.4\% & 81.6\% & 81.4\% & 973 \\ 
			Sequence Detection & 74.2\% & 85.8\% & 74.2\% & 78.6\% & 31713 \\ 
		\end{tabular}} 
	\vspace{1\baselineskip}

	\begin{center} 
		\begin{tabular}{c|cccc}
			\multicolumn{5}{c}{Confusion Matrix: One Video Detection} \\ 
			\diagbox{Truth}{Predicted} & Fight & Gunshot & Fire & Normal \\ \hline 
			Fight & \textbf{82.9\%} & 2.5\% & 0.7\% & 13.9\% \\ 
			Gunshot & 10.3\% & \textbf{58.4\%} & 12.9\% & 18.4\% \\ 
			Fire & 8.4\% & 20\% & \textbf{61.6\%} & 10\% \\ 
			Normal & 10.3\% & 0.8\% & 1.5\% & \textbf{87.4\%} \\ 
		\end{tabular} 
	\end{center}
	
	\begin{center} 
		\begin{tabular}{c|cccc} \multicolumn{5}{c}{Confusion Matrix: One Sequence Detection} \\ 
			\diagbox{Truth}{Predicted} & Fight & Gunshot & Fire & Normal \\ \hline 
			Fight & \textbf{64\%} & 6\% & 2\% & 28\%\\ 
			Gunshot & 11.7\% & \textbf{58.6\%} & 10\% & 19.7\% \\
			Fire & 13.7\% & 16.6\% &\textbf{55.7\%} & 14\% \\ 
			Normal & 7.6\% & 5.8\% & 3.9\% & \textbf{82.7\%} \\ 
		\end{tabular} 
	\end{center}
\end{table}

\noindent Despite the fact that there is much more data to process in continuous stream analysis, our model seems to perform quite well and is fairly close to the performance of processing completed videos, as shown in table \ref{multiClassPerf}. To facilitate comparisons of our various models, we evaluated our multi-class model on each portion of our dataset in tables \ref{multiBagarre}, \ref{multiShooting}, and \ref{multiFire}. At first glance, a multi-class model is much more practical to use, but is it as efficient as specialized models ?
For the Fight class, we observe similar performance for video detection and an improvement for sequence detection (continuous stream), with better detection of the normal class, as shown in tables \ref{bagarrePerf}, \ref{multiBagarre}. Regarding the other two classes, Fire (tables \ref{FeuPerf}, \ref{multiFire}) and Gunshot (tables \ref{ShootingPerf}, \ref{multiShooting}), we notice a drop in accuracy of about 10 to 15\% for the Fire class and up to 20\% for the Gunshot class, accompanied by a decrease in F1-Score ranging from 5\% to 10\%.

\begin{table}[H] 
	\captionof{table}{\textbf{Performance on Fights}} 
	\label{multiBagarre} 
	\resizebox{\columnwidth}{!}{ 
		\begin{tabular}{c | c | c | c | c | c}
			 Evaluation Mode & Accuracy & Precision & Recall & F1-Score & Elements \\ \hline 
			Video Detection & 84.9\% & 87.1\% & 84.9\% & 86\% & 773 \\
			Sequence Detection & 84.6\% & 87.8\% & 84.6\% & 86.1\% & 9865 \\ 
		\end{tabular}} 
	\vspace{1\baselineskip}

	\begin{center} 
		\begin{tabular}{c|cccc} \multicolumn{5}{c}{Confusion Matrix: One Video Detection} \\ 
			\diagbox{Truth}{Predicted} & Fight & Gunshot & Fire & Normal \\ \hline 
			Fight & \textbf{82.8\%} & 2.6\% & 0.8\% & 13.8\% \\ 
			Normal & 11.3\% & 0.8\% & 0.8\% & \textbf{87.1\%} \\ 
		\end{tabular} 
	\end{center}
	
	\begin{center} 
		\begin{tabular}{c|cccc} \multicolumn{5}{c}{Confusion Matrix: One Sequence Detection} \\ 
			\diagbox{Truth}{Predicted} & Fight & Gunshot & Fire & Normal \\ \hline 
			Fight & \textbf{64\%} & 6\% & 2\% & 28\% \\
			Normal & 9.9\% & 0.9\% & 0.4\% & \textbf{88.8\%} \\ 
		\end{tabular} 
	\end{center} 
\end{table}

\begin{table}[H] 
	\captionof{table}{\textbf{Performance on Gunshots}} 
	\label{multiShooting} \resizebox{\columnwidth}{!}{ 
		\begin{tabular}{c | c | c | c | c | c} Evaluation Mode & Accuracy & Precision & Recall & F1-Score & Elements \\ \hline 
			Video Detection & 70.8\% & 89.4\% & 70.8\% & 77.2\% & 134 \\ 
			Sequence Detection & 76\% & 95.1\% & 76\% & 83.8\% & 13856 \\ 
		\end{tabular}} 
	\vspace{1\baselineskip}

	\begin{center} 
		\begin{tabular}{c|cccc} \multicolumn{5}{c}{Confusion Matrix: One Video Detection} \\ 
			\diagbox{Truth}{Predicted} & Fight & Gunshot & Fire & Normal \\ \hline 
			Gunshot & 10.4\% & \textbf{58.4\%} & 12.9\% & 18.3\% \\ 
			Normal & 3.6\% & 1.7\% & 7\% & \textbf{87.7\%} \\ 
		\end{tabular} 
	\end{center}
	
	\begin{center} 
		\begin{tabular}{c|cccc} \multicolumn{5}{c}{Confusion Matrix: One Sequence Detection} \\ 
			\diagbox{Truth}{Predicted} & Fight & Gunshot & Fire & Normal \\ \hline 
			Gunshot & 11.6\% & \textbf{58.5\%} & 10\% & 19.9\% \\ 
			Normal & 5\% & 10.5\% & 7.6\% & \textbf{76.9\%} \\ 
		\end{tabular} 
	\end{center}
 \end{table}
	
\begin{table}[H] 
	\captionof{table}{\textbf{Performance on Fires}} 
	\label{multiFire} 
	\resizebox{\columnwidth}{!}{ 
		\begin{tabular}{c | c | c | c | c | c} Evaluation Mode & Accuracy & Precision & Recall & F1-Score & Elements \\ \hline 
			Video Detection & 69.8\% & 87.4\% & 69.8\% & 76.9\% & 93 \\ 
			Sequence Detection & 76.6\% & 86.9\% & 76.6\% & 81.4\% & 9718 \\ 
		\end{tabular}} 
	\vspace{1\baselineskip}
	
	\begin{center} 
		\begin{tabular}{c|cccc} \multicolumn{5}{c}{Confusion Matrix: One Video Detection} \\ 
			\diagbox{Truth}{Predicted} & Fight & Gunshot & Fire & Normal \\ 
			\hline Fire & 8.4\% & 20\% & \textbf{61.6\%} & 10\% \\ 
			Normal & 3\% & 0\% & 12.2\% & \textbf{84.8\%} \\ 
		\end{tabular} 
	\end{center}
	
	\begin{center} 
		\begin{tabular}{c|cccc} \multicolumn{5}{c}{Confusion Matrix: One Sequence Detection} \\ 
			\diagbox{Truth}{Predicted} & Fight & Gunshot & Fire & Normal \\ \hline 
			Fire & 13.7\% & 16.6\% & \textbf{55.7\%} & 14\% \\ 
			Normal & 4\% & 13.2\% & 9.7\% & \textbf{85\%} \\ 
		\end{tabular}
	 \end{center} 
\end{table}

\noindent Despite its good performance, our multi-class model remains inferior to specialized models due to its occasional confusion between different types of incidents. To address this, we decided to train one final version of our model by grouping all anomalies into a single class to simplify analysis. As a result, the dataset now includes a Normal class representing no issues and an Abnormal class indicating an incident requiring intervention, encompassing fights, gunshots, and fires. To facilitate comparisons, as usual, we evaluated this new model on each section of our dataset.

\begin{table}[H] 
	\captionof{table}{\textbf{Performance on the Normal/Abnormal Dataset}}
	\label{normalAbnormal} 
	\resizebox{\columnwidth}{!}{ 
		\begin{tabular}{c | c | c | c | c} 
			Evaluation Mode & Accuracy & Precision & Recall & F1-Score \\ \hline 
			Video Detection & 85.71\% & 82.83\% & 86.74\% & 84.74\% \\ 
			Sequence Detection & 84.27\% & 94.82\% & 85.31\% & 89.82\% \\ 
		\end{tabular}
	} 
	\vspace{1\baselineskip}

	\begin{center} 
		\begin{tabular}{c|cc} 
			\multicolumn{3}{c}{Confusion Matrix: One Video Detection} \\ 
			\diagbox{Truth}{Predicted} & Abnormal & Normal \\ \hline 
			Abnormal & \textbf{84.84\%} & 15.16\% \\ 
			Normal & 13.26\% & \textbf{86.74\%} \\ 
		\end{tabular} 
	\end{center}
	
	\begin{center} 
		\begin{tabular}{c|cc} 
			\multicolumn{3}{c}{Confusion Matrix: One Sequence Detection} \\ 
			\diagbox{Truth}{Predicted} & Abnormal & Normal \\ \hline 
			Abnormal & \textbf{79.72\%} & 20.28\% \\ 
			Normal & 14.68\% & \textbf{85.32\%} \\ 
		\end{tabular} 
	\end{center} 
\end{table}

\begin{table}[H] 
	\captionof{table}{\textbf{Performance on the Fight/Normal Dataset}} 
	\label{normalAbnormalBagarre} 
	\resizebox{\columnwidth}{!}{ 
		\begin{tabular}{c | c | c | c | c} Evaluation Mode & Accuracy & Precision & Recall & F1-Score \\ \hline 
			Video Detection & 84.59\% & 83.66\% & 85.82\% & 84.73\% \\ 
			Sequence Detection & 88.53\% & 94.45\% & 91.65\% & 93.03\% \\ 
		\end{tabular}
	} 
	\vspace{1\baselineskip}

	\begin{center} 
		\begin{tabular}{c|cc} \multicolumn{3}{c}{Confusion Matrix: One Video Detection} \\ 
			\diagbox{Truth}{Predicted} & Fight & Normal \\ \hline 
			Fight & \textbf{83.37\%} & 16.63\% \\ 
			Normal & 14.17\% & \textbf{85.83\%} \\ 
		\end{tabular} 
	\end{center}
	
	\begin{center} 
		\begin{tabular}{c|cc} 
			\multicolumn{3}{c}{Confusion Matrix: One Sequence Detection} \\ 
			\diagbox{Truth}{Predicted} & Fight & Normal \\ \hline 
			Fight & \textbf{72.77\%} & 27.23\% \\ 
			Normal & 8.35\% & \textbf{91.65\%} \\ 
		\end{tabular} 
	\end{center} 
\end{table}

\begin{table}[H]
\captionof{table}{\textbf{Performance on the Shooting/Normal Dataset}}
\label{normalAbnormalShooting}
\resizebox{\columnwidth}{!}{
\begin{tabular}{c | c | c | c | c }
Evaluation Mode & Accuracy & Precision & Recall & F1-Score \\ \hline
Video Detection & 88.80\% & 82.81\% & 92.98\% & 87.60\% \\ 
Sequence Detection & 79.74\% & 99.34\% & 79.26\% & 88.17\% \\ 
\end{tabular}} \vspace{1\baselineskip}

\begin{center}
\begin{tabular}{c|cc}
\multicolumn{3}{c}{Confusion Matrix: Video Detection} \\ 
 \diagbox{Truth}{Predicted} & Shooting & Normal \\ \hline
Shooting & \textbf{85.72\%} & 14.28\% \\ 
Normal & 7.02\% & \textbf{92.98\%} \\ 
\end{tabular}
\end{center}

\begin{center}
\begin{tabular}{c|cc}
\multicolumn{3}{c}{Confusion Matrix: Sequence Detection} \\ 
 \diagbox{Truth}{Predicted} & Shooting & Normal \\ \hline
Shooting & \textbf{89.44\%} & 10.56\% \\ 
Normal & 20.73\% & \textbf{79.27\%} \\ 
\end{tabular} 
\end{center}
\end{table}

\begin{table}[H]
\captionof{table}{\textbf{Performance on the Fire/Normal Dataset}}
\label{normalAbnormalFire}
\resizebox{\columnwidth}{!}{
\begin{tabular}{c | c | c | c | c}
Evaluation Mode & Accuracy & Precision & Recall & F1-Score \\ \hline
Video Detection & 91.39\% & 87.87\% & 87.87\% & 87.87\% \\ 
Sequence Detection & 88.21\% & 93.24\% & 90.03\% & 91.61\% \\ 
\end{tabular}} \vspace{1\baselineskip}

\begin{center}
\begin{tabular}{c|cc}
\multicolumn{3}{c}{Confusion Matrix: Video Detection} \\ 
 \diagbox{Truth}{Predicted} & Fire & Normal \\ \hline
Fire & \textbf{93.33\%} & 6.67\% \\ 
Normal & 12.12\% & \textbf{87.88\%} \\ 
\end{tabular}
\end{center}

\begin{center}
\begin{tabular}{c|cc}
\multicolumn{3}{c}{Confusion Matrix: Sequence Detection} \\ 
 \diagbox{Truth}{Predicted} & Fire & Normal \\ \hline
Fire & \textbf{83.67\%} & 16.33\% \\ 
Normal & 9.96\% & \textbf{90.04\%} \\ 
\end{tabular} 
\end{center}
\end{table}

\begin{figure}[H]
\includegraphics[width=\linewidth]{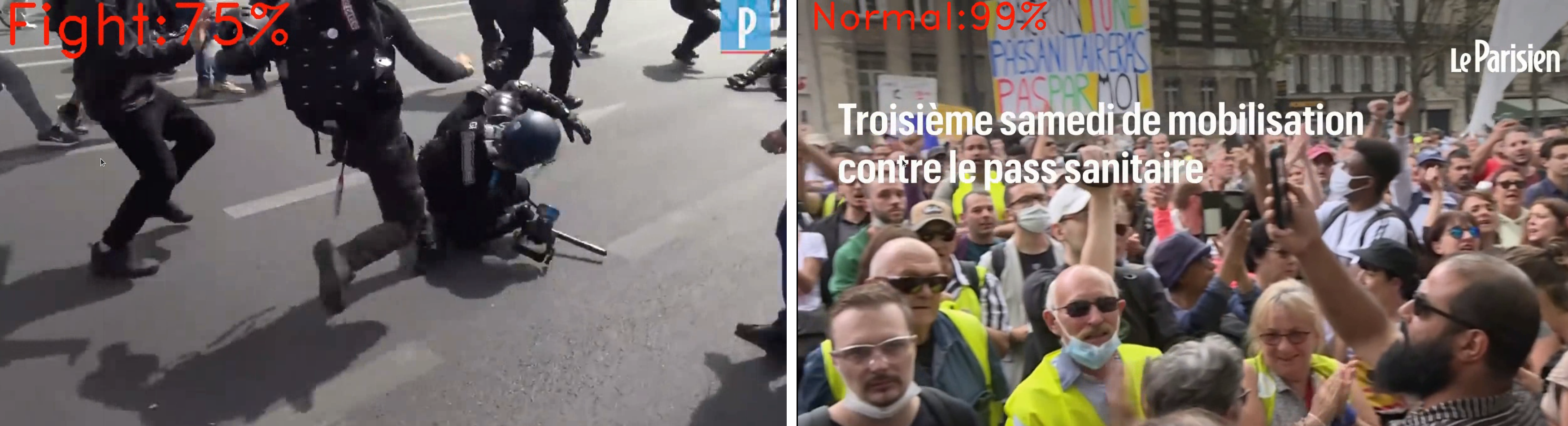}
\caption{Fight detection, Source: Le Parisien}
\label{bagarreExemple}
\end{figure}

\begin{figure}[H]
\includegraphics[width=\linewidth]{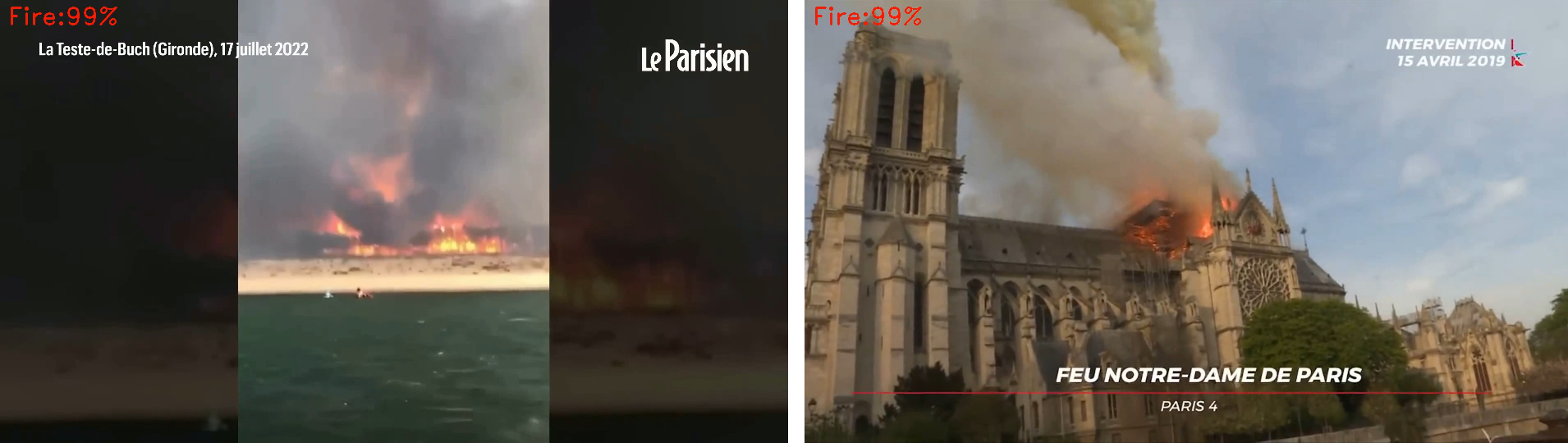}
\caption{Fire detection, Source: Le Parisien}
\label{incendieExemple}
\end{figure}

\begin{figure}[H]
\includegraphics[width=\linewidth]{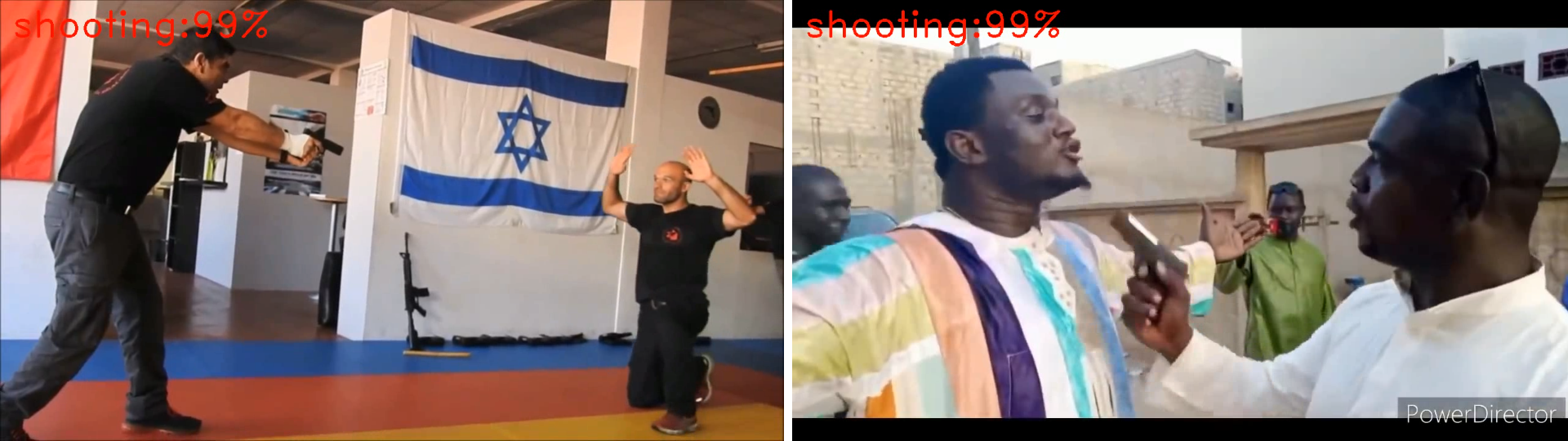}
\caption{Gunshot detection}
\label{shootingExemple}
\end{figure}

\noindent When comparing these statistics (\ref{multiClassPerf}, \ref{normalAbnormal}), we observe that this new two-class model (normal/abnormal) outperforms our multi-class model. Moreover, it is generally equivalent to or better than our specialized models (tables \ref{multiBagarre} vs. \ref{normalAbnormalBagarre}, \ref{multiShooting} vs. \ref{normalAbnormalShooting}, \ref{multiFire} vs. \ref{normalAbnormalFire}). This suggests that the multi-class model correctly eliminates the normal class but sometimes misclassifies the type of anomaly.
In scenarios where only one type of anomaly is of interest, a specialized model should be preferred. However, when handling multiple types of anomalies, the choice of model depends on the problem at hand. If we aim to simply detect any type of incident and allow a human to deduce its specific nature, a normal/abnormal model is a better candidate. Conversely, if we need to precisely identify the type of anomaly, we must use a multi-class model. Although it is less effective than the binary model for handling multiple types of anomalies, the multi-class model remains viable for real-world conditions.
To evaluate the performance of our multi-class model in real-world scenarios, we tested it on a set of ten unedited videos not included in our dataset. The images below are extracted from these videos and illustrate the model's detections along with their confidence scores, displayed in red in the top-left corner of each image.
For the ``fight`` class, we selected images from anti-health-pass protests that involved violent incidents (figure \ref{bagarreExemple}). For the ``fire`` class, we used footage from the Notre-Dame Cathedral fire and a wildfire in the Gironde region in July 2022 (figure \ref{incendieExemple}). As there were no recent incidents involving gunshots, we tested our model on generic videos retrieved online (figure \ref{shootingExemple}).

\noindent Table \ref{perfVitesse} indicates the execution speed of our multi-class model on the ten unedited videos described above. Although our code is not optimized, the detection time is generally respectable, ranging between 104 and 744 milliseconds.

\begin{table}[H]
\begin{center}
\caption{Execution Speed of Our Model}
\label{perfVitesse}
\resizebox{\columnwidth}{!}{
\begin{tabular}{ c | c | c | c }
Video Duration & Video \acrshort{FPS} & Average Detection Time & Total Processing Time \\ \hline
34s & 30 & 468ms & 23s \\ 
20s & 30 & 500ms & 13s \\ 
2min 20 & 25 & 263ms & 46s \\ 
4min 56 & 30 & 104ms & 46s \\ 
2min 06 & 25 & 128ms & 20s \\ 
4min 36 & 30 & 122ms & 50s \\ 
20s & 30 & 285ms & 8s \\ 
1min 30 & 30 & 248ms & 33s \\ 
1min 12 & 30 & 140ms & 15s \\ 
3min 08 & 30 & 744ms & 1min 56 \\ 
\end{tabular}}
\end{center}
\end{table}

\section{Spatial Analysis: Object Detection}
\label{sec:AnalyseSpatiale}
\subsection{Model Selection}

In this section, we present our experiments and results in object detection.
Our first experiment involved training the \acrshort{YOLO}V3 and \acrshort{YOLO}V4 architectures on a subset of our dataset containing only two classes: “gun” and “weapons,” to compare their performance.
To achieve this, we developed our own evaluation algorithm. This algorithm starts by loading the ground truth labels associated with the predicted image. Then, it compares the ground truth boxes for each class with all remaining detections for that class, filtering them using the \gls{NMS} operation to find the one with the highest \gls{IoU}. For reference, the \gls{NMS} operation filters overlapping bounding boxes for the same object, retaining only those with the highest confidence scores. Once this bounding box is identified, we calculate the \gls{IoU} to measure the similarity percentage between the detected box and the ground truth box.
Our algorithm is therefore not efficient, resulting in slow execution. However, during the evaluation phase, processing speed is not a critical issue. We added certain parameters to our system, such as the overlap rate and confidence threshold used in the \gls{NMS} operation, as well as the ability to set minimum and maximum thresholds for the \gls{IoU} and a minimum required size for predicted boxes. Any detection not meeting the parameters set for \gls{NMS} is considered invalid and excluded from the statistics. Regarding the parameters for calculating the \gls{IoU}, we introduced a new metric called badBox, representing a correct detection (True Positive) where the bounding box does not meet the specified criteria. These parameters allow us to assess the performance of our model under real-world conditions.

\begin{table}[H]
\begin{center}
\caption{\acrshort{YOLO}V3: Confidence 55\%, Threshold 70\%, IoU Min 0\%, Max 100\%, Min Box Size 0px}
\label{yolov3Perf}
\begin{tabular}{c | c | c | c | c | c}
Class & Detected & Total & True Positive & False Positive & False Negative \\ \hline
Gun & 3285 & 3041 & 72.38\% & 4.93\% & 27.62\% \\ 
Weapons & 577 & 1308 & 31.19\% & 3.98\% & 68.81\% \\ 
\end{tabular}
\end{center}
\hfill
\begin{center}
\begin{tabular}{c | c | c | c | c}
Class & \gls{IoU} & Precision & Recall & F1-Score \\ \hline
Gun & 79.04\% & 93.62\% & 72.38\% & 81.64\% \\ 
Weapons & 70.05\% & 88.7\% & 31.19\% & 46.15\% \\ 
\end{tabular}
\end{center}
\hfill\null
\end{table}

\begin{table}[H]
\begin{center}
\caption{\acrshort{YOLO}V4: Confidence 55\%, Threshold 70\%, IoU Min 0\%, Max 100\%, Min Box Size 0px}
\label{yolov4Perf}
\begin{tabular}{c | c | c | c | c | c}
Class & Detected & Total & True Positive & False Positive & False Negative \\ \hline
Gun & 4929 & 3041 & 79.61\% & 6.94\% & 20.39\% \\ 
Weapons & 1633 & 1308 & 52.14\% & 5.58\% & 47.68\% \\ 
\end{tabular}
\end{center}
\hfill
\begin{center}
\begin{tabular}{c | c | c | c | c}
Class & \gls{IoU} & Precision & Recall & F1-Score \\ \hline
Gun & 82.22\% & 91.98\% & 79.61\% & 85.35\% \\ 
Weapons & 75.83\% & 90.33\% & 52.14\% & 66.12\% \\ 
\end{tabular}
\end{center}
\end{table}

\noindent A comparison of Tables \ref{yolov3Perf} and \ref{yolov4Perf} reveals that version 4 of \acrshort{YOLO} outperforms version 3, even though it is slightly slower (processing times remain acceptable overall). We therefore selected version 4 for subsequent experiments.
By analyzing our mislabeled files, we discovered that \acrshort{YOLO}V4 occasionally confused the two classes, leading to reduced performance during evaluation. This can be attributed to the high similarity between the images of the two classes. In our context, any firearm, regardless of its type, represents a risk. Consequently, we decided to merge these two classes to prevent performance degradation.

\begin{table}[H]
\begin{center}
\caption{Comparison of \acrshort{YOLO}V4 Performance (1 Class) with \acrshort{YOLO}V4 (2 Classes)}
\label{yolo1classevs2}
\begin{tabular}{c | c | c | c}
Model & True Positive & False Positive & False Negative \\ \hline
YoloV4: 2 Classes & \textbf{65.88\%} & \textbf{6.26\%} & \textbf{34.04\%} \\ 
YoloV4: 1 Class & 65.05\% & 18.44\% & 34.94\% \\ 
\end{tabular}
\end{center}
 
\begin{center}
\begin{tabular}{c | c | c | c | c}
Model & \gls{IoU} & {Precision} & {Recall} & {F1-Score} \\ \hline
YoloV4: 2 Classes & \textbf{79.03\%} & \textbf{91.16\%} & \textbf{65.88\%} & \textbf{75.74\%} \\ 
YoloV4: 1 Class & 70.42\% & 77.91\% & 65.05\% & 70.90\% \\ 
\end{tabular}

\end{center}
\end{table}

\begin{figure}[H]
\centering
\includegraphics[width=\linewidth, height=8cm]{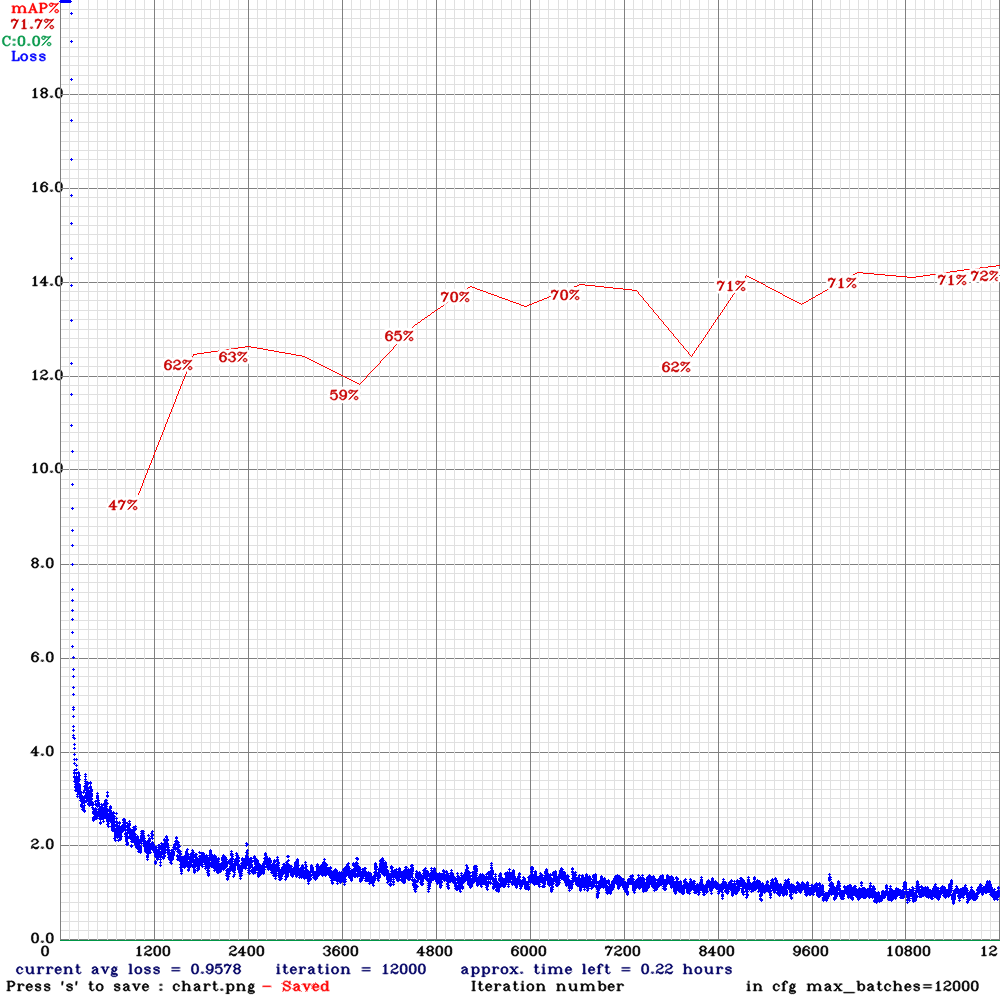}
\caption{Learning Curve for YOLOV4: 2 Classes}
\label{learningCurveYoloV4-2}
\end{figure}

\begin{figure}[H]
\centering
\includegraphics[width=\linewidth, height=8cm]{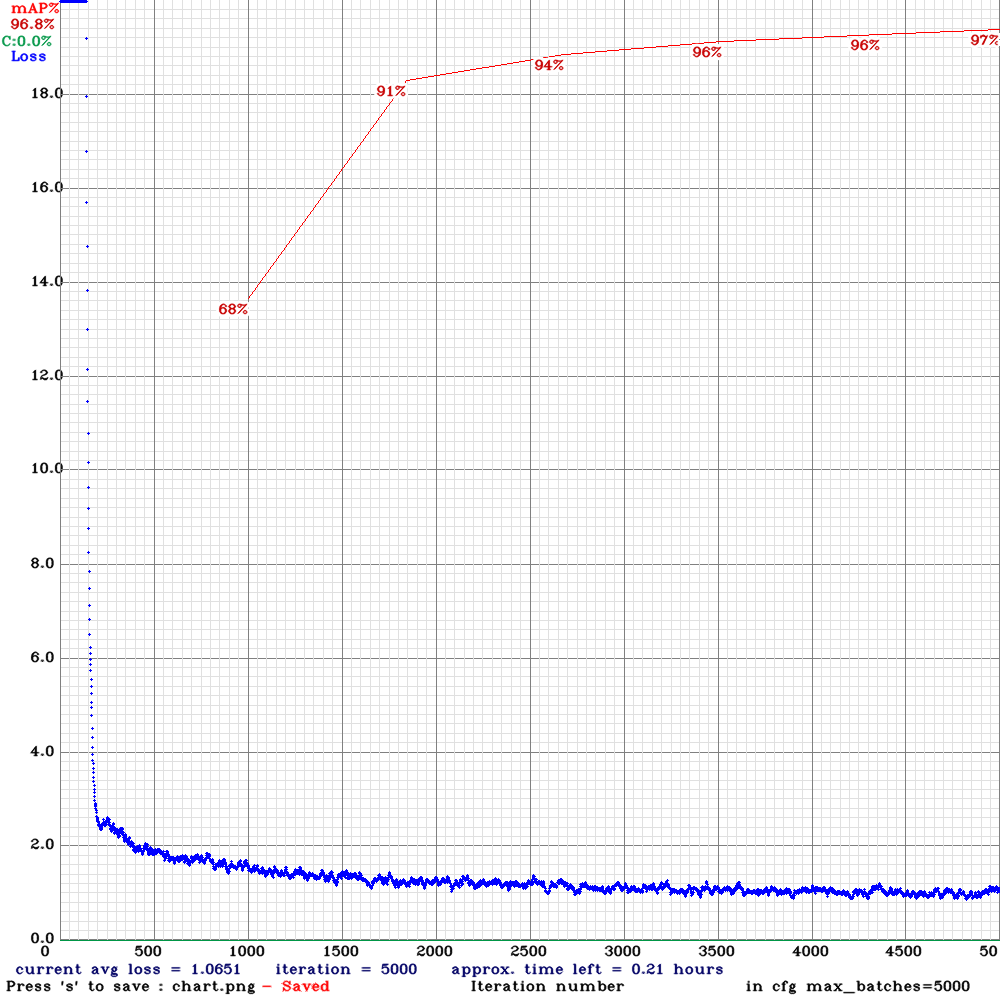}
\caption{Learning Curve for YOLOV4: 1 Class}
\label{learningCurveYoloV4-1}
\end{figure}

\noindent At first glance, the section of Table \ref{yolo1classevs2} representing the merged classes appears to show lower performance. However, based on the learning curve generated during training, the \gls{MAP} is significantly better (Figures \ref{learningCurveYoloV4-2}, \ref{learningCurveYoloV4-1}).

\subsection{Results}

\begin{figure}[H]
\centering
\includegraphics[width=\linewidth]{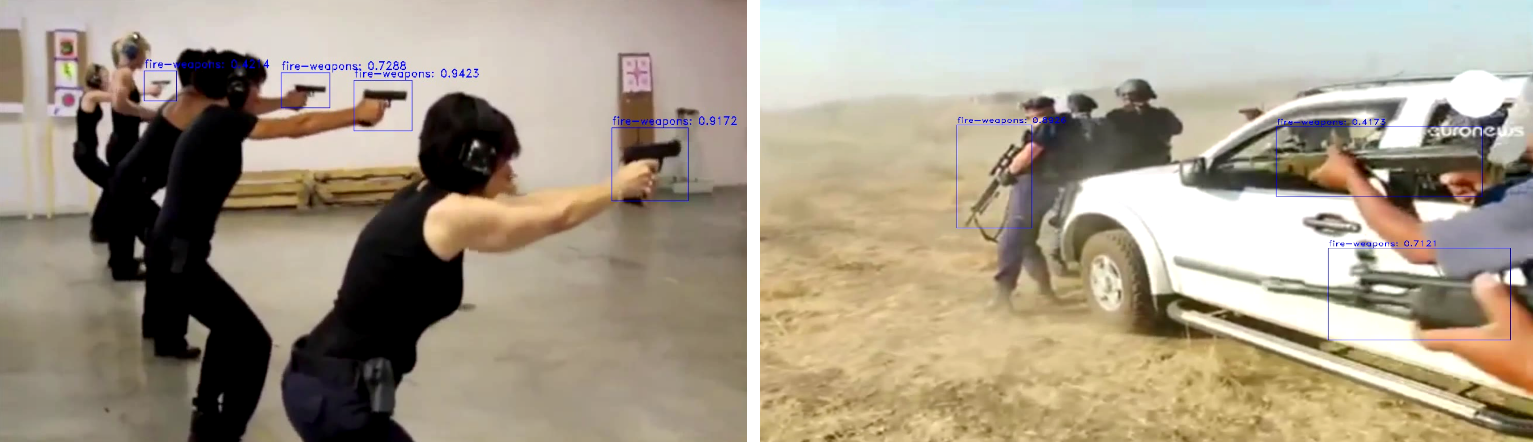}
\caption{Gun Detection via \acrshort{YOLO}V4}
\label{yoloGunExemple}
\end{figure}

\begin{figure}[H]
\centering
\includegraphics[width=\linewidth]{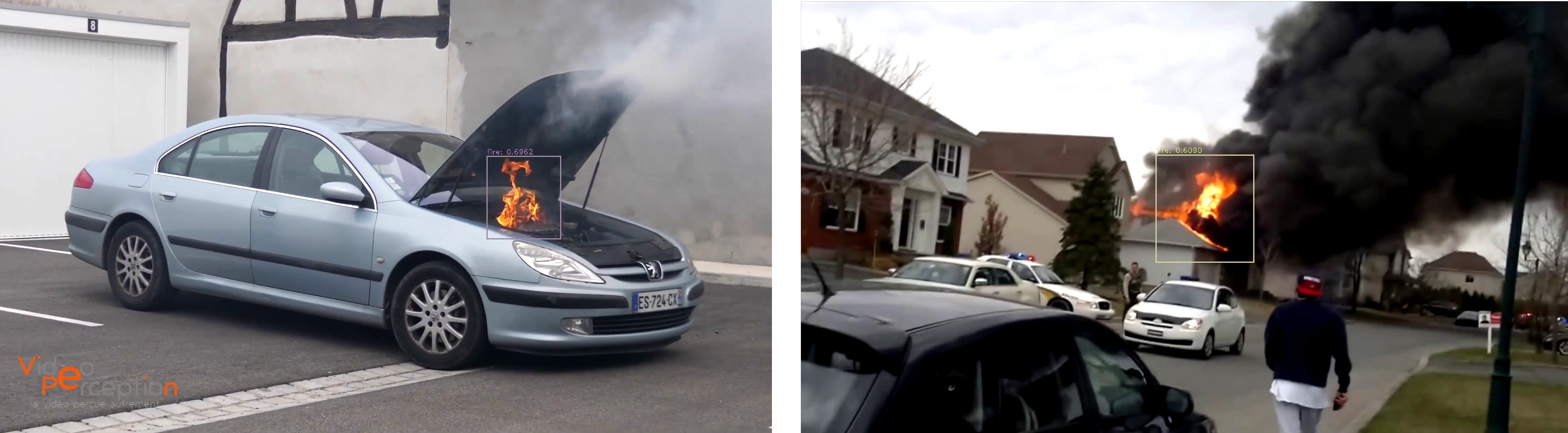}
\caption{Flame Detection via \acrshort{YOLO}V4}
\label{yoloFireExemple}
\end{figure}

After achieving satisfactory results with the \acrshort{YOLO}V4 model trained on a single class, we evaluated the model's performance under anomaly detection conditions. Our initial approach was to define an anomaly as a detected firearm with sufficiently high precision; otherwise, no anomaly would be reported (Figures \ref{yoloGunExemple}, \ref{yoloGun}).
Of course, this condition is somewhat subjective, and the performance of our model depends on the situation and its reliability in detecting these objects. Based on the confusion matrix, it appears that using \acrshort{YOLO} trained to recognize firearms for gunshot detection yields reasonably good performance (Table \ref{yoloGun}, Figure \ref{yoloGunExemple}). Therefore, we decided to apply the same approach to fire detection by training \acrshort{YOLO} to recognize flames (Table \ref{YoloFire}, Figure \ref{yoloFireExemple}).

\begin{table}[H]
\begin{center}
\caption{\acrshort{YOLO}: Gunshot Detection, Confidence at 55\%}
\label{yoloGun}
\begin{tabular}{c | c | c | c}
Accuracy & Precision & Recall & F1-Score \\ \hline
80\% & 16.07\% & 81.19\% & 27.79\% \\ 
\end{tabular}
\end{center}

\begin{center}
\begin{tabular}{c|c c}
\multicolumn{3}{c}{Confusion Matrix} \\ 
\diagbox{Truth}{Predicted} & Gunshot & Normal \\ \hline
Gunshot & \textbf{81.12\%} & 18.88\% \\ 
Normal  & 20\% & \textbf{80\%} \\ 
\end{tabular}
\end{center}
\end{table}

\begin{table}[H]
\begin{center}
\caption{\acrshort{YOLO}: Fire Detection, Confidence at 55\%}
\label{YoloFire}
\begin{tabular}{c | c | c | c}
Accuracy & Precision & Recall & F1-Score \\ \hline
90.7\% & 89.1\% & 99.2\% & 93.8\% \\ 
\end{tabular}
\end{center}

\begin{center}
\begin{tabular}{c|cc}
\multicolumn{3}{c}{Confusion Matrix} \\ 
\diagbox{Truth}{Predicted} & Fire & Normal \\ \hline
Fire & \textbf{69.5\%} & 30.5\% \\ 
Normal & 0.7\% & \textbf{99.3\%} \\ 
\end{tabular}

\end{center}
\end{table}

\section{Conclusion}

Our tests allowed us to define a set of models with the following common features: the choice of \acrshort{YOLO}V7 (even though some of our tests were conducted before this version was released and used V4); the use of \acrshort{GRU} and \acrshort{VGG}19 convolution for our \acrshort{CRNN}; the decision to merge certain classes in our dataset; and the emphasis on traceability. This leaves room for variations, particularly in terms of sequential or parallel modes and the choice between multi-class or binary configurations.
Regarding these variations, our study allows us to propose the following recommendations:

\begin{enumerate}
\item If the goal is to identify the type of anomaly that occurred, the best results are achieved using \acrshort{YOLO}V7 and a multi-class \acrshort{CRNN} in sequential mode. When anomalies are related to human behaviors, applying pre-processing steps such as pose estimation with background removal can yield excellent results.

\item For real-time applications where speed is critical, the best reliability/speed trade-off is obtained using \acrshort{YOLO}V7 and our \acrshort{GRU}-based \acrshort{CRNN} in parallel mode, with the following rules:

\begin{itemize}
\item If an anomaly is detected and a flame is identified, the ongoing event is classified as a fire.

\item If an anomaly is detected and a person is identified with a nearby weapon, the event is classified as a gunshot.
\end{itemize}

\item If the goal is solely to alert about any potential danger, it is preferable to use a binary model (normal/anomalous) instead of a multi-class model. While less precise in identifying the specific type of anomaly, this type of model achieved the best overall performance across all models when combined with \acrshort{YOLO} (in either parallel or sequential mode depending on the requirements).
\end{enumerate}

\part*{} 
\chapter*{Conclusion}
\markboth{Conclusion}{}
\addstarredchapter{Conclusion}

This conclusion will be divided into three parts: the first will be a general summary of my work, the second will discuss its limitations, and the last will outline various perspectives to consider.

\section{General Summary}

Nowadays, many places are equipped with surveillance cameras intended to ensure our safety. All of these cameras cannot be monitored in real-time and are instead used as deterrents, reviewed after events occur. During the live analysis of these images, the monitor is generally responsible for multiple cameras. This surveillance task, therefore, represents a large workload and relies on the monitor’s attention.
We proposed here a set of models that form an anomaly detection system capable of detecting anomalies in a relatively short time from video footage or continuous streams.
To address this issue, I explored two fields related to computer vision: action recognition and object detection. My idea was to combine these two techniques to emulate a human-like behavior that would enable us to detect if an anomaly is occurring, determine its type, and locate it within the image and sequence.
In our research, we observed two main cases leading to anomalies. One is linked to the actions performed by various objects or entities on the screen, and the other is related to particular objects whose mere presence raises suspicion of an anomaly. Our goal is therefore to identify the various objects on the screen and then analyze their behavior to detect any potential anomalies.
In this doctoral project, we focused on supervised approaches. We thus created two datasets. The first consists of images representing firearms, people, and flames, and the second consists of videos depicting fires, fights, and gunshots.
The proposed system comprises two models that can be arranged in parallel or in series according to needs. For object detection, we chose \acrshort{YOLO}V7 for its performance in terms of processing speed and its ability to address various image processing challenges, including human pose analysis. For anomaly detection, we opted for a \acrfull{CRNN} composed of \acrshort{VGG}19 and a \acrshort{GRU}, a model that we trained with a video generator on finite video data, each summarized by a single sequence of images and subsequently tested for video or sequence classification.
It can perform detection on videos in batch mode or detection on 20-frame sequences in a continuous stream. The presented system is thus capable of providing very good results, reaching up to 90\% precision and recall in certain cases. Unfortunately, being a detection system, the anomaly is detected a posteriori. It therefore does not operate in real-time, but its response times are quite fast regardless of the architecture used. In series, we achieve an average refresh rate between 1 and 1.5 seconds, and a rate below 1.2 seconds for parallel mode. This speed could be further improved in parallel mode by configuring \acrshort{YOLO} to ignore certain frames in the sequence, for example, by analyzing every other frame instead of all frames, which is clearly a waste of time for continuous streams where inter-frame differences are minimal. Nonetheless, regardless of the mode used (parallel or series), the processing speeds remain well below a monitor’s reaction time.

\section{Limitations of My Work}

\begin{figure}[H]
\begin{center}
\includegraphics[scale=0.4]{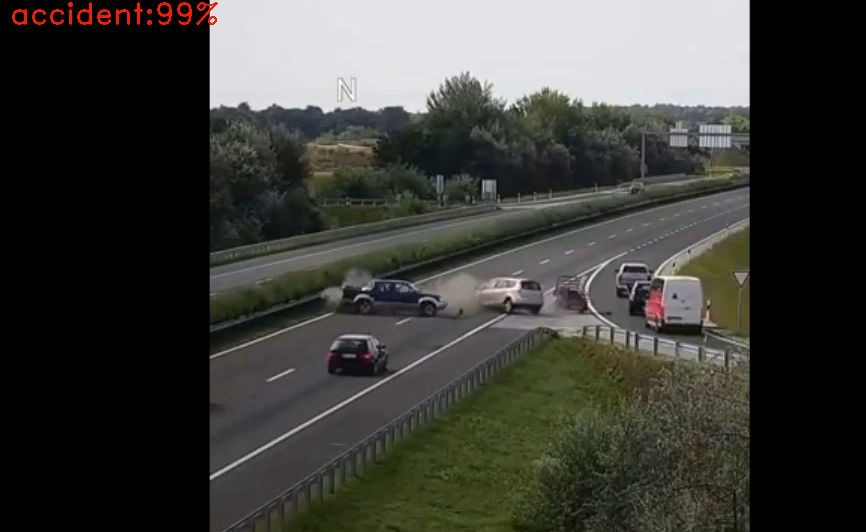}
\caption{Accident Detection}
\label{accidentExemple}
\end{center}
\end{figure}

In this thesis, very few classes were used to train the model. However, thanks to our architecture (\acrshort{CGRU}), it is also possible to achieve good results on other types of anomalies, such as car accidents as shown in Figure \ref{accidentExemple}, provided an appropriate dataset is used.

Despite good performance from our system and acceptable processing times that can still be improved, we have not yet been able to target the area where anomalies occur. \\

We also noted that in certain specific cases, the conditions set to reduce the false-negative rate were not adequate. We wanted, in parallel mode, to change the label of any sequence labeled as normal if a flame or a person with an apparent weapon was detected. Unfortunately, under certain conditions, these behaviors may be considered normal, as in airports where it is common to encounter armed military personnel, a situation that does not necessarily represent a risk. \\

Finally, we note that during this doctoral project, we were unable to test our system on footage from surveillance cameras. Our performance metrics were thus based solely on videos from our test set. Furthermore, due to the fact that data collection and cleaning is a lengthy and costly task, we trained our model using only three classes. Each new anomaly we wish to add will therefore require the collection of new videos or even new images, which may be challenging to obtain in certain contexts. Additionally, in cases where our models are used in parallel, it will sometimes be necessary to develop new detection rules.

\section{Perspectives}

A first potential improvement involves replacing the \gls{IoU} calculation between a person and a firearm in our parallel architecture with a \gls{DIoU} calculation, which would consider the centers of each bounding box. This change would prevent an alert from being triggered if a weapon is detected but merely carried. \\

We could also replace manually defined rules in parallel mode with a machine learning model, such as an Isolation Forest specialized in anomaly detection, or an \acrshort{MLP} accompanied by oversampling, for example. These models could take as input the output of our \acrshort{CGRU} and \acrshort{YOLO}, offering greater reliability while simplifying the addition of new classes. \\

An area that was not thoroughly explored in our work is using spatial analysis in parallel to alter the label of certain frames when a suspicious object is detected, rather than altering the label of the entire sequence (which is the case with the current parallel mode). \\

Other paths to improve our work include using different models or innovative approaches. For example, we could evaluate the performance of a \acrshort{ViT} model, which has its own attention mechanism, in comparison with our \acrshort{CRNN}. It would also be interesting to explore the use of unsupervised neural networks, such as autoencoders or \gls{DINO} (V1, V2), to avoid the need for data collection and labeling each time new classes are added. \\

In the current architecture, another idea to test would be to replace the \acrshort{VGG} convolution with \acrshort{YOLO} within our \acrshort{CRNN}. \acrshort{YOLO} is an object detection neural network architecture based on convolutions, which could provide additional information about detected objects and their positions in each frame of the video. This approach could enhance our model’s ability to capture visual features in each moment of the video and potentially improve its performance. \\

In this thesis, we did not explore image segmentation due to real-time constraints. However, this idea could be interesting to test in anomaly detection cases where processing speed is not crucial. The goal would be to determine if this preprocessing improves the obtained results. To evaluate this approach, we could use the \acrshort{YOLO}v7 architecture specialized for this type of processing or other models such as Faster-\acrshort{RCNN} already mentioned in this thesis or even \acrfull{SAM}, an open-source segmentation model recently proposed by Meta \citep{kirillov2023segany}. \\

In terms of explainability, an interesting approach would be to integrate object detection. Integrating \acrshort{YOLO} with our visualization techniques, such as our activation maps, could help us better understand our model’s decisions and allow us to better represent the area where the detected incident occurred. \\

Finally, although not all videos contain sound, combining our approach with audio data analysis could, in some cases, identify certain anomalies such as gunshots, explosions, or even car accidents with greater reliability.

\printglossary[type=\acronymtype,title=Acronymes,nonumberlist]
\printglossary

\nocite{*}
\printbibliography[heading=bibintoc, keyword={me}, title={My publications}]
\printbibliography[heading=bibintoc, title={Bibliography}]

\listoftables
\listoffigures
\tableofcontents

\end{document}